\pdfoutput=1

\documentclass[11pt]{article}

\usepackage[final]{acl}

\usepackage{times}
\usepackage{latexsym}

\usepackage[T1]{fontenc}

\usepackage[utf8]{inputenc}

\usepackage{microtype}

\usepackage{inconsolata}

\usepackage{graphicx}

\usepackage{booktabs}
\usepackage{array}
\usepackage[table]{xcolor}
\usepackage{amsmath}  
\usepackage{color}
\usepackage{hyperref}

\definecolor{phase1color}{HTML}{EAF2F8} 
\definecolor{phase2color}{HTML}{E9F7EF} 
\definecolor{phase3color}{HTML}{FEF9E7} 
\definecolor{phase4color}{HTML}{FDF2E9} 
\definecolor{phase5color}{HTML}{F4ECF7} 

\title{Beyond the Leaderboard: Rethinking Medical Benchmarks for Large Language Models }

\author{
    Wenting Chen$^{1}$ \quad Guo Yu$^{2}$ \quad Yiu-Fai Cheung$^{3}$  \quad \textbf{Meidan Ding}$^{2}$   \\  
    \textbf{Jie Liu}$^{4}$ \quad \textbf{Zizhan Ma}$^{3}$\thanks{~~Zizhan Ma is the corresponding author.} \quad \textbf{Wenxuan Wang}$^{5}$ \quad \textbf{Linlin Shen}$^{2}$ \\
    \vspace{0.1cm}
    $^1$ Stanford University \quad $^2$ Shenzhen University \quad $^3$ The Chinese University of Hong Kong \\
    $^4$ City University of Hong Kong \quad $^5$ Renmin University of China \\
    \vspace{0.1cm}
    \texttt{wentchen@stanford.edu} \quad
    \texttt{wangwenxuan@ruc.edu.cn} \quad
    \texttt{zizhan.ma@link.cuhk.edu.hk}
}

\begin{document}
\maketitle
\begin{abstract}

Large language models (LLMs) show significant potential in healthcare, prompting numerous benchmarks to evaluate their capabilities. However, concerns persist regarding the reliability of these benchmarks, which often lack clinical fidelity, robust data management, and safety-oriented evaluation metrics. To address these shortcomings, we introduce \textit{MedCheck}, the first lifecycle-oriented assessment framework designed for medical benchmarks. Our framework deconstructs benchmark development into five stages from design to governance, and provides a comprehensive checklist of 46 medically-tailored criteria. Using \textit{MedCheck}, we conducted an in-depth empirical evaluation of 56 medical LLM benchmarks. Our analysis uncovers widespread, systemic issues, including a profound disconnect from clinical practice, a crisis of data integrity due to unmitigated contamination risks, and a systematic neglect of safety-critical evaluation dimensions like model robustness and uncertainty awareness. Based on these findings, \textit{MedCheck} serves as a diagnostic framework to audit existing benchmarks and an actionable guideline for a more standardized, reliable, and transparent approach to evaluating AI in healthcare. 

\end{abstract}

\section{Introduction}
\label{sec:introduction}

Large language models (LLMs) are demonstrating significant potential in healthcare, leading to a proliferation of benchmarks designed to evaluate their capabilities \citep{jin2021what, pal2022medmcqa}. These evaluation tools have evolved from early exam-style question answering (QA) to encompass more complex clinical tasks like report summarization and diagnosis \citep{wu2025bridge, liu2024clinicbench, wu2025medsbench}.

\begin{figure}[t]
  \centering
  \includegraphics[width=1\columnwidth]{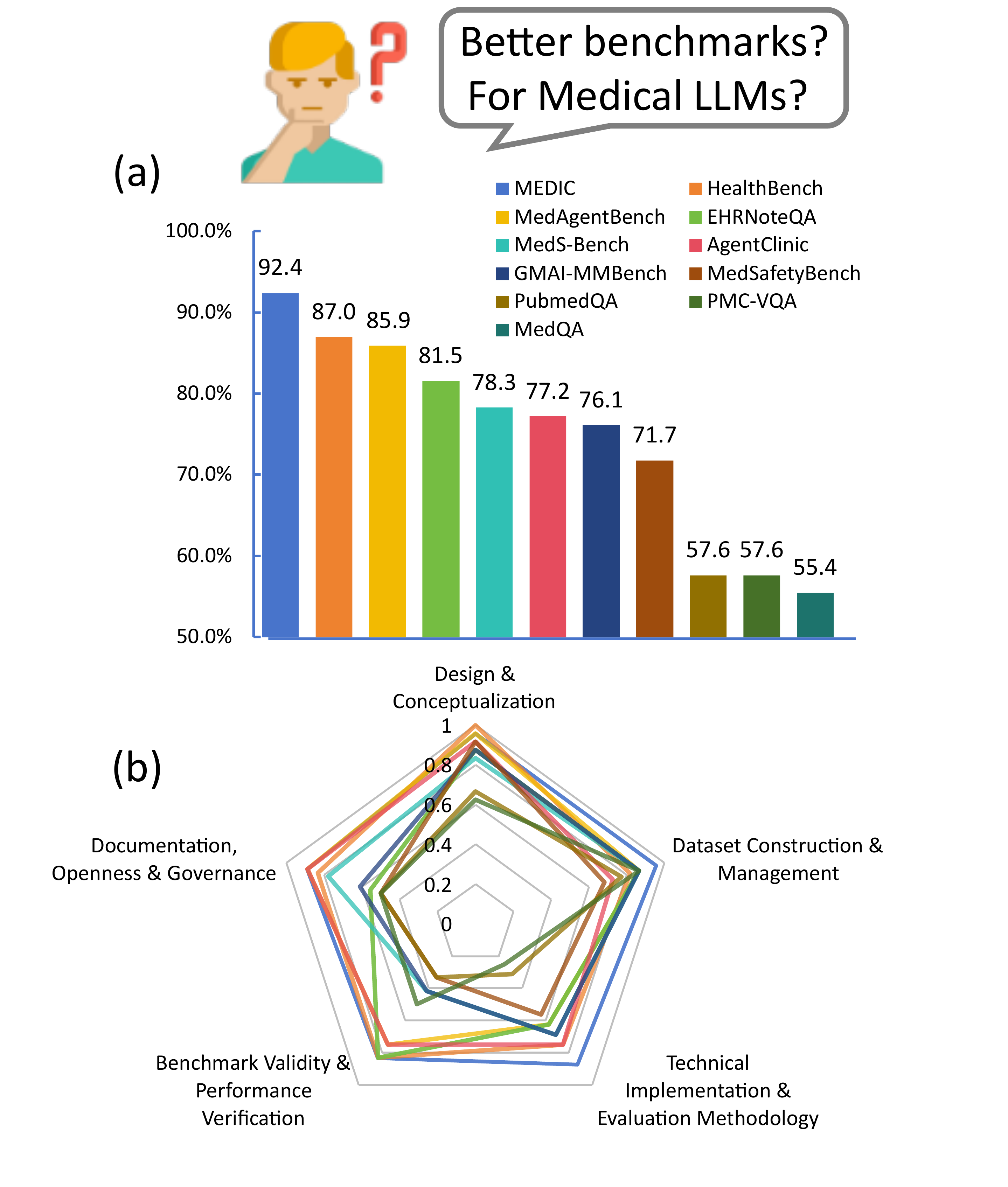}
  \caption{(a) Overall and (b) phase-by-phase performance of medical LLM benchmarks. The significant variance, identified through our MedCheck evaluation, highlights the core motivation for this work: the development of a principled framework for systematic benchmark evaluation.}
  \label{fig:question}
\end{figure}

However, despite their widespread adoption, the reliability of many newer benchmarks is a growing concern. Echoing critiques from within the clinical informatics community \citep{wornow2023shaky}, many evaluations rely heavily on closed-form, multiple-choice questions (MCQA) that, while testing factual knowledge, fail to assess open-ended clinical reasoning or account for real-world variability \citep{wu2025medsbench, zhang2025llmeval}. Furthermore, a significant number are constructed from academic materials rather than authentic clinical data, which raises critical concerns about their clinical fidelity and the risk of data contamination—where models are evaluated on data seen during training \citep{ouyang2024climedbench, wu2025bridge}. These systemic flaws, particularly prevalent in rapidly developed benchmarks designed for general-purpose LLMs, can create an illusion of progress where models are optimized for tasks that lack genuine clinical utility, a problem long recognized by researchers working with real-world electronic health record (EHR) data \citep{wornow2023shaky,blagec2023benchmark,alaa2025medical}.

Recent efforts have proposed general frameworks to improve evaluation quality, such as BetterBench for general-purpose AI \citep{reuel2024betterbench} and How2Bench for code \citep{cao2025how}. However, while valuable, these frameworks are not tailored for the medical domain, which demands a more rigorous approach accounting for specialized terminology, patient data ethics, and the paramount importance of safety.

To develop trustworthy clinical AI, the field must transition from an ad-hoc, publication-driven approach to a disciplined, engineering-oriented paradigm for evaluation \citep{laskar-etal-2024-systematic, yan2024medical-benchmark-survey}. Adopting a lifecycle-aware perspective, similar to principles in mature engineering fields, is a prerequisite for ensuring safety and efficacy in the medical domain \citep{park2020evaluating}.

We introduce \textbf{\textit{MedCheck}}, the first comprehensive, lifecycle-oriented assessment framework for medical LLM benchmarks. \textit{MedCheck} deconstructs benchmark development into five continuous stages, from design to governance, and provides a checklist of 46 medically-tailored criteria. We demonstrate \textit{MedCheck}'s utility by applying it in an in-depth evaluation of 56 medical benchmarks. Our analysis reveals widespread, systemic issues: a profound disconnect from clinical practice, a crisis of data integrity from unmitigated contamination risks, and a systematic neglect of safety-critical dimensions like model robustness and uncertainty awareness.
Our contributions are threefold:
\begin{itemize}
    \item We introduce \textit{MedCheck}, the first comprehensive, lifecycle-oriented evaluation framework with 46 criteria specifically designed for medical benchmarks.
    \item We conduct an in-depth evaluation of 56 medical benchmarks, revealing widespread, systemic weaknesses across the current evaluation landscape.
    \item We offer \textit{MedCheck} as a practical checklist to guide the development of more reliable, transparent, and clinically relevant benchmarks for AI in healthcare.
\end{itemize}

\section{Related works}
\label{sec:related-works}

\subsection{The Evolution of Medical LLM Benchmarks}

Evaluation of medical LLMs has been dominated by benchmarks derived from medical qualification examinations \citep{jin2021what, pal2022medmcqa, liu2023cmexam} and scholarly literature \citep{jin2019pubmedqa, he2020pathvqa}. While useful for assessing foundational medical knowledge, their format and content often lack direct clinical applicability.

To bridge this gap, recent efforts include data-centric frameworks like BigBIO \citep{fries2022bigbio} and public clinical datasets like MIMIC-IV \citep{johnson2023mimic-iv}. There is also a growing emphasis on creating benchmarks that better mirror real-world clinical tasks, such as comprehensive benchmarks that cover report summarization, diagnosis, and treatment planning \citep{wu2025bridge, liu2024clinicbench, wu2025medsbench, bedi2025medhelm}, as well as agentic benchmarks that simulate sequential clinical decision-making \citep{jiang2025medagentbench, schmidgall2024agentclinic}. Although these efforts represent a significant step forward, their development has often outpaced the creation of rigorous standards for their own evaluation, a gap our work aims to fill.

\subsection{Towards a Science of Benchmark Evaluation}

The proliferation of benchmarks has ignited discussion about their reliability and validity. \citet{alaa2025medical} empirically demonstrated a poor correlation between LLM performance on existing benchmarks and in real-world clinical scenarios, highlighting a crisis in the construct validity of benchmarks—the degree to which a test measures what it claims to measure \citep{cronbach1955constructvalidity}. \citet{wornow2023shaky} also finds that foundation modals are often evaluated on tasks that do not provide meaningful insights on their usefulness to health systems. This has catalyzed a new research area focused on the science of benchmarking itself.

\begin{figure*}[h!]
  \centering
  \includegraphics[width=0.9\textwidth]{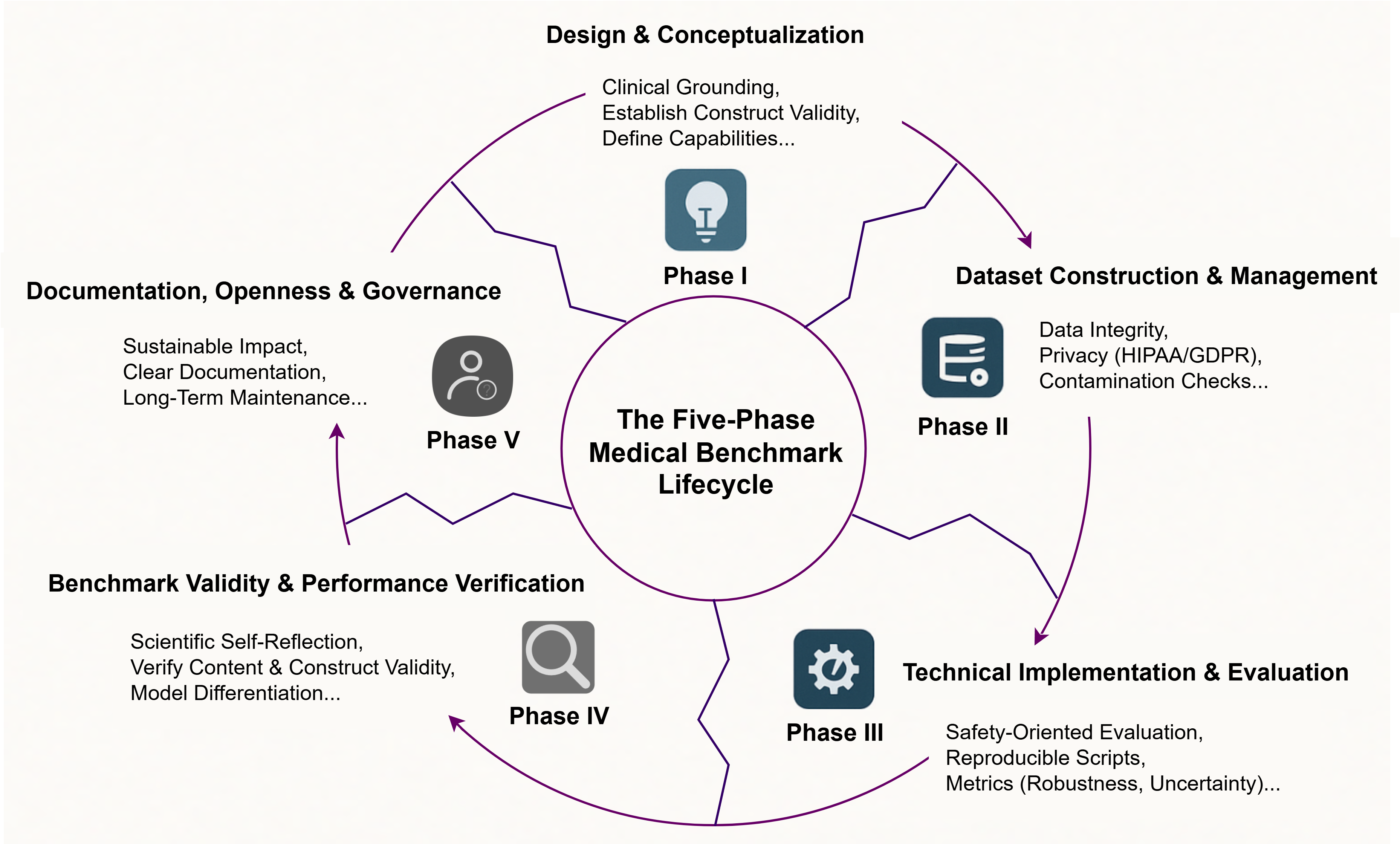}
  \caption{The proposed five-phase lifecycle for medical benchmark engineering. This model illustrates the interconnected and continuous stages of development, from initial design and conceptualization to long-term documentation and governance.}
  \label{fig:lifecycle}
\end{figure*}

Pioneering works have proposed general frameworks to standardize the evaluation of AI benchmarks. For instance, \citet{bhardwaj-etal-2024-machine} developed a checklist for data curation best practices. More comprehensively, BetterBench \citep{reuel2024betterbench} provides a 46-criteria framework for general-purpose AI benchmarks, and How2Bench \citep{cao2025how} offers a 55-item checklist for code-related LLM benchmarks. These frameworks promote a lifecycle-aware perspective, revealing common issues in data quality, reproducibility, and transparency. There are also initiatives like TRIPOD-LLM \citep{gallifant2025tripod-llm} which propose standardized reporting guidelines to enhance transparency and reproducibility.

However, these valuable solutions are either general-purpose or focused on reporting. High-stakes clinical applications demand a more rigorous, context-aware approach that accounts for specialized terminology, patient data ethics, and the paramount importance of safety and reliability. While existing surveys have profiled the medical LLM landscape \citep{liu2024medical-llm-survey, yan2024medical-benchmark-survey}, they stop short of providing a structured framework for assessing benchmark quality. Our work bridges this critical gap by introducing \textit{MedCheck}, the first assessment framework specifically engineered for the unique complexities of the medical domain. 

For a detailed discussion of the foundational evaluation and reporting frameworks of the clinical informatics community, refer to the Appendix \ref{sec:appendix_frameworks}.

\section{Design}
\label{sec:design}

To systematically deconstruct the complexities and shortcomings of existing medical LLM benchmarks, we establish a conceptual framework for their development. This section introduces a novel five-phase lifecycle model for benchmark engineering and details the rigorous methodology employed in our comprehensive analysis.

\subsection{The Five-Phase Medical Benchmark Lifecycle}

Our proposed lifecycle, as depicted in Figure \ref{fig:lifecycle},  provides a structured paradigm for the engineering of high-quality medical benchmarks. Each phase addresses a core set of objectives and potential pitfalls.

\textbf{Phase I: Design and Conceptualization.}

This foundational phase moves beyond mere task definition to establish the benchmark's theoretical underpinnings, focusing on its construct validity—the degree to which it accurately measures its intended theoretical construct, such as "clinical reasoning" \citep{cronbach1955constructvalidity, mehandru2025erreason, schmidgall2024agentclinic}. This phase defines the benchmark's specific, medically relevant purpose and intended contribution, as exam performance may not correlate with real-world utility \citep{alaa2025medical}.

\begin{figure*}[h!]
  \centering
  \includegraphics[width=0.9\textwidth]{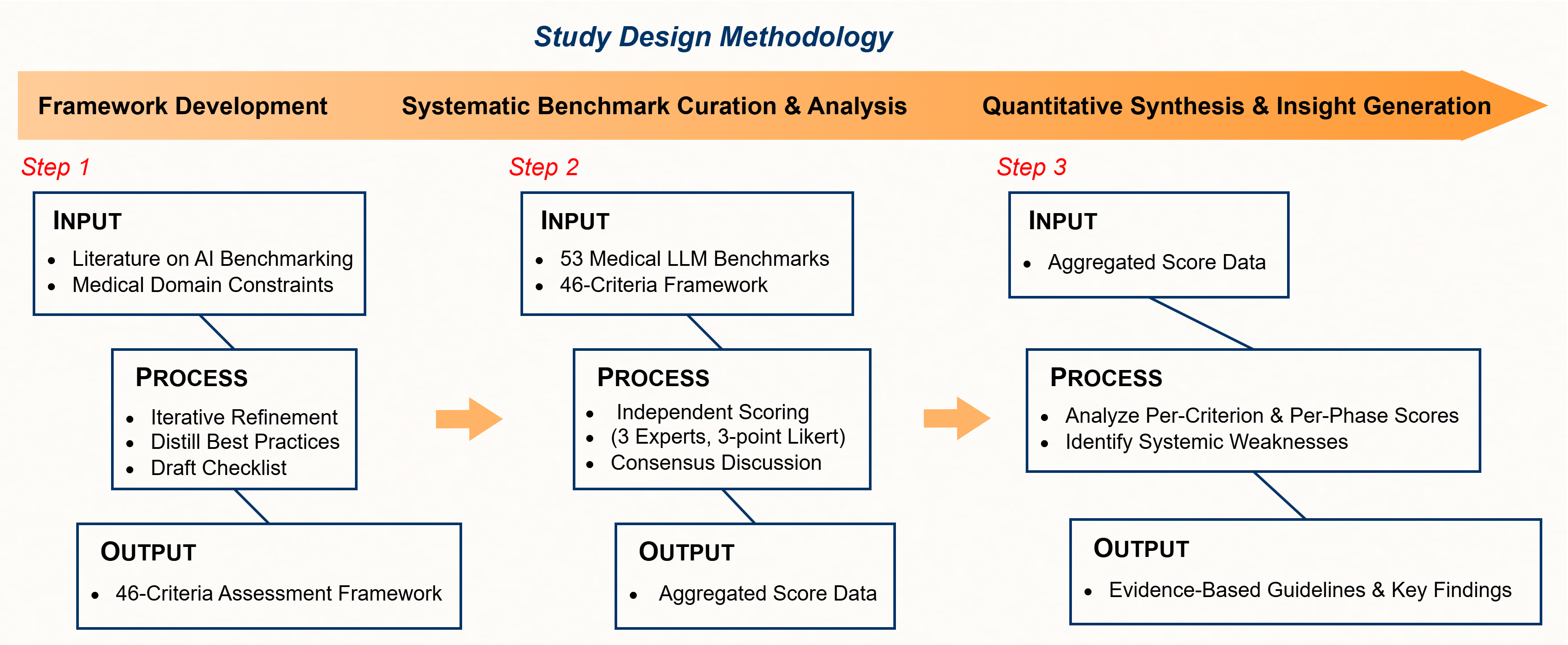}
  \caption{The three-step methodology employed in this study. Our approach involved (1) developing the 46-criteria assessment framework, (2) systematically curating and scoring 56 benchmarks against it, and (3) performing a quantitative synthesis to generate systemic insights.}
  \label{fig:studydesign}
\end{figure*}

\textbf{Phase II: Dataset Construction and Management.} 

This phase constitutes the benchmark's empirical core, focusing on the curation of authentic, diverse, representative, and ethically sourced data. This demands strict adherence to medical privacy regulations like HIPAA and GDPR \citep{bhardwaj-etal-2024-machine, eke2025role}. A primary challenge is proactively addressing data contamination, where evaluation data has been seen by an LLM during training \citep{ouyang2024climedbench, wu2025bridge}. Such contamination leads to inflated performance scores and misleading leaderboards \citep{dong2024-data-contamination, cheng2025-data-contamination}, making rigorous processes for data quality assurance, de-identification, and contamination detection paramount.

\textbf{Phase III: Technical Implementation and Evaluation Methodology.} This operational phase transforms a dataset into an evaluation toolkit with reproducible scripts and, crucially, metrics beyond simple accuracy \citep{chang-etal-2023-survey}. As closed-form MCQA may not assess deep reasoning \citep{wu2025medsbench, molfese-etal-2025-right-answer}, it is vital to evaluate the logical coherence of a model's reasoning \citep{dai2025climb}, its robustness to noisy inputs \citep{han2024medsafetybench}, and its capacity to articulate uncertainty—a cornerstone of safe clinical practice \citep{shelmanov-etal-2024-uq-llms}.

\textbf{Phase IV: Benchmark Validity and Performance Verification.} This phase provides the empirical validation for the benchmark as a measurement instrument. It involves presenting evidence for content validity (the content is representative of the clinical domain) and, critically, substantiating the construct validity claims established in Phase I \citep{alaa2025medical}. A validated benchmark must also demonstrate its ability to reliably differentiate between models of varying capabilities, thus providing a clear signal of progress in the field \citep{reuel2024betterbench}.

\textbf{Phase V: Documentation, Openness, and Governance.} This community-facing phase ensures the benchmark's long-term value and impact. It involves clear documentation and open-source principles for transparency and reproducibility \citep{arvan-etal-2022-reproducibility, cohen-etal-2018-reproducibility}. Critically, it demands a robust governance model for long-term maintenance, versioning, and community feedback.

\textit{MedCheck} evaluates benchmarks as complete engineered systems, not just data collections. It recognizes that even when datasets are reused, significant value can be added through innovative task design, rigorous validation, and sustainable governance.

\subsection{Study Design}

Our three-step methodology (Figure \ref{fig:studydesign}) was designed to ensure objectivity, reproducibility, and depth.

\textbf{Step 1: Framework Development.} We developed the 46-criteria \textit{MedCheck} framework, which serves as the analytical lens for this study. We developed the framework by first conducting an extensive literature review of existing best practices in general AI benchmarking \citep{reuel2024betterbench, cao2025how} and machine learning data curation \citep{bhardwaj-etal-2024-machine}, which we then distilled and iteratively refined into the final 46-criteria framework, grounded in the medical domain's unique ethical and practical constraints, such as patient privacy (e.g., HIPAA compliance), the need for evidence-based standards, and the high stakes associated with potential patient harm \citep{farhud2021ethical, eke2025role}. The complete 46-criteria checklist is provided in Appendix \ref{sec:appendix_full_checklist}.

\textbf{Step 2: Systematic Benchmark Curation and Analysis.} We curated and analyzed a corpus of 56 medical LLM benchmarks, listed in Appendix \ref{sec:appendix_benchmark_list}. The selection process is described in Appendix \ref{sec:appendix_benchmark_search} To ensure objectivity, a rigorous scoring protocol was implemented. Our evaluation process incorporates both LLM-as-judge and domain expert review to balance scalability with accuracy. Specifically, we first provided the paper text and documentation of each benchmark to a large language model and requested an initial scoring based on our 46 criteria. Then, a panel of three NLP researchers with clinical informatics experience reviewed and adjusted these scores on a 3-point Likert scale (0: not met; 1: partially met; 2: fully met). All assessments were based exclusively on publicly available artifacts, including published papers, code repositories, and official websites. Any scoring discrepancies were resolved through a consensus discussion to arrive at a final score for each criterion. We provide the detailed scoring guidelines for the three NLP researcher annotators, along with the full human-in-the-loop annotation protocol, in Appendix~\ref{sec:appendix_evaluation_setup}.

\textbf{Step 3: Quantitative Synthesis and Insight Generation.} The individual scores were aggregated to enable a multi-level quantitative analysis. We calculated per-criterion average scores to identify specific, widespread weaknesses across the field. These were then rolled up into per-phase scores to assess the maturity of each stage in the development lifecycle. Finally, an overall score was computed for each benchmark to gauge its overall quality. This quantitative synthesis allowed us to move beyond anecdotal critiques and identify widespread, descriptive patterns of deficiencies across the entire landscape of medical LLM evaluation, forming the evidence base for the guidelines presented in the following section. A detailed breakdown of these quantitative results can be found in Appendix \ref{sec:appendix_results}.

\section{MedCheck: A Guideline for Medical Benchmarks}
\label{sec:guideline}

Based on our five-phase lifecycle and extensive analysis, we present the \textit{MedCheck} framework as an actionable guideline. For each phase, we explain its core principle and summarize our key findings, further detailed in Appendix \ref{sec:appendix_results}. Our analysis also differentiates between benchmarks designed for foundational medical knowledge and those for clinical practice. A detailed domain-specific analysis can be found in Appendix \ref{sec:appendix_classification}, which shows that our framework appropriately evaluates benchmarks according to their intended purpose. In the tables that follow, each criterion number is hyperlinked to its full definition and scoring rubric in Appendix \ref{sec:appendix_full_checklist}.

\begin{table}[h!]
\centering
\small
\caption{MedCheck Criteria for Phase I: Design and Conceptualization}
\rowcolors{2}{phase1color}{white}
\begin{tabular}{p{0.1\linewidth} p{0.8\linewidth}}
\toprule
\textbf{No.} & \textbf{Description} \\
\midrule
\hyperlink{item:1}{1} & Does the benchmark define the targeted LLM capabilities in medicine for evaluation (e.g., QA, diagnostic reasoning)? \\
\hyperlink{item:2}{2} & Does it describe specific clinical or research applications and their potential value? \\
\hyperlink{item:3}{3} & Does it highlight its unique contribution or innovation compared to existing benchmarks? \\
\hyperlink{item:4}{4} & Does it define the specific LLM functions for evaluation? \\
\hyperlink{item:5}{5} & Does it define the scope of the medical specialties of the benchmark? \\
\hyperlink{item:6}{6} & Is it designed to meet the needs of LLM researchers and medical challenges? \\
\hyperlink{item:7}{7} & Have qualified medical experts been involved in benchmark development? \\
\hyperlink{item:8}{8} & Does it rely on recognized medical sources (e.g., clinical guidelines, databases)? \\
\hyperlink{item:9}{9} & Does it align with international medical standards (e.g., ICD, SNOMED CT, LOINC)? \\
\hyperlink{item:10}{10} & Are the metrics clear and closely related to clinical tasks? \\
\hyperlink{item:11}{11} & Does it assess multiple dimensions (e.g., safety, completeness, interpretability)? \\
\hyperlink{item:12}{12} & Does it consider potential risks and biases in model outputs? \\
\bottomrule
\end{tabular}
\end{table}

\subsection{Guideline for Design \& Conceptualization}
\textbf{Explanation.} This foundational phase anchors a benchmark in clinical relevance and scientific novelty. It necessitates the precise definition of the targeted LLM capabilities, medical scope, and application value, all of which must be grounded in documented domain expertise and authoritative sources. A robust design benefits from considering or aligning with established medical standards or terminologies (e.g., ICD, SNOMED CT) where appropriate to reflect real-world practice and facilitate downstream integration with clinical systems. It should also incorporate the use of clearly defined, multi-dimensional metrics that extend beyond accuracy, and the proactive consideration of safety and bias from inception \citep{arora2025healthbench}.

\textbf{Findings: The Clinical Disconnect.} While nearly all benchmarks (98\%) define high-level objectives, our analysis reveals a systemic issue we term the \textit{Clinical Disconnect}. A total of 50\% (28 of 56) fail to align with any formal medical standards (e.g., ICD, SNOMED CT). Furthermore, 45\% (25 of 56) do not incorporate safety and fairness into their design, and 34\% (19 of 56) evaluate only a single dimension like accuracy, neglecting critical aspects such as completeness. This disconnect stems from an "academic-first, clinical-second" mindset, where developers favor convenient data sources like exam questions from MedQA \citep{jin2021what} and MedMCQA \citep{pal2022medmcqa} over data reflecting complex clinical workflows. This convenience-driven design compromises clinical fidelity and construct validity, leading to models optimized for irrelevant and potentially unsafe tasks \citep{alaa2025medical}.

\begin{table}[h!]
\centering
\small
\caption{MedCheck Criteria for Phase II: Dataset Construction \& Management}
\rowcolors{2}{phase2color}{white}
\begin{tabular}{p{0.1\linewidth} p{0.8\linewidth}}
\toprule
\textbf{No.} & \textbf{Description} \\
\midrule
\hyperlink{item:13}{13} & Are the original sources clearly stated and traceable? \\
\hyperlink{item:14}{14} & Are the sources authoritative and well-justified? \\
\hyperlink{item:15}{15} & Is the data origin clear, and is synthetic data validated? \\
\hyperlink{item:16}{16} & Is the dataset representative of the target population? \\
\hyperlink{item:17}{17} & Are diversity goals defined with supporting quantitative analysis? \\
\hyperlink{item:18}{18} & Is the data properly cleaned and standardized? \\
\hyperlink{item:19}{19} & Are privacy measures described and regulation-compliant? \\
\hyperlink{item:20}{20} & Is the data format consistent and unambiguous? \\
\hyperlink{item:21}{21} & Is expert-involved data review in place? \\
\hyperlink{item:22}{22} & Are reference answers accurate and validated? \\
\hyperlink{item:23}{23} & Are contamination risks detected and handled? \\
\bottomrule
\end{tabular}
\end{table}

\subsection{Guideline for Dataset Construction \& Managemen}
\textbf{Explanation.} This phase addresses the integrity of the benchmark's core asset: its data. This principle mandates the use of traceable and authoritative data sources, with validated authenticity for any synthetic data. The dataset must be demonstrably representative and diverse, supported by quantitative analysis. Rigorous quality control—encompassing cleaning, standardization, expert review, and validated reference answers—is imperative, as are regulation-compliant privacy measures and the proactive mitigation of data contamination risks \citep{ouyang2024climedbench, wu2025bridge}.

\textbf{Findings: A Crisis of Foundational Validity.} Our analysis reveals critical data management weaknesses that undermine the field's empirical foundation. While most benchmarks are reasonably transparent about their primary data sources, subsequent quality control is severely lacking. A staggering 88\% (49 of 56) fail to address data contamination. While post-hoc detection is challenging for closed-source models, the field lacks proactive mitigation strategies within developer control, such as the implementation of canary strings or temporal data cutoffs to explicitly signal exclusion from future pre-training crawls. Furthermore,  66\% (37 of 56) are insufficiently diverse or representative, and 55\% (31 of 56) lack any clear data audit or expert review mechanism. This failure to ensure data integrity threatens the field's validity. Unvetted data may contain factual errors, while contamination leads to artificially inflated scores, creating a false perception of model capability \citep{magar-schwartz-2022-data-contamination, deng2024-data-contamination, bender-etal-2021-dangers}. This renders leaderboards misleading and undermines scientific credibility by building on unreliable evidence \citep{holisticai-2024-data-contamination}.

\begin{table}[h!]
\centering
\small
\caption{MedCheck Criteria for Phase III: Technical Implementation \& Evaluation Methodology}
\rowcolors{2}{phase3color}{white}
\begin{tabular}{p{0.1\linewidth} p{0.8\linewidth}}
\toprule
\textbf{No.} & \textbf{Description} \\
\midrule
\hyperlink{item:24}{24} & Is the evaluation tool easy to install and use? \\
\hyperlink{item:25}{25} & Are detailed documentation and environment settings provided to support reproducibility? \\
\hyperlink{item:26}{26} & Are baseline models or human performance results provided for comparison? \\
\hyperlink{item:27}{27} & Are there evaluations for the model’s reasoning process? \\
\hyperlink{item:28}{28} & Are there evaluations testing the model’s robustness (e.g., input perturbations)? \\
\hyperlink{item:29}{29} & Does the benchmark design help evaluate the generalization ability of models to unseen data? \\
\hyperlink{item:30}{30} & Are there evaluations testing the model’s ability to express uncertainty? \\
\hyperlink{item:31}{31} & Does the benchmark support both closed-source APIs and open-source models? \\
\bottomrule
\end{tabular}
\end{table}

\subsection{Guideline for Technical Implementation \& Evaluation Methodology}
\textbf{Explanation.} This phase focuses on transforming a dataset into a complete evaluation toolkit. It requires the provision of accessible, reproducible, and well-documented code that supports diverse model types and offers performance baselines for context. Fundamentally, the methodology must transcend accuracy to assess safety-critical capabilities: the model's reasoning process, its robustness to input variations, its generalization to unseen data, and its capacity to articulate uncertainty—all of which are vital for trustworthy medical AI \citep{dai2025climb, qiu2025medr, han2024medsafetybench, shelmanov-etal-2024-uq-llms}.

\textbf{Findings: Systematic Neglect of Safety-Critical Capabilities.} This is the most underdeveloped phase in our analysis (average score: 52.4\%), revealing a profound gap between current practice and the needs of reliable medical AI. An alarming 89\% (50 of 56) of benchmarks have no mechanism to test for model robustness, and 91\% (51 of 56) fail to evaluate a model's ability to handle uncertainty. Furthermore, 48\% (27 of 56) neglect the model's reasoning process, focusing only on the final answer. These omissions constitute a systematic neglect of safety. A model's reasoning, robustness, and uncertainty awareness are cornerstones of clinical trustworthiness \citep{farhud2021ethical}. A brittle model that fails with slight data variations is dangerous \citep{bmj-2024-central-oversight}, and an overconfident model that gives incorrect advice without expressing uncertainty is a direct threat to patient safety. By not measuring these capabilities, the community implicitly deems them unimportant, increasing the risk of deploying brittle, opaque, and unsafe systems.

\begin{table}[h!]
\centering
\small
\caption{MedCheck Criteria for Phase IV: Benchmark Validity \& Performance Verification}
\rowcolors{2}{phase4color}{white}
\begin{tabular}{p{0.1\linewidth} p{0.8\linewidth}}
\toprule
\textbf{No.} & \textbf{Description} \\
\midrule
\hyperlink{item:32}{32} & Does the benchmark cover the claimed medical knowledge and skills? \\
\hyperlink{item:33}{33} & Do the tasks realistically simulate clinical settings? \\
\hyperlink{item:34}{34} & Can it distinguish between models of different levels? \\
\hyperlink{item:35}{35} & Are benchmark scores correlated with real-world clinical performance? \\
\hyperlink{item:36}{36} & Does the benchmark demonstrate internal consistency to ensure that different components reliably assess the same capability? \\
\hyperlink{item:37}{37} & Are statistical tests used to verify results? \\
\bottomrule
\end{tabular}
\end{table}

\subsection{Guideline for Benchmark Validity \& Performance Verification}
\textbf{Explanation.} This phase concerns the scientific validation of the benchmark as a measurement instrument. It requires empirical evidence supporting both content validity (comprehensive coverage of the claimed domain) and construct validity (realistic task simulation and accurate measurement of the intended capability \citep{cronbach1955constructvalidity}). A validated benchmark must also demonstrate its utility through proven discriminative power, correlation with real-world clinical performance, high internal consistency, and the application of statistical tests to verify results \citep{reuel2024betterbench}.

\textbf{Findings: The Risk of Misdirected Progress.} Formal scientific validation of benchmarks themselves is rare. While most benchmarks can differentiate models, only 54\% (30 of 56) provide a compelling analysis of their content validity, and just 38\% (21 of 56) are grounded in scenarios with high real-world authenticity. Without proper validation, a benchmark's results are difficult to interpret. Poor content validity creates blind spots, while poor construct validity means the benchmark may not measure the claimed capability at all, leading to misleading conclusions and misdirected research efforts \citep{alaa2025medical}. This lack of self-reflection risks optimizing for metrics that lack genuine clinical utility. To bridge this gap, developers must integrate clinician-in-the-loop validation, ensuring that automated evaluation metrics strictly align with physician preferences and patient safety outcomes.

\begin{table}[h!]
\centering
\small
\caption{MedCheck Criteria for Phase V: Documentation, Openness, and Governance}
\rowcolors{2}{phase5color}{white}
\begin{tabular}{p{0.1\linewidth} p{0.8\linewidth}}
\toprule
\textbf{No.} & \textbf{Description} \\
\midrule
\hyperlink{item:38}{38} & Is there clear, comprehensive benchmark documentation? \\
\hyperlink{item:39}{39} & Are the evaluation criteria and instructions clear and easy to follow? \\
\hyperlink{item:40}{40} & Are limitations and potential risks openly discussed? \\
\hyperlink{item:41}{41} & Has the benchmark undergone formal academic peer review? \\
\hyperlink{item:42}{42} & Are the code and data publicly available with proper licensing? \\
\hyperlink{item:43}{43} & Is a clear usage and citation guideline provided? \\
\hyperlink{item:44}{44} & Is there a clear plan for updates and version control? \\
\hyperlink{item:45}{45} & Is there an public channel for user feedback? \\
\hyperlink{item:46}{46} & Is the long-term maintenance responsibility clearly stated? \\
\bottomrule
\end{tabular}
\end{table}

\subsection{Guideline for Documentation, Openness, \& Governance}
\textbf{Explanation.} This final phase ensures a benchmark's long-term utility and trustworthiness. It requires comprehensive and transparent documentation, including clear instructions and a candid discussion of limitations. Adherence to open-source principles—publicly accessible code and data with appropriate licensing, clear citation guidelines, and academic peer review—is crucial. A robust governance model, which specifies long-term maintenance responsibilities, versioning plans, and a public feedback channel, is essential for sustained relevance and impact.

\textbf{Findings: A Fragmented and Unsustainable Ecosystem.} While most benchmarks provide public access to their assets (55 of 56), the governance required for sustainable impact is critically lacking. A significant 39\% (22 of 56) do not specify a usage license, creating barriers to adoption. Most critically, 80\% (45 of 56) have no clear maintenance plan, and 63\% (35 of 56) lack a public feedback channel. This renders them "fire-and-forget" artifacts destined for obsolescence. This practice fosters a fragmented and unsustainable ecosystem, where effort is wasted on disposable artifacts rather than on building a lasting, reliable infrastructure for scientific progress \citep{arvan-etal-2022-reproducibility}. To ensure longevity, the community must move toward institutional stewardship models, where host organizations provide explicit, long-term funding and technical commitments to prevent valuable benchmarks from becoming static, unmaintained relics.

\section{Discussion}
\label{sec:discussion}

Our analysis of 56 medical benchmarks reveals a field at a critical crossroads. While research is active, systemic deficiencies threaten the validity and utility of this work.

\subsection{Implications: The Urgent Need for a Paradigm Shift}
The status quo is scientifically unsound and clinically irresponsible. The current trajectory fosters an "illusion of progress," where scores on clinically irrelevant or contaminated tasks mask a lack of genuine advancement. By failing to account for evaluation noise, the current paradigm risks optimizing models for pattern memorization rather than the deep, non-linear reasoning required for complex patient care \citep{laskar-etal-2024-systematic, chang-etal-2023-survey}. This misdirects research efforts, misinforms stakeholders, and delays the responsible integration of AI into healthcare by creating brittle, biased systems that could harm patients \citep{obermeyer-etal-2019-dissecting}.

A paradigm shift is urgently needed—a move from ad-hoc dataset creation toward a disciplined, lifecycle-aware practice of benchmark engineering. This requires treating benchmarks not as disposable artifacts for a single paper, but as scientific instruments demanding rigorous design, validation, and maintenance \citep{reuel2024betterbench}.

\subsection{Our Guideline as a Catalyst for Change}
Our five-phase lifecycle and 46-criteria checklist offer a practical toolkit and actionable roadmap for this shift, helping developers create higher-quality benchmarks and users assess existing ones. Adopting this framework can directly mitigate the critical issues identified in our study by promoting: \textit{Medical Grounding} (mandating expert involvement and alignment with medical standards), \textit{Data Integrity} (enforcing transparent sourcing, quality control, and contamination checks), \textit{Safety-Oriented Evaluation} (requiring assessment of reasoning, robustness, and uncertainty), \textit{Scientific Validity} (emphasizing content and construct validation), and \textit{Sustainable Impact} (promoting open practices and long-term governance). Furthermore, it is essential to recognize that the multi-dimensional objectives within the \textit{MedCheck} lifecycle are mutually reinforcing rather than fundamentally incompatible. For instance, stringent data privacy (Phase II) and open-source transparency (Phase V) act as co-existing requirements: rigorous de-identification is a strict prerequisite before a clinical dataset can be ethically open-sourced. High standards in one dimension fundamentally enable and secure practices in the others, creating a cohesive ecosystem for trustworthy AI evaluation.

\subsection{Future Directions: The Next Frontier of Medical Benchmark Research}
Future efforts should move beyond the static, multiple-choice paradigm and pursue three critical directions:

\textbf{Embracing Dynamic and Interactive Benchmarks.} Future benchmarks must reflect the dynamic nature of clinical encounters by assessing sequential decision-making and information gathering. Pioneering works like MediQ \citep{li2024mediq} and AgentClinic \citep{schmidgall2024agentclinic} exemplify this necessary evolution toward evaluating core clinical reasoning. Such interactive setups force models to actively elicit missing patient information, better mirroring the iterative and often ambiguous nature of real-world differential diagnosis \citep{trimble-hamilton-2016-diagnostic-reasoning}.

\textbf{Prioritizing Empirical Construct Validity.} The field must empirically validate that benchmarks truly measure the clinical constructs they claim to \citep{alaa2025medical}. This requires innovative methodologies that correlate benchmark scores with performance on real-world clinical data, such as predicting patient outcomes from EHRs. Without this validation, we risk optimizing for metrics that lack clinical significance.

\textbf{Building a Collaborative Evaluation Ecosystem.} To foster quality and transparency, the community should develop a living repository, akin to platforms like betterbench.stanford.edu \citep{reuel2024betterbench}. Such a platform would enable continuous evaluation of benchmarks against a standardized framework like \textit{MedCheck}. This would create a feedback loop incentivizing higher-quality benchmarks and more informed decisions.

\section{Conclusion}

To address the persistent reliability concerns in medical AI evaluation, we introduced \textit{MedCheck}, a comprehensive, lifecycle-oriented assessment framework comprising 46 criteria. Our empirical evaluation of 56 benchmarks exposed systemic deficiencies across the field, namely a profound disconnect from real-world clinical practice, severe data integrity vulnerabilities stemming from unmitigated contamination risks, and a widespread neglect of safety-critical capabilities. These findings underscore the urgent need to transition from ad-hoc dataset curation toward a disciplined, engineering-focused approach. \textit{MedCheck} serves as the foundational toolkit for this paradigm shift, guiding the community to move beyond superficial leaderboard rankings and rethink medical LLM benchmarks entirely, ultimately fostering the development of genuinely safe, trustworthy, and clinically effective artificial intelligence.

\section*{Limitations}

We acknowledge this study's limitations. First, our analysis of 56 benchmarks, while extensive, is not exhaustive given the field's rapid growth. Second, the scoring process, despite a rigorous protocol, retains a degree of subjectivity inherent in qualitative assessment. Third, our findings are based exclusively on public artifacts, which may not capture all unpublished development practices. Finally, the \textit{MedCheck} framework itself is a snapshot of current best practices and will require future revisions to keep pace with evolving AI capabilities and ethical standards. Specifically, as multimodal and autonomous medical agents emerge, future iterations must expand to rigorously audit multimodal clinical grounding and agentic safety.


\bibliography{custom}
\clearpage
\appendix

\section*{Appendix}
\label{sec:appendix}

In this appendix, we provide additional information about \textit{MedCheck}. Appendix \ref{sec:appendix_frameworks} discusses clinically-grounded evaluation and reporting frameworks. Appendix \ref{sec:appendix_benchmark_search} shows the benchmark searching and selecting procedure, and Appendix \ref{sec:appendix_evaluation_setup} details our evaluation setup. Appendix \ref{sec:appendix_classification} outlines the classification of clinical and medical benchmarks. Appendix \ref{sec:appendix_benchmark_list} lists the 56 evaluated medical LLM benchmarks, and Appendix \ref{sec:appendix_results} provides extra quantitative analysis of the evaluation results. Appendix \ref{sec:appendix_full_checklist} shows the complete list of evaluation criteria. Finally, Appendix \ref{sec:appendix_case_study} provides detailed scoring and explanations for a representative benchmark, and Appendix \ref{sec:appendix_clinical_actionable_report} presents an actionable diagnostic report example based on a clinical-oriented benchmark. 

\section{Clinically-Grounded Evaluation and Reporting Frameworks}
\label{sec:appendix_frameworks}

Beyond the benchmark datasets themselves, clinical informatics researchers have developed a parallel stream of work focused on establishing rigorous methodologies for AI model evaluation, validation, and transparent reporting. This research, published in top-tier medical and clinical informatics journals, emphasizes clinical utility, patient safety, and real-world applicability over leaderboard performance alone. A cornerstone of this ecosystem is the decades-long MIT and Harvard Medical School effort to develop, de-identify, and ethically share large-scale EHR datasets like MIMIC. The MIMIC project represents more than a data source—it embodies the principles of data curation, long-term maintenance, and community governance that are central to \textit{MedCheck}'s framework, particularly in Phase II and Phase V.

Recent work emerging from this clinically-grounded tradition has produced several critical resources. The TRIPOD-LLM statement, for example, provides the first reporting guideline specifically for studies using LLMs, ensuring that methods and results are communicated transparently (Gallifant et al., 2025). While TRIPOD-LLM guides the reporting of a study, our \textit{MedCheck} framework is a complementary tool designed to assess the quality of the benchmark used within that study. In parallel, comprehensive evaluation platforms like Med-HELM have been developed to assess LLMs on a wide array of clinically-grounded tasks using real EHR data. Med-HELM's focus is on holistically evaluating the model's capabilities, whereas \textit{MedCheck}'s focus is on the prior, meta-level task of evaluating the measurement instrument itself.

The very premise of evaluating LLMs on medical tasks is predicated on the knowledge they contain. The landmark work by Singhal et al. (2023) was among the first to systematically demonstrate that LLMs do, in fact, encode a substantial amount of clinical knowledge. This finding underscores the urgency and importance of developing robust, clinically-valid benchmarks to accurately measure and safely harness this encoded knowledge, moving beyond exam-style questions to tasks that reflect the complexity of real-world clinical practice.

\section{Benchmark searching and selection}
\label{sec:appendix_benchmark_search}

To ensure the objectivity and comprehensiveness of our evaluation, we implemented a systematic literature search and selection protocol divided into three stages: identification, screening, and eligibility. We first queried major academic databases—including Google Scholar, ACL Anthology, arXiv, and PubMed—for publications from January 2018 to July 2025. The search strategy employed Boolean combinations of keywords across subject, domain, and artifact type, specifically incorporating terms such as "medical LLM," "medical benchmark," "healthcare," "clinical," "clinical decision," and data-specific identifiers like "EHR" (electronic health record) and "EMR." This initial retrieval, focused on titles and abstracts to ensure relevance, yielded a total of 482 candidate publications after removing duplicates.

Subsequently, we applied stringent exclusion criteria to filter these candidates. We excluded studies strictly limited to traditional discriminative paradigms (e.g., BERT-based classification) that have not been adapted for evaluating the generative reasoning capabilities of LLMs. Furthermore, we removed benchmarks lacking publicly accessible data or code repositories essential for our reproducibility analysis, as well as studies that merely re-evaluated existing datasets without introducing novel contributions. This screening process narrowed the candidate pool to 85 potential benchmarks.

From this screened corpus, we selected the final set of 56 benchmarks based on specific criteria regarding impact and community adoption. Addressing the ambiguity of "widely used," we defined significance through quantitative metrics tailored to the publication timeframe. For established benchmarks published prior to 2023, we required a minimum citation count of 50 to verify community adoption. Conversely, for emerging benchmarks published in 2024 and 2025, we selected those that were either accepted by top-tier venues (e.g., ACL, NeurIPS, Nature Medicine) or represented a unique task category not covered by older benchmarks. This multi-stage protocol ensures that our analysis captures the most significant and representative contributions to the medical LLM landscape.

\section{Evaluation Setup}
\label{sec:appendix_evaluation_setup}

In the evaluation process of this study, the primary focus was on manual scoring conducted by domain experts, with the use of LLMs providing supplementary assistance. While LLM were strictly instructed to follow the evaluation criteria, they inevitably exhibited hallucinations. Thus, the manual evaluation process remained the cornerstone of the benchmark assessment, with LLM serving as a tool to facilitate expert review by quickly retrieving relevant information and providing references for scoring.

This section will first describe the manual evaluation process, outlining how human annotators were trained, calibrated, and involved in scoring. It will then explain how the LLM-as-judge approach was implemented to enhance efficiency and assist in the evaluation process.

\subsection{Guideline for annotator}
To ensure reliable and consistent scoring of benchmarks according to the proposed evaluation framework, we established a rigorous annotation protocol executed by three expert annotators. The process comprised two phases: (1) a preparation and calibration phase, and (2) a formal scoring phase, as detailed below.

\label{sec:guideline for annotator} 

\begin{table}[h!]
\centering
\small
\caption{Overall Scoring Criteria}
\label{tab:scoring_criteria} 
\begin{tabular}{p{0.1\linewidth} p{0.85\linewidth}}
\toprule
\textbf{Score} & \textbf{Description} \\
\midrule
0 & The criterion is completely absent, severely lacking, or contains fundamental flaws. \\
  & \textit{Judgment basis:} No relevant information; clearly incorrect information. \\
\midrule
1 & Initial attempt made, but incomplete, lacking details, methodologically crude, or unvalidated. \\
  & \textit{Judgment basis:} Mentioned but not implemented; implemented but without evaluation; described ambiguously, making validation impossible. \\
\midrule
2 & Clear, complete, transparent, and aligned with current domain consensus or literature-recommended best practices. \\
  & \textit{Judgment basis:} Explicit description + implementation + validation; citation of authoritative methods; open-source / reproducible / peer-reviewed / recognized by the community. \\
\bottomrule
\end{tabular}
\end{table}

\paragraph{Preparation and Calibration Phase}
    
One week prior to scoring, all three annotators were provided with the full evaluation criteria (Appendix ~\ref{sec:appendix_full_checklist}). After confirming that each annotator had thoroughly reviewed and understood the criteria, we convened an online calibration meeting to accomplish the following objectives:
(a) Jointly review the evaluation criteria;
(b) Discuss and resolve discrepancies, with particular attention to clarifying ambiguous standards or dimensions with similar names, and establish consensus on terminology and definitions;
(c) Harmonize overall scoring expectations (As shown in Table~\ref{tab:scoring_criteria});
(d) Clarify acceptable sources of evidence: scores must be grounded in publicly accessible documentation (e.g., main paper text, appendices, GitHub repositories, technical reports, official websites). If required information is absent, the item receives a default score of 0, with a comment stating “Information not provided.”

After each annotator completed scoring at least three benchmarks, a second alignment meeting was held to address emerging ambiguities or inconsistencies encountered during initial scoring. Based on this discussion, all previously submitted scores were reviewed and revised as necessary to maintain inter-annotator consistency.

To ensure reliability and directly address the limitation of single-annotator-per-case for the main corpus, we established a shared evaluation subset of 5 randomly selected benchmarks before the formal scoring phase. All three annotators independently scored this subset. We calculated the inter-annotator agreement using Fleiss' Kappa, achieving a strong agreement score ($\kappa = 0.78$), which indicates substantial consensus on the interpretation of the criteria.

\paragraph{Formal Scoring Phase}
    
A total of 56 benchmarks were uniformly and randomly distributed among the annotators. Each annotator received a randomly ordered list of benchmarks, accompanied by the LLM-generated scores and corresponding explanations for each benchmark, and was instructed to manually evaluate 1–2 benchmarks per day, adhering to the following requirements:

(a) For each scored item, provide specific evidence in the comment field (As illustrated in the “Explanation” section of Appendix~\ref{sec:appendix_case_study});
(b) Ignore authorship and institutional affiliations; avoid basing judgments on general impressions or prior familiarity with a benchmark;
(c) Apply criteria strictly as defined—do not broaden or narrow the scope based on personal interpretation;
(d) The LLM-generated scores and explanations are intended solely to serve as auxiliary aids—facilitating rapid information retrieval and offering preliminary scoring suggestions—and must not constitute the primary basis for evaluation. Their content must be rigorously verified for factual accuracy before being considered as reference material;
(e) For items with insufficient or ambiguous information, assign a score of 0, 1, or 2 as appropriate, but include a brief justification in the comments (e.g., “The paper mentions data cleaning but does not detail the procedure”) and prefix the note with “[Uncertain]”;
(e) For borderline cases where a benchmark's documentation is vague or implicitly implies a feature, annotators were strictly instructed to default to the lower score to penalize the lack of transparency. A higher score was only awarded if concrete, verifiable evidence (e.g., explicit code snippets, specific GitHub commits, or dedicated appendix sections) could be explicitly cited in the comments. Any remaining uncertainties (flagged with "[Uncertain]") were brought to the weekly expert panel meetings, where consensus was reached through rigorous group discussion and majority voting;
(f) Daily monitoring of individual annotator score distributions was conducted to detect and mitigate systematic bias (e.g., consistently high or low scoring tendencies).

Throughout the process, any issues or discrepancies in scoring were resolved through weekly consensus meetings among the expert panel, culminating in a final agreed-upon score for each criterion. This rigorous protocol substantially mitigated the well-documented susceptibility of manual scoring to various confounding factors—such as subjective bias, cognitive load, and order effects—and ensured that the evaluation process was maximally transparent, consistent, and reproducible.

\subsection{LLM-as-Judge Scoring}
While the primary evaluation in MedCheck is conducted manually by domain experts, we explored the use of LLM-as-judge scoring during the preliminary scoring phase to balance scalability with accuracy.

We include our evaluation criteria, the paper text, and documentation of medical benchmarks in the prompt. We then use GPT-4o-Search-Preview with temperature = 0 to generate scores in JSON format. To ensure the LLM produces reasonable scores based on our criteria, we explicitly instructed it to strictly adhere to the evaluation guidelines and provide an explanation for each score. Additionally, to guarantee the LLM has access to up-to-date public documentation, we enabled web search functionality, allowing it to retrieve supplementary information (e.g., the benchmark’s official website, GitHub repository, etc.) when necessary. During the manual scoring process, the panel of domain experts refers to the LLM-generated scores and explanations, but they must verify the accuracy of the LLM-provided explanations and ultimately assign the final scores.

To address potential concerns regarding confirmation bias from LLM-generated scores, we explicitly frame our approach as an \textbf{"Assisted-Verification"} design. In the context of auditing extensive and complex benchmark documentation, independent "blind" human annotation often suffers from high false-negative rates due to cognitive fatigue and information overload (i.e., annotators inadvertently missing existing evidence). In our protocol, the LLM functions strictly as an information retrieval assistant to surface relevant sections and ensure high recall. The human expert then rigorously verifies the factual existence of the cited documentation to ensure high precision. Because the assessment relies on objective verification of factual evidence rather than subjective judgment, the risk of bias is minimized while the accuracy of evidence retrieval is significantly improved.

In summary, the evaluation process relied fundamentally on the expertise of human annotators to ensure the accuracy and consistency of the final scores. LLMs played a strictly auxiliary role—expediting data retrieval and providing preliminary scoring suggestions—while the ultimate judgment remained firmly in the hands of domain experts. This hybrid methodology synergistically combines the rigor and reliability of manual evaluation with the efficiency and scalability of LLMs, thereby ensuring an evaluation framework that is both comprehensive and practically viable.

\section{Classification of Clinical and Medical Benchmarks}
\label{sec:appendix_classification}

We classify the 56 evaluated benchmarks into two distinct categories: \textbf{Clinical (42 benchmarks, 75\%)} and \textbf{Medical (14 benchmarks, 25\%)}. These categories exhibit a fundamental distinction in their objectives, application scenarios, and data sources. We argue that evaluating a benchmark from one category using the criteria intended for the other would be methodologically flawed. Therefore, to ensure a fair and context-aware assessment, the criteria used to define each category are detailed below.

\begin{itemize}
    \item \textbf{Clinical Benchmarks (42/56, 75\%):} These benchmarks primarily evaluate the application of LLMs in real-world clinical workflows.
    \begin{itemize}
        \item \textbf{Objective:} Assess capabilities like processing EHRs, clinical reasoning with dynamic or incomplete information, and supporting diagnostic decisions.
        \item \textbf{Scenario:} Simulate authentic clinical encounters, such as patient consultations, risk prediction, or treatment planning.
        \item \textbf{Data Source:} Primarily use real-world data, including EHRs, clinical case notes, and doctor-patient dialogues.
    \end{itemize}

    \item \textbf{Medical Benchmarks (14/56, 25\%):} These benchmarks focus on assessing an LLM's foundational medical knowledge and theoretical reasoning.
    \begin{itemize}
        \item \textbf{Objective:} Test the model's mastery of established medical facts and concepts.
        \item \textbf{Scenario:} Typically involve standardized, knowledge-based tasks, often in a multiple-choice question format.
        \item \textbf{Data Source:} Primarily use academic materials, such as medical exam questions, textbooks, and research literature.
    \end{itemize}
\end{itemize}

\section{List of Evaluated Benchmarks}
\label{sec:appendix_benchmark_list}

We assessed the following 56 benchmarks, categorized as either Clinical (C) or Medical (M) (alphabetical order):
\begin{itemize}
    \item AfriMed-QA \cite{adebayo2024afrimed} (C)
    \item AgentClinic \cite{schmidgall2024agentclinic} (C)
    \item Asclepius \cite{liu2025asclepius} (C)
    \item BRIDGE \cite{wu2025bridge} (C)
    \item CHBench \cite{guo2024chbench} (C)
    \item CheXpert \cite{irvin2019chexpert} (C)
    \item CLIMB \cite{dai2025climb} (C)
    \item CliMedBench \cite{ouyang2024climedbench} (C)
    \item ClinicBench \cite{liu2024clinicbench} (C)
    \item CMB \cite{wang2023cmb} (M)
    \item cMedQA2 \cite{zhang2018multi} (C)
    \item CMExam \cite{liu2023cmexam} (M)
    \item COGNET-MD \cite{panagoulias2024cognet} (M)
    \item DataDEL \cite{yang2024datadel} (C)
    \item DR.BENCH \cite{gao2023dr} (C)
    \item EHRNoteQA \cite{kweon2024ehrnoteqa} (C)
    \item EndoBench \cite{liu2025endobench} (C)
    \item ER-REASON \cite{mehandru2025erreason} (C)
    \item GMAI-MMBench \cite{chen2024gmai} (C)
    \item HeadQA \cite{vilares2019head} (M)
    \item HealthBench \cite{arora2025healthbench} (C)
    \item LCD \cite{yoon2025lcd} (C)
    \item LLMEval-Med \cite{zhang2025llmeval} (C)
    \item MedAgentBench \cite{jiang2025medagentbench} (C)
    \item MedAgentsBench \cite{tang2025medagentsbench} (M)
    \item MedCalc \cite{khandekar2024medcalc} (C)
    \item MedChain \cite{liu2024medchain} (C)
    \item MedConceptsQA \cite{shoham2024medconceptsqa} (M)
    \item MEDEC \cite{abacha2025medec} (C)
    \item MedExQA \cite{kim2024medexqa} (M)
    \item MEDIC \cite{kanithi2024medic} (C)
    \item MediQ \cite{li2024mediq} (C)
    \item MedJourney \cite{wu2025medjourney} (C)
    \item MedMCQA \cite{pal2022medmcqa} (M)
    \item MedOdyssey \cite{fan2025medodyssey} (C)
    \item MedQA \cite{jin2021what} (M)
    \item MedR-Bench \cite{qiu2025medr} (C)
    \item MedRisk \cite{liu2025medrisk} (C)
    \item MedSafetyBench \cite{han2024medsafetybench} (C)
    \item MedS-Bench \cite{wu2025medsbench} (C)
    \item MIMIC-IV-ED \cite{xie2022benchmarking} (C)
    \item MLEC-QA \cite{li2021mlec} (M)
    \item MMMU (Health \& Medicine) \cite{yue2024mmmu} (C)
    \item MVME (AI hospital) \cite{fan2025mvme-ai-hospital} (C)
    \item OmniMedVQA \cite{hu2024omnimedvqa} (C)
    \item PathMMU \cite{sun2025pathmmu} (C)
    \item PathVQA \cite{he2020pathvqa} (M)
    \item PMC-VQA \cite{zhang2024pmc} (C)
    \item PubMedQA \cite{jin2019pubmedqa} (M)
    \item ReasonMed \cite{sun2025reasonmed} (M)
    \item SLAKE \cite{liu2021slake} (C)
    \item TrialPanorama \cite{wang2025trialpanorama} (C)
    \item VQA-Med \cite{ben2021overview} (C)
    \item VQA-RAD \cite{lau2018dataset} (C)
    \item webMedQA \cite{he2019webmedqa} (C)
    \item XMedBench \cite{wang2024xmedbench} (M)
\end{itemize}

\section{Quantitative Results}

\label{sec:appendix_results}
This appendix provides a summary of the quantitative analysis conducted on the 56 medical LLM benchmarks.

\subsection{Publication Year and Task Focus of Benchmarks}
Our collection covers open-source medical large model benchmarks published between 2018 and 2025.

\begin{figure}[htb]
\centering
\includegraphics[width=0.5\textwidth]{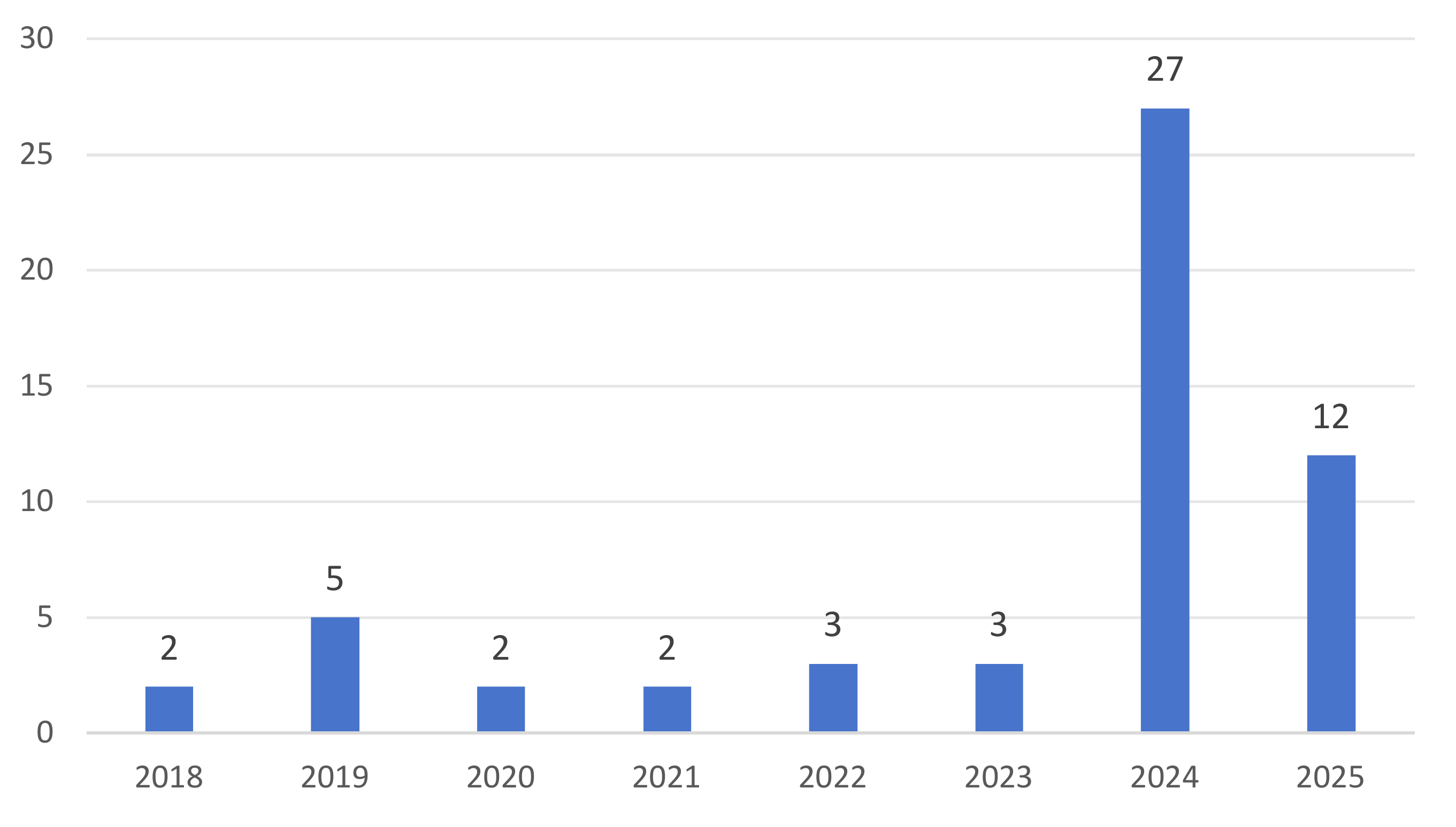}
\caption{Publication Year Distribution of the 56 Medical Benchmarks}
\label{fig6}
\end{figure}

\textbf{Annual Trend.} As shown in Figure \ref{fig6}, the publication years of the 56 benchmarks we collected. A clear turning point can be observed in 2024, when the volume of medical benchmark publications significantly increased. Prior to 2024, only 17 benchmarks were released. The brief surge in 2019, driven by the rise of deep learning and Transformer models, catalyzed short-term growth in medical AI and a peak in benchmark publications. However, no substantial upward trend was evident before 2024.

In contrast, with the emergence of large models such as ChatGPT and Med-PaLM in 2024, there was an explosive growth in medical NLP and multimodal benchmarks. A total of 27 medical benchmarks were published in 2024, surpassing the cumulative number of the previous six years. As of July 10, 2025, 12 new benchmarks were already released in the first half of 2025, showing a steady growth trend.

\begin{figure}[htb]
\centering
\includegraphics[width=0.5\textwidth]{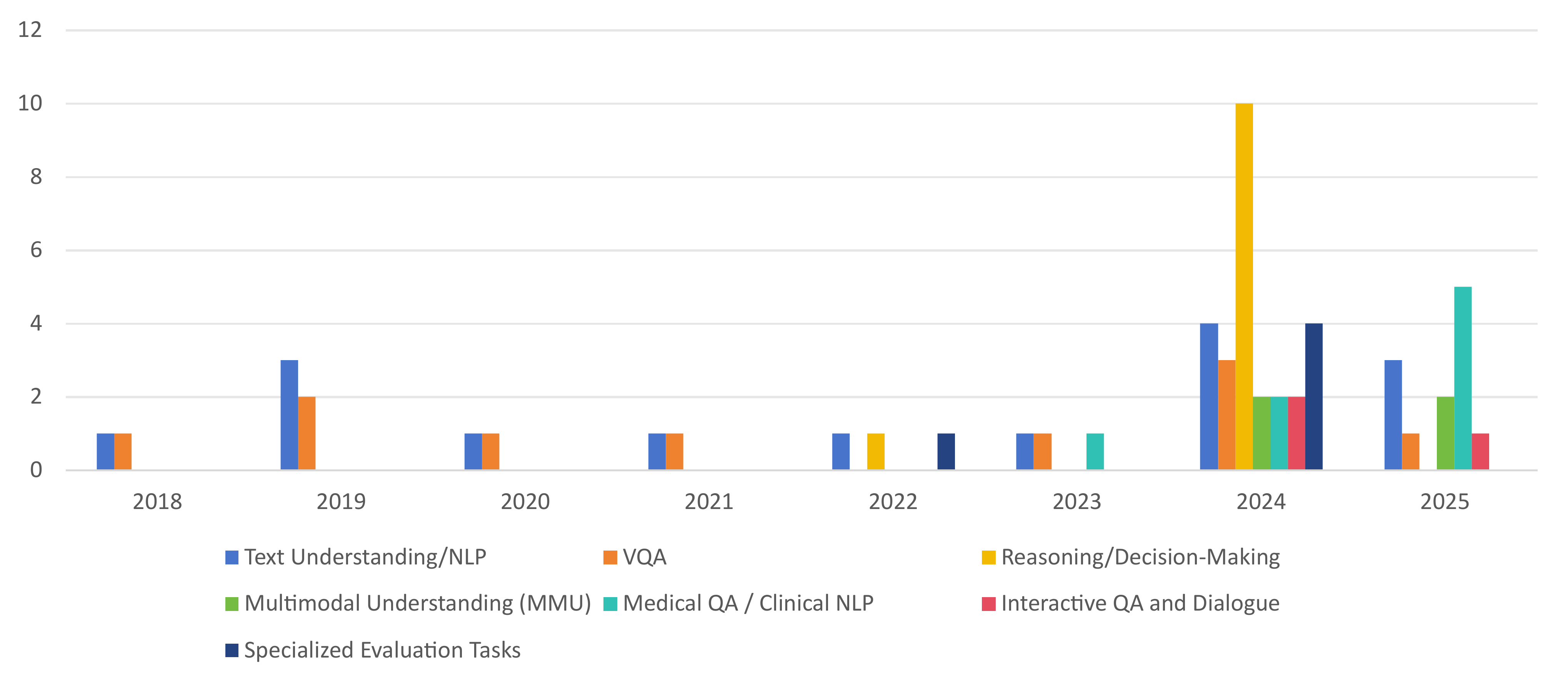}
\caption{Temporal Distribution of Benchmarks by Task Type}
\label{fig7}
\end{figure}

\textbf{Task Trend.} Before 2024, most benchmarks primarily focused on evaluating models' capabilities in NLP and VQA. As shown in Figure \ref{fig7}, following the explosion in benchmark quantity post-2024, the range of tasks covered expanded to six categories, with new benchmarks addressing specific tasks for large models, such as multimodal understanding, reasoning and decision-making, dialogue/interactive QA , as well as medical text interpretation and safety in medical recommendation generation.

\begin{figure}[htb]
\centering
\includegraphics[width=0.5\textwidth]{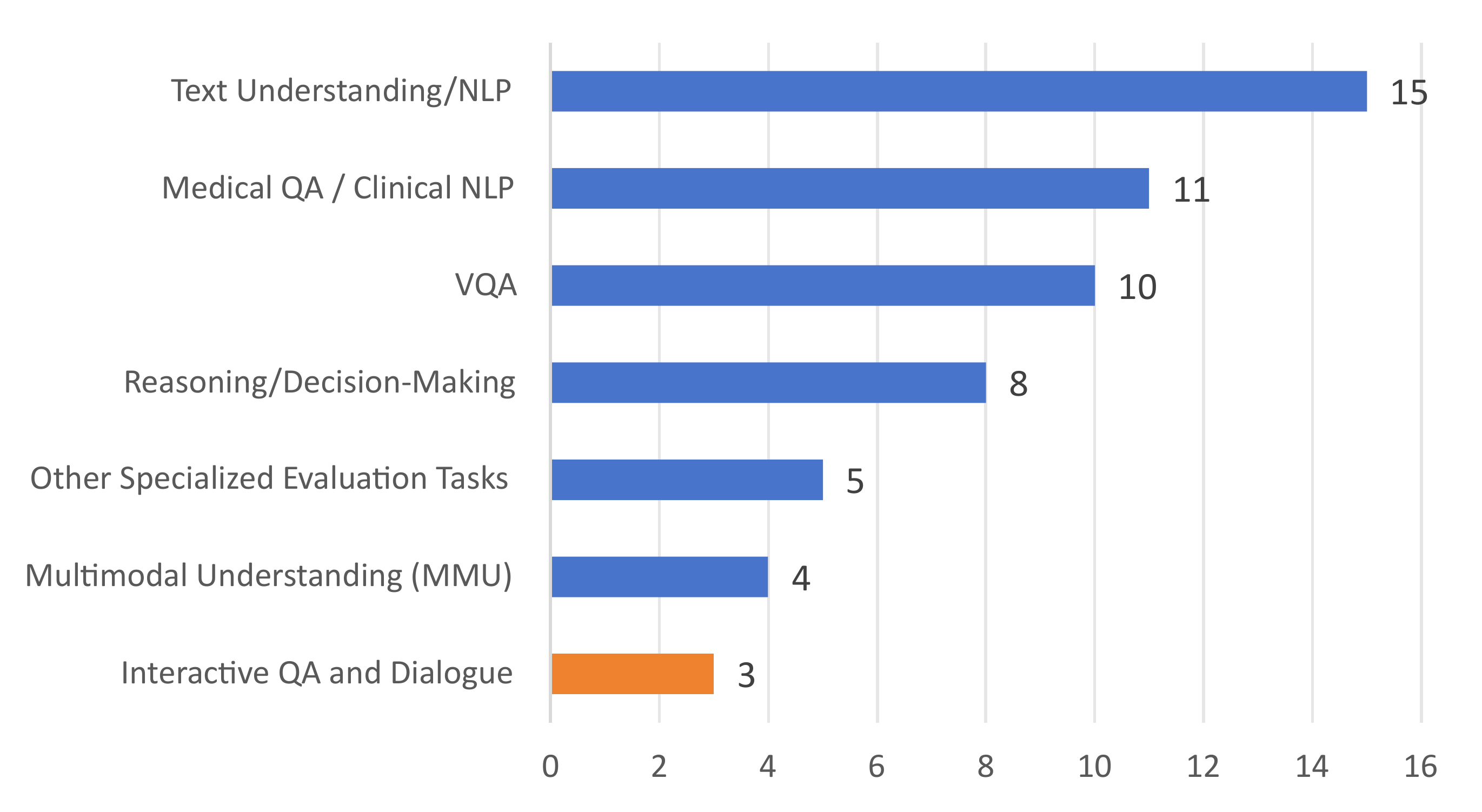}
\caption{Task Type Distribution of the 56 Medical Benchmarks}
\label{fig8}
\end{figure}

In summary, as illustrated in Figure \ref{fig8}, more than half (64\%) of the existing medical benchmarks focus on assessing models' text comprehension, NLP, and VQA abilities. This indicates that most medical benchmarks historically centered around structured or unstructured medical text understanding. However, in recent years, there has been an increasing trend towards medical question answering and clinical NLP, with the highest number of such benchmarks being published in the first half of 2025.

\subsection{Natural Language}
\begin{figure}[htb]
\centering
\includegraphics[width=0.5\textwidth]{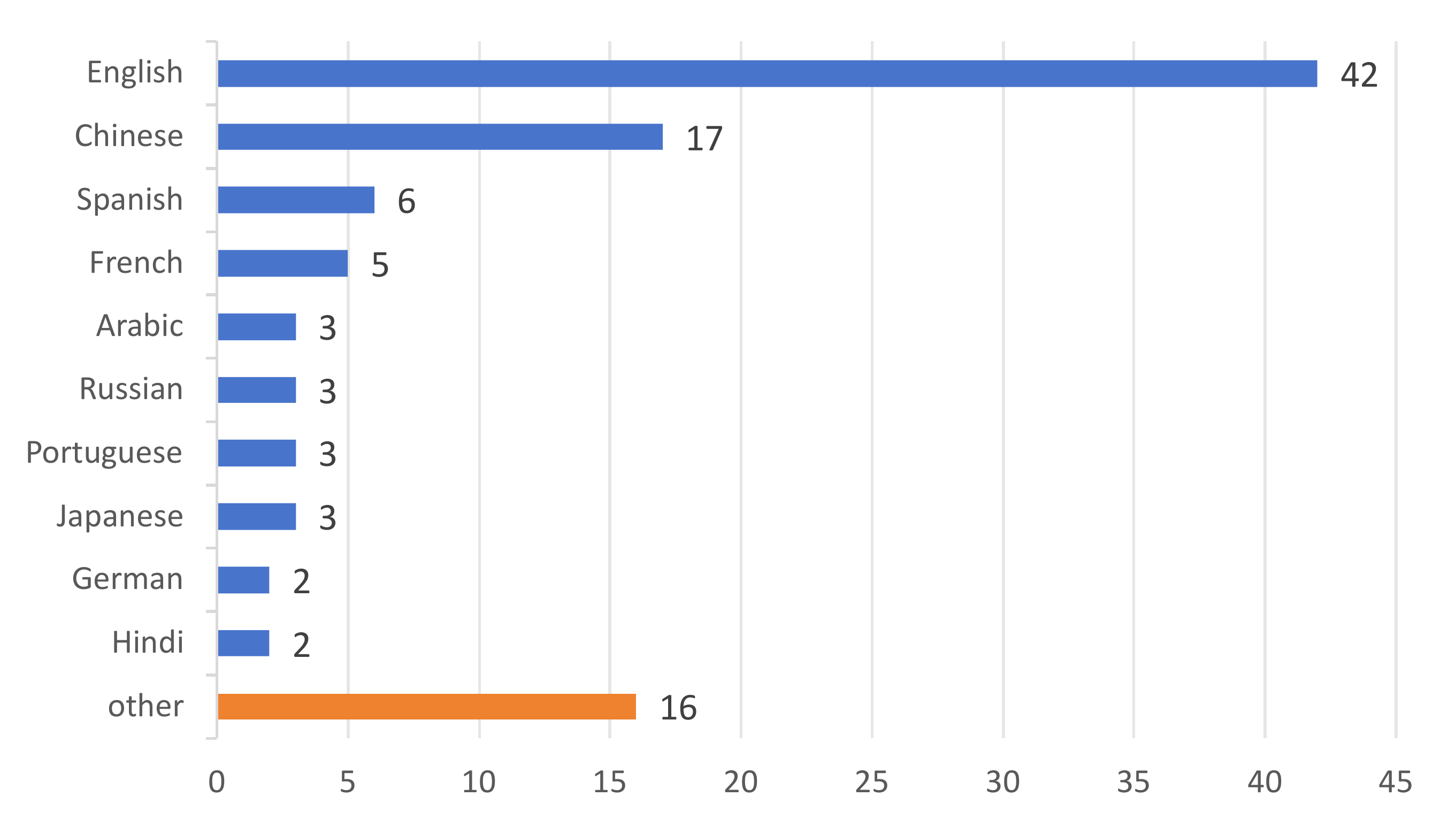}
\caption{Natural Language Distribution of the Benchmarks}
\label{fig9}
\end{figure}

Figure \ref{fig9} presents the distribution of natural language usage across the 56 collected benchmarks. It is evident that English dominates, being used in 45 of the benchmarks (approximately 80\%), establishing itself as the primary language for evaluating medical LLMs. Chinese ranks second, appearing in 17 benchmarks (approximately 30\%), indicating a notable level of development in Chinese medical AI. In contrast, other languages such as Arabic, Russian, Portuguese, Japanese, and Hindi—despite being spoken by large global populations—remain underrepresented in publicly available benchmarks, with each language appearing in only 2 to 6 benchmarks on average. This highlights that multilingual medical AI remains in its early stages.

Finally, it may be concluded that current medical LLM benchmarks exhibit a pronounced trend of language centralization. English, as the de facto language of international medical communication, dominates the vast majority of evaluation tasks. Although multilingual benchmarks are still in their infancy, they hold substantial potential in promoting global medical equity and cross-cultural adaptation. Moving forward, greater efforts are needed to improve the availability of multilingual data resources and to develop cross-lingual evaluation strategies.

\subsection{Medical Diversity}
\begin{figure}[htb]
\centering
\includegraphics[width=0.5\textwidth]{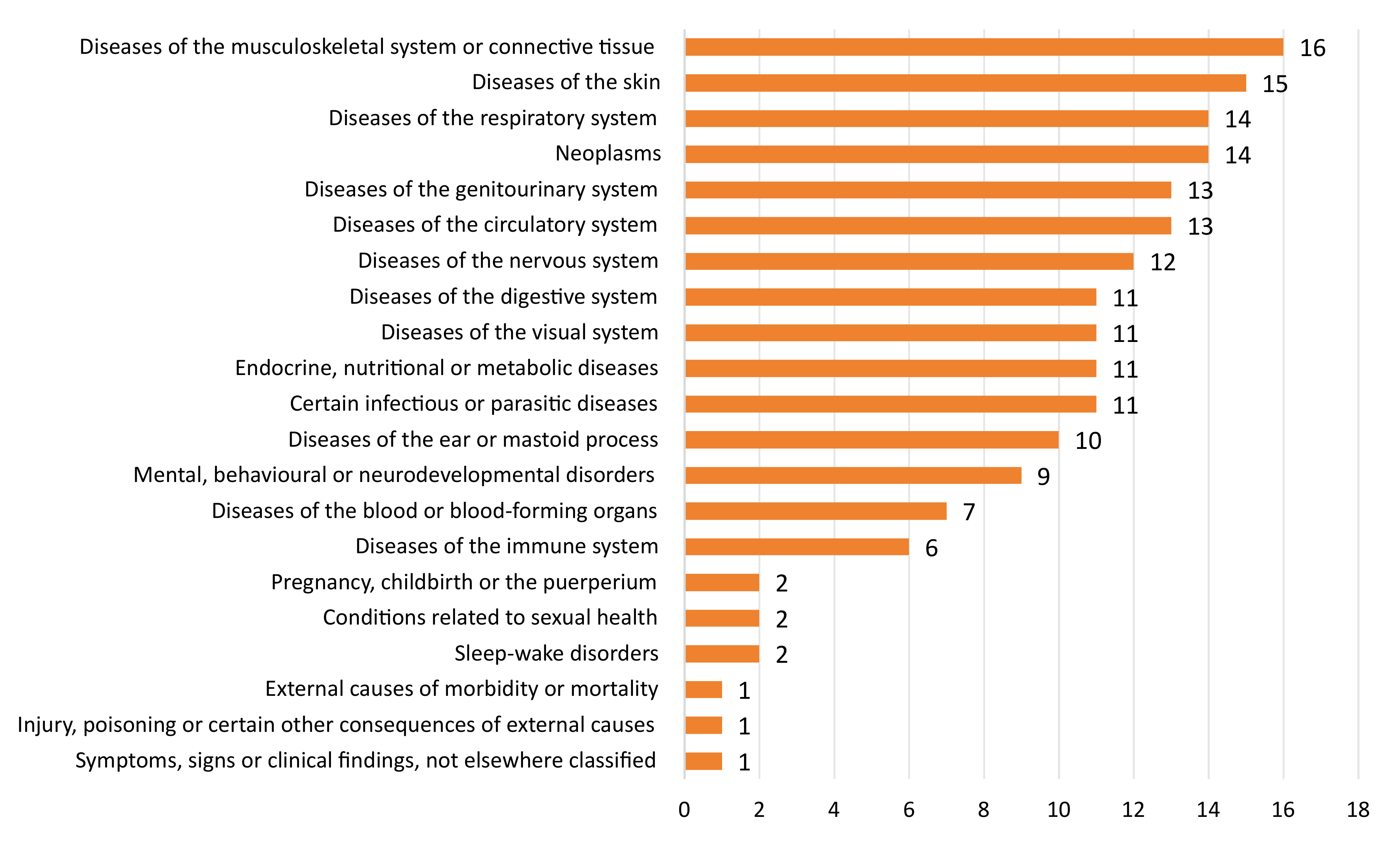}
\caption{Disease Distribution Based on ICD-11 Classification}
\label{fig10}
\end{figure}

\textbf{Distribution of Diseases.} Figure \ref{fig10} illustrates the distribution of disease categories represented in the 56 medical benchmarks, categorized according to the first 23 chapters of the ICD-11 classification system (with Certain conditions originating in the perinatal period and Developmental anomalies excluded due to the absence of relevant benchmarks). The analysis reveals that current research on medical LLMs is heavily concentrated in disease areas such as Diseases of the musculoskeletal system or connective tissue (16 benchmarks), Diseases of the skin (15), Diseases of the respiratory system (14), and Neoplasms (13). These categories typically have abundant publicly available imaging data (e.g., X-rays, CT scans, pathology slides), which facilitates the development of standardized evaluation protocols based on both image and text modalities.

In contrast, low-frequency disease categories such as Mental and behavioral disorders (e.g., depression, anxiety) rely on psychological scales, clinical interviews, or life history data, which are difficult to standardize and rarely captured in benchmark datasets. Similarly, Immunological and Hematological diseases often require the integration of laboratory indicators, clinical progression, and imaging, making them difficult to model using unimodal approaches. From a clinical perspective, however, these underrepresented diseases are not necessarily rare in real-world settings. Many exhibit high outpatient visit rates, cause significant patient distress, or carry substantial social burden. This highlights a misalignment between research frequency and clinical demand. Future research in medical LLMs should be designed with greater awareness of this gap, ensuring that high-demand but rarely covered disease populations are not systematically neglected.

\begin{figure}[htb]
\centering
\includegraphics[width=0.5\textwidth]{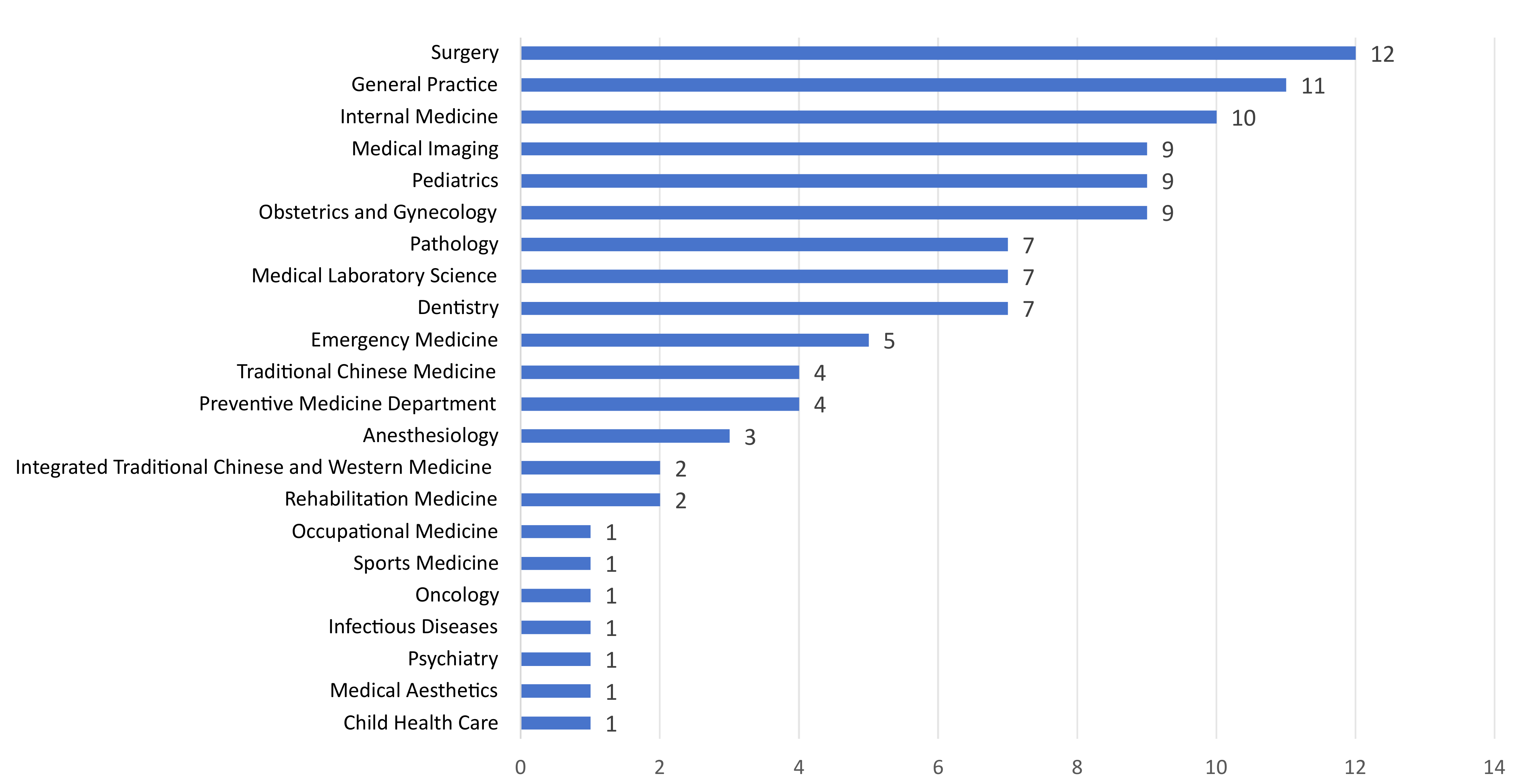}
\caption{Medical Specialty Distribution Based on the Official Directory of Clinical Departments}
\label{fig11}
\end{figure}

\textbf{Medical Specialties.} While ICD-11 provides a useful framework for disease-based diversity analysis, it lacks granularity in distinguishing among interdisciplinary or population-specific fields such as clinical medicine, obstetrics and gynecology, and pediatrics. To complement ICD-11-based analysis, we additionally introduce a second diversity metric based on the official list of clinical departments issued by the National Health Commission of China. Figure \ref{fig11} shows the distribution of benchmarks across these medical specialties (excluding departments such as Endemic Diseases that are not represented in any benchmark or that overlap with ICD-11 categories).

This classification reveals a similar pattern: departments with well-structured data and clearly defined tasks dominate the landscape, such as Surgery (12 benchmarks), General Practice (11), and Medical Imaging (9). In contrast, specialties associated with vulnerable populations or complex care—such as Rehabilitation Medicine, Occupational Health, and Child Health Care—are significantly underrepresented.

This imbalance may result in limited generalizability and fairness in real-world deployment of medical LLMs, particularly in addressing the needs of underserved or structurally marginalized groups. Therefore, future benchmark design should aim for more balanced coverage across medical specialties and actively promote the standardization and open sharing of data from underrepresented departments. This will support the development of AI systems that are not only technically robust but also clinically inclusive.

\subsection{Benchmark Performance Heatmap Example}
\label{sec:appendix_heatmap}

\begin{figure}[!h]
    \centering
    \includegraphics[width=\columnwidth]{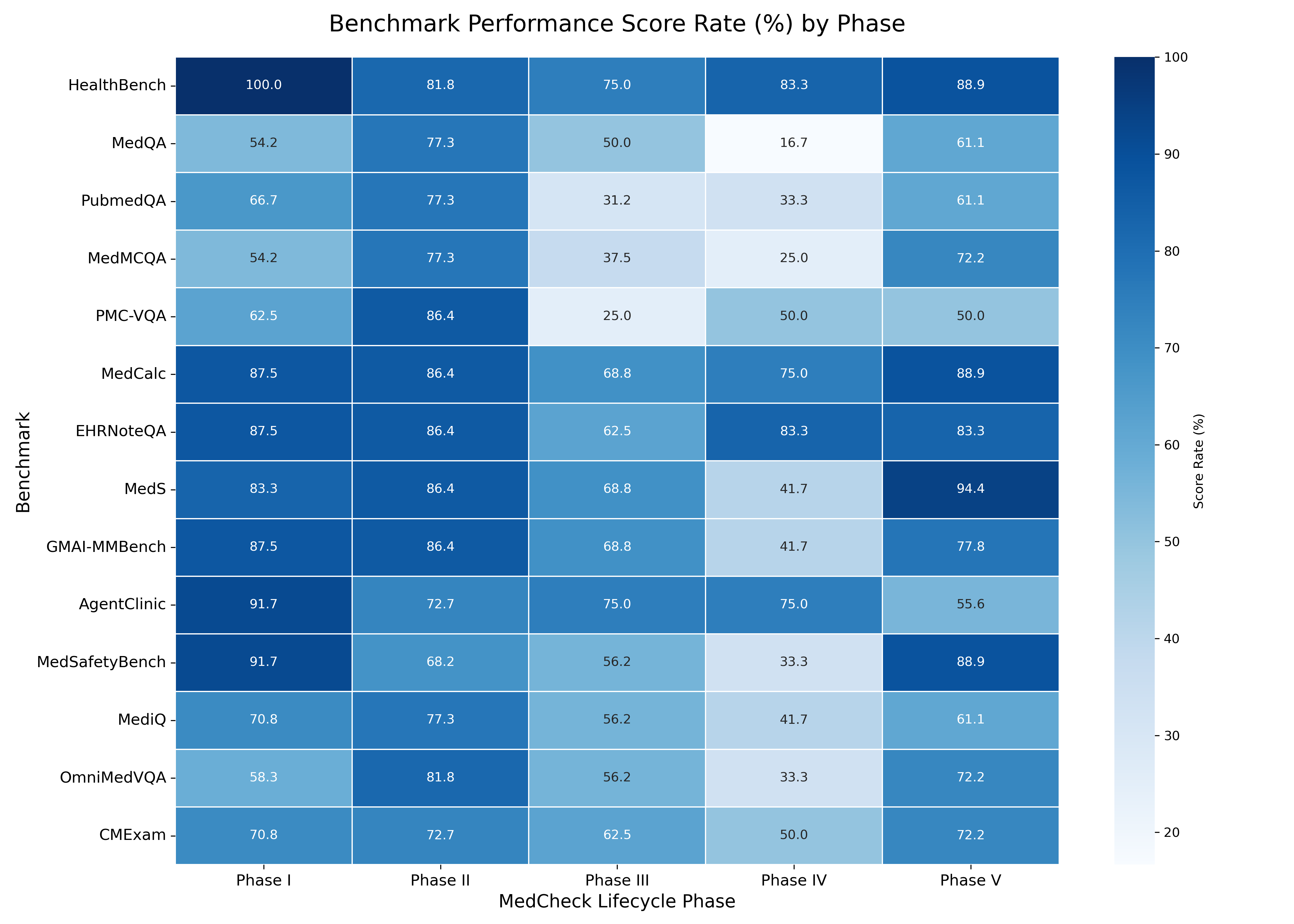}
    
    \caption{Performance score rates (\%) for 14 medical LLM benchmarks across the five phases of the MedCheck evaluation framework.}
    
    \label{fig:heatmap}
\end{figure}

The heatmap in Figure~\ref{fig:heatmap} visualizes the performance of 14 out of 56 of the medical LLM benchmarks evaluated evaluated against the five-phase \textit{MedCheck} framework. Each row represents a benchmark, and columns correspond to the framework's phases. The color intensity reflects the score rate (\%), highlighting the comparative strengths and weaknesses of each benchmark and revealing systemic trends.

\subsection{Statistics About Phase I: Design and Conceptualization}

This section presents a systematic evaluation of the design logic and conceptual clarity of existing medical benchmarks.  we examine their alignment with medical capabilities, practical application scenarios, and innovation objectives. The assessment is structured around three core aspects: (1) the clarity of domain positioning from the perspective of AI capabilities and clinical scope, (2) the methodological soundness in terms of metric design and evaluation diversity, and (3) the integration of safety and fairness considerations as part of responsible AI practices.

\begin{figure}[htb]
\centering
\includegraphics[width=0.5\textwidth]{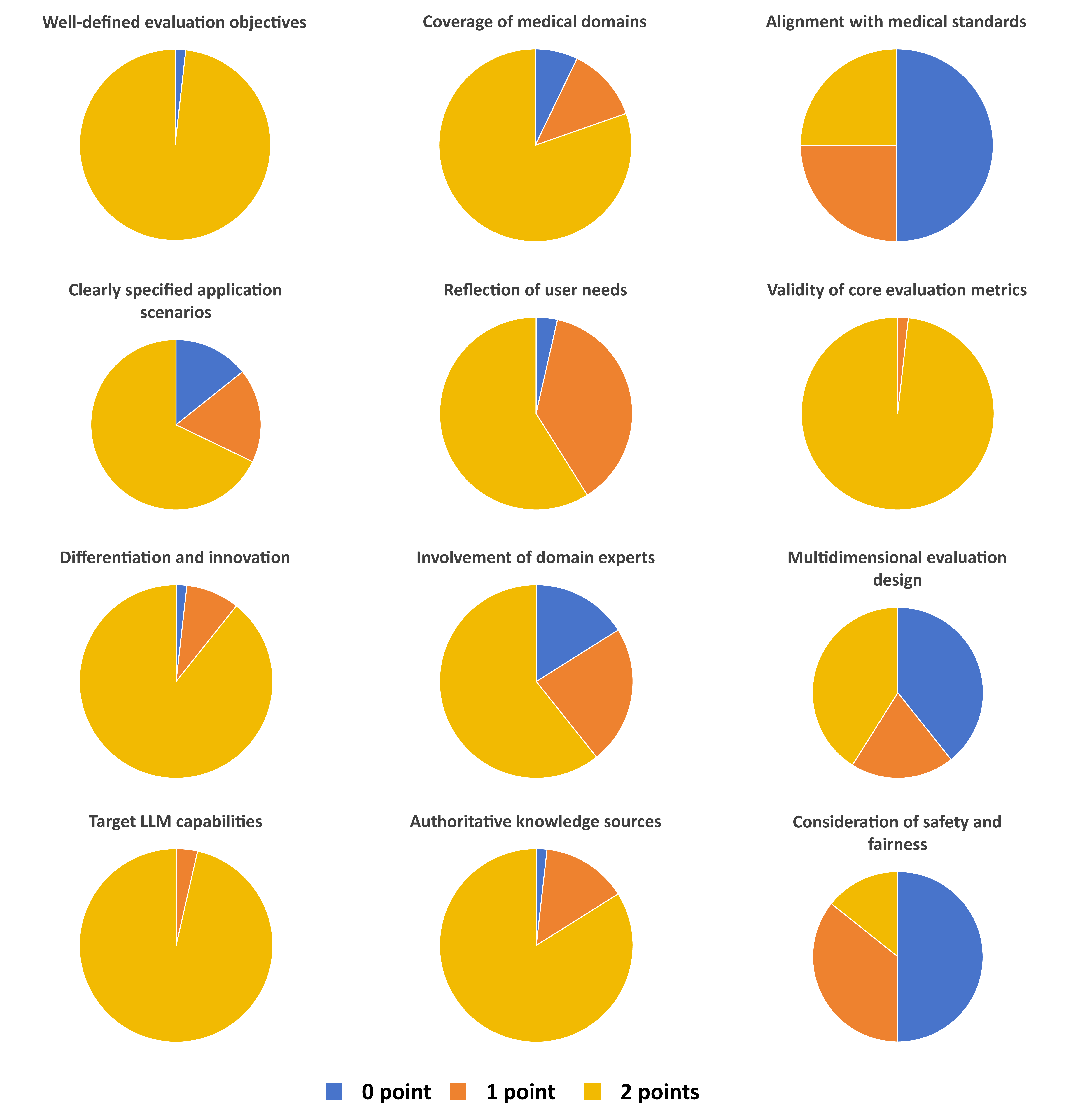}
\caption{Scoring Performance of the 56 Benchmarks in the Design and Conceptualization Stage}
\label{fig1}
\end{figure}

As shown in Figure \ref{fig1}, the average normalized score across benchmarks is 74.5\% (17.9 out of 24), with a mean of 1.5 points per criterion, indicating general compliance with baseline expectations. However, performance on several key items remains suboptimal and warrants further attention.

\begin{figure}[htb]
\centering
\includegraphics[width=0.5\textwidth]{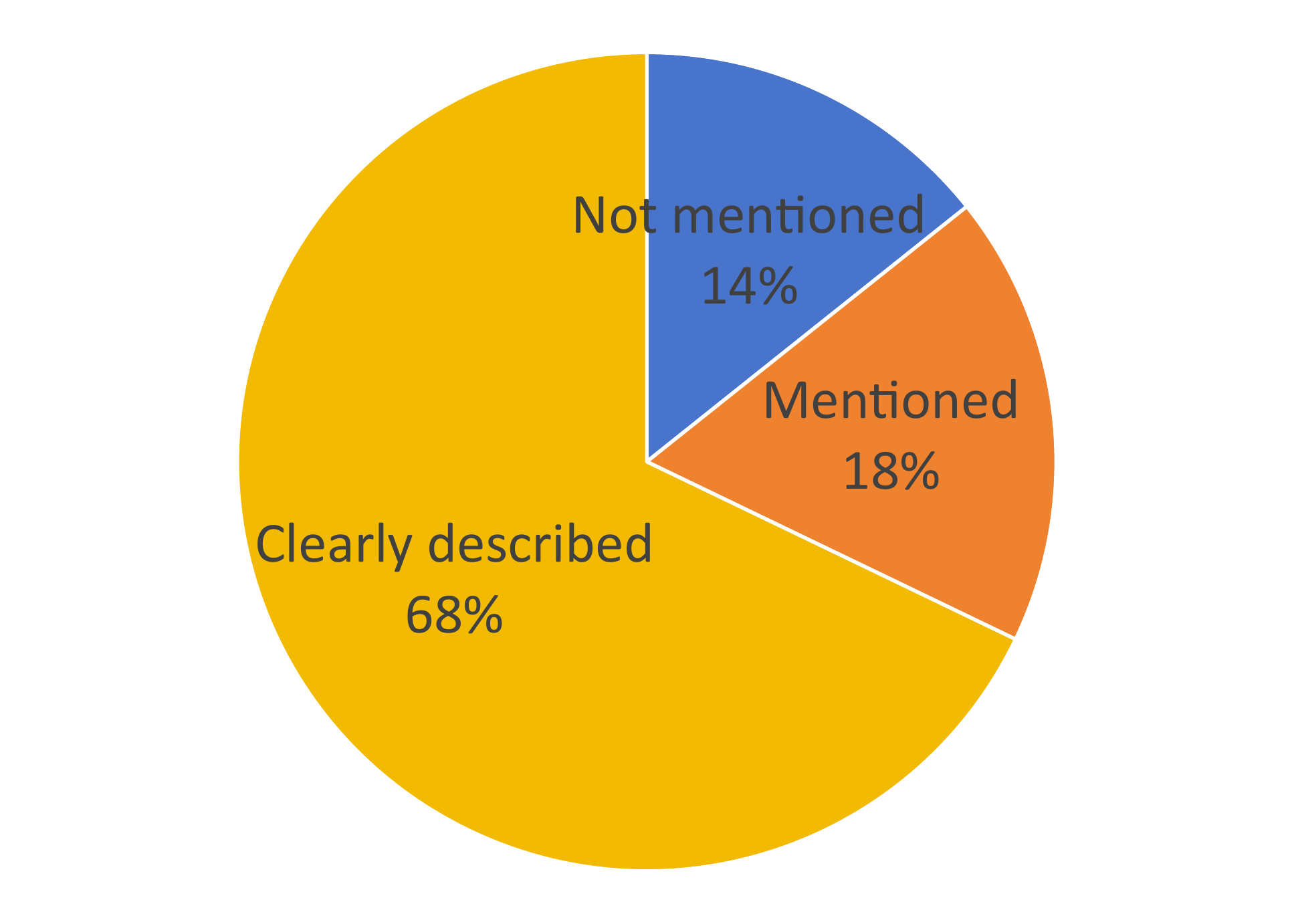}
\caption{Performance on Clarity of Application Scenario}
\label{fig13}
\end{figure}

\textbf{Clearly specified application scenarios.} To ensure practical relevance, benchmarks should clearly define their intended clinical or biomedical research scenarios. Such grounding facilitates real-world applicability and aligns evaluation with deployment needs. As illustrated in Figure \ref{fig13}, 68\% of benchmarks specify concrete application scenarios, articulating clinical value and expected utility. In contrast, 18\% provide only general references to decision-support contexts, and 14\% fail to specify any scenario, reducing interpretability and downstream usability.

\begin{figure}[htb]
\centering
\includegraphics[width=0.5\textwidth]{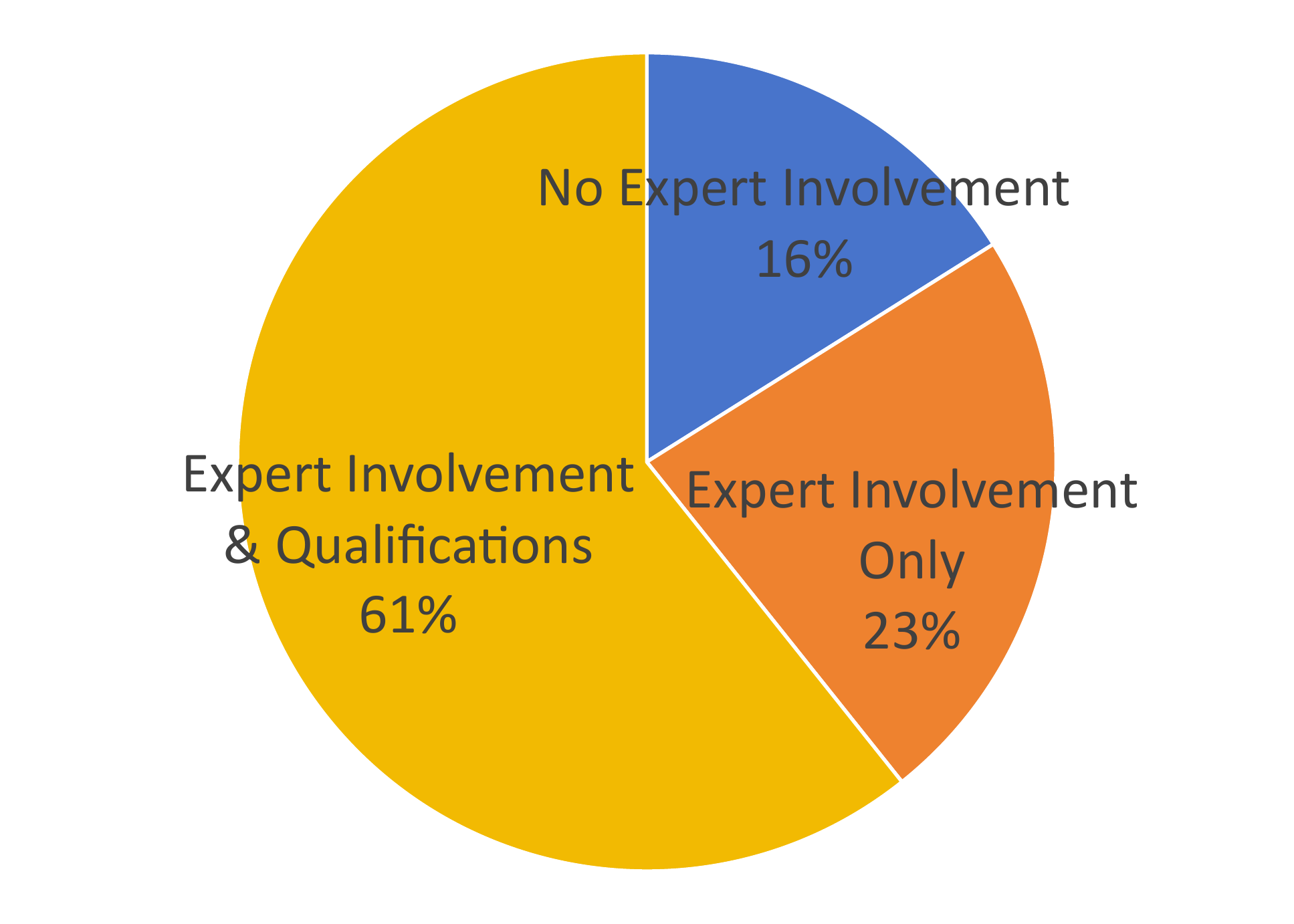}
\caption{Performance on  Domain Experts Involvement}
\label{fig14}
\end{figure}

\textbf{Involvement of domain experts.} Expert input is a critical component in medical AI development. It ensures the appropriateness of task design, data annotation quality, and answer validity. However, as shown in Figure \ref{fig14}, only 23\% of benchmarks explicitly report domain expert involvement, often without detailing the roles or qualifications. 16\% do not mention expert participation, raising concerns about the benchmarks’ clinical validity. Future benchmark design should adopt standardized expert reporting practices to enhance credibility, transparency, and methodological rigor.

\begin{figure}[htb]
\centering
\includegraphics[width=0.5\textwidth]{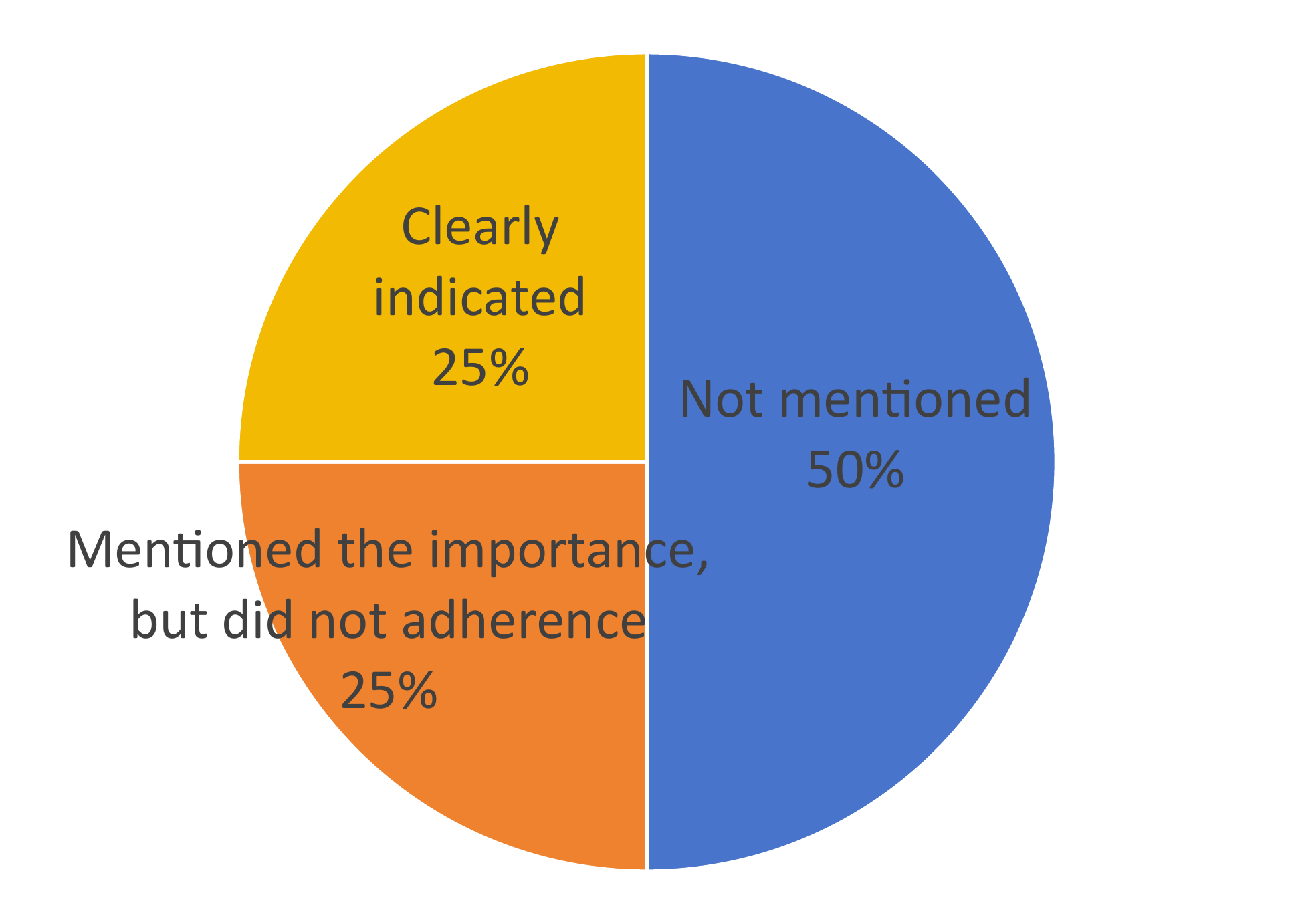}
\caption{Performance on  Medical Standards Alignment}
\label{fig15}
\end{figure}

\textbf{Alignment with medical standards.} Adherence to international medical standards (e.g., ICD-11, SNOMED CT, LOINC) is essential for achieving semantic interoperability, annotation consistency, and cross-system comparability. As Figure \ref{fig15} indicates, 50\% of benchmarks do not clarify whether their data conforms to such standards. This omission undermines the benchmark’s reusability and weakens the comparability of evaluation outcomes. Without standard-based structure and terminology, label ambiguity and inconsistent categorization may arise, which compromises the reliability of model performance evaluation.

\begin{figure}[htb]
\centering
\includegraphics[width=0.5\textwidth]{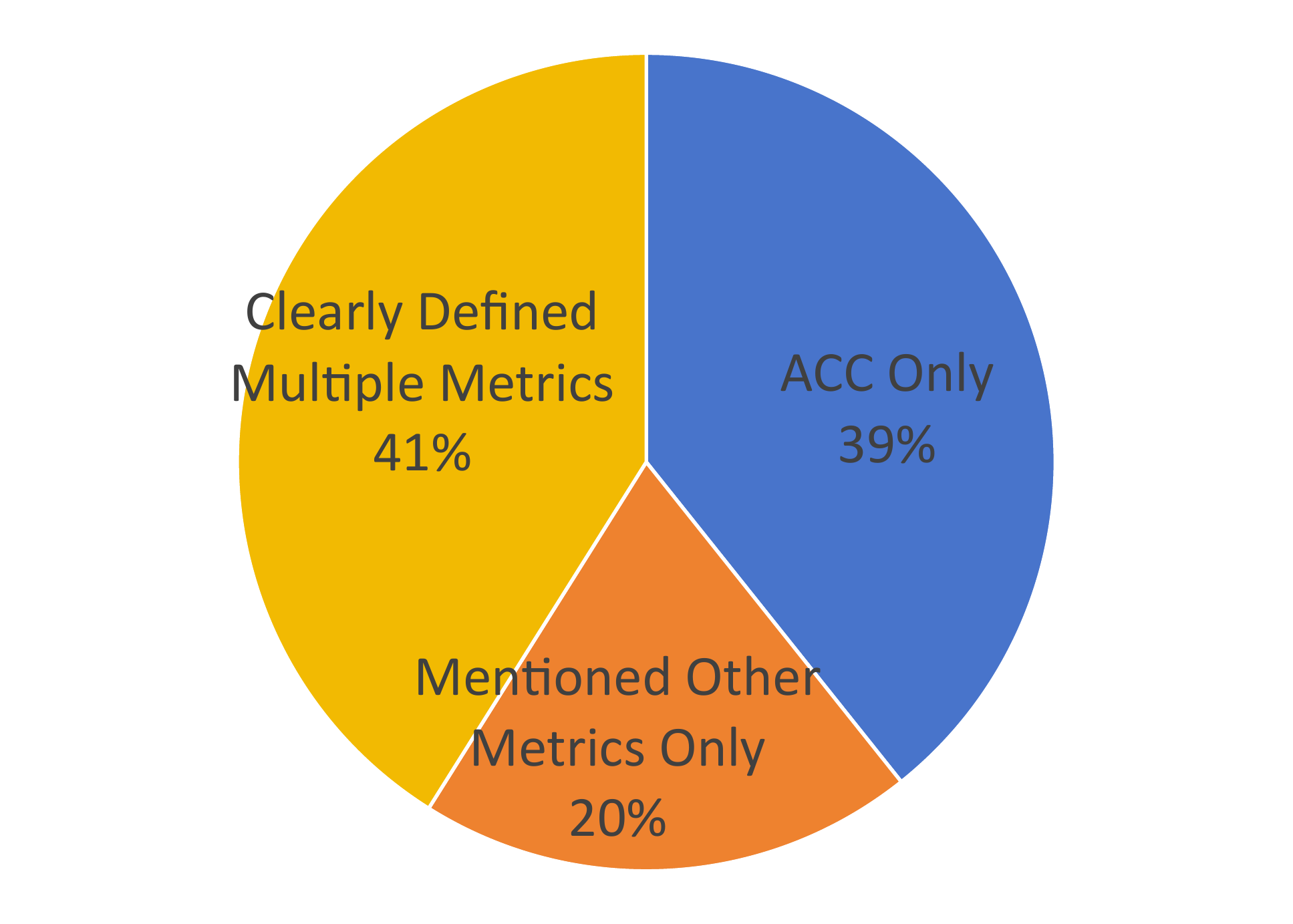}
\caption{Performance on Multi-dimensional Evaluation}
\label{fig16}
\end{figure}

\textbf{Multidimensional evaluation design.} In medical AI, a model's performance cannot be solely measured by accuracy. Outputs that are correct but unsafe, incomplete, or biased may have detrimental consequences. Multidimensional evaluation frameworks are required to assess whether models' output is not only correct but also reliable. As presented in Figure \ref{fig16}, 39\% of benchmarks use only a single metric, typically accuracy, and 20\% mention other metrics without detailed implementation. Metrics such as uncertainty estimation, answer refusal, and harmful output detection are absent from most benchmarks, despite their relevance in high-stakes medical decision-making.

\begin{figure}[htb]
\centering
\includegraphics[width=0.5\textwidth]{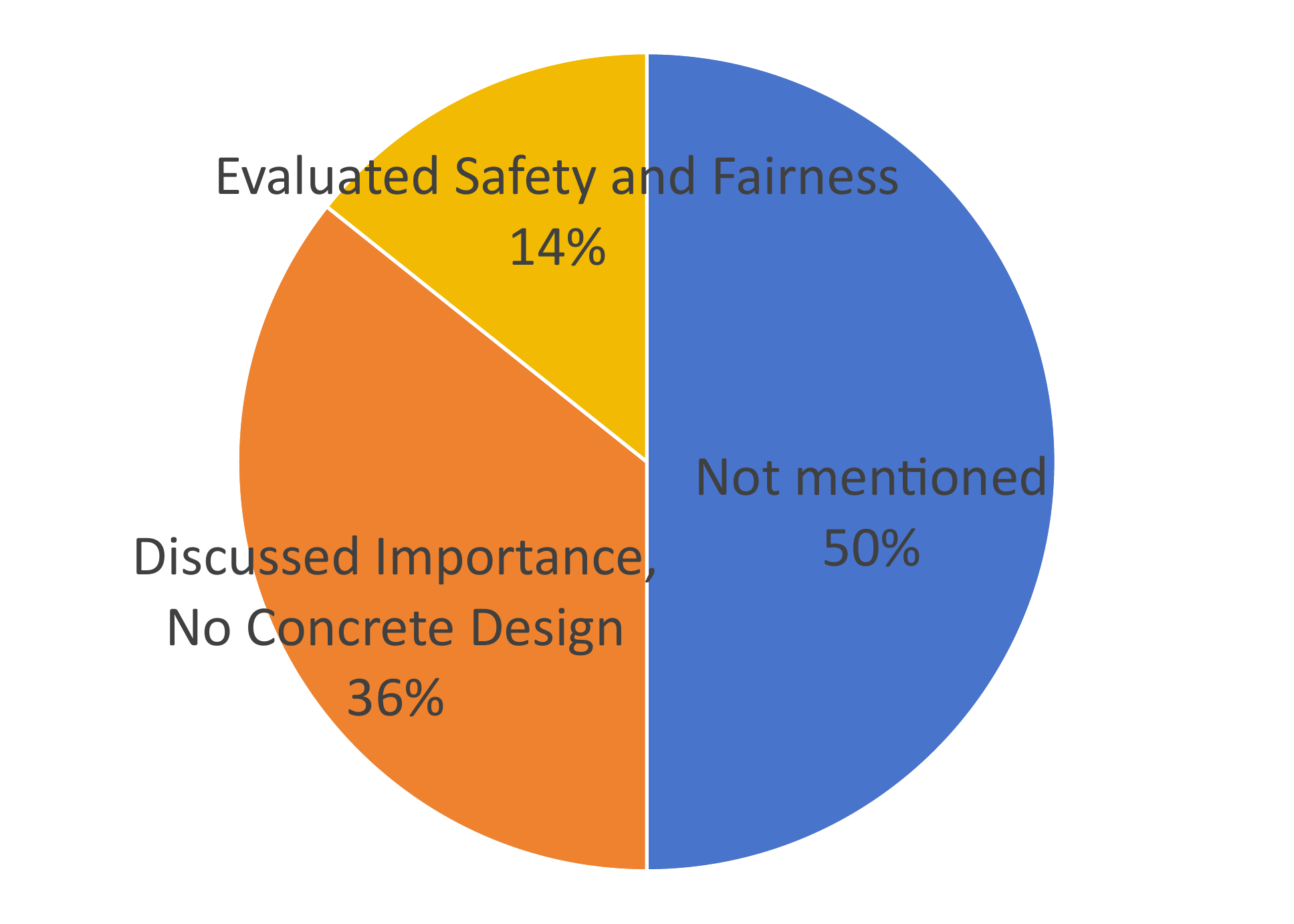}
\caption{Performance on Consideration of Safety and Fairness}
\label{fig17}
\end{figure}

\textbf{Consideration of safety and fairness}. Robust evaluation frameworks must account for safety and fairness, especially in sensitive domains like healthcare. As shown in Figure \ref{fig17}, only 14\% of benchmarks include explicit safety or fairness evaluations, while 36\% acknowledge their importance without implementation. The remaining 50\% omit these aspects entirely. Neglecting safety assessments increases the risk of misleading or overconfident outputs, which may harm clinical decisions. The lack of fairness evaluation may propagate systematic bias across gender, age, race, or disease distribution, exacerbating healthcare disparities. To build trustworthy and ethical medical AI, safety and fairness must be treated as core evaluation dimensions.

\subsection{Statistics About Phase II: Dataset Construction and Management}
\begin{figure}[htb]
\centering
\includegraphics[width=0.5\textwidth]{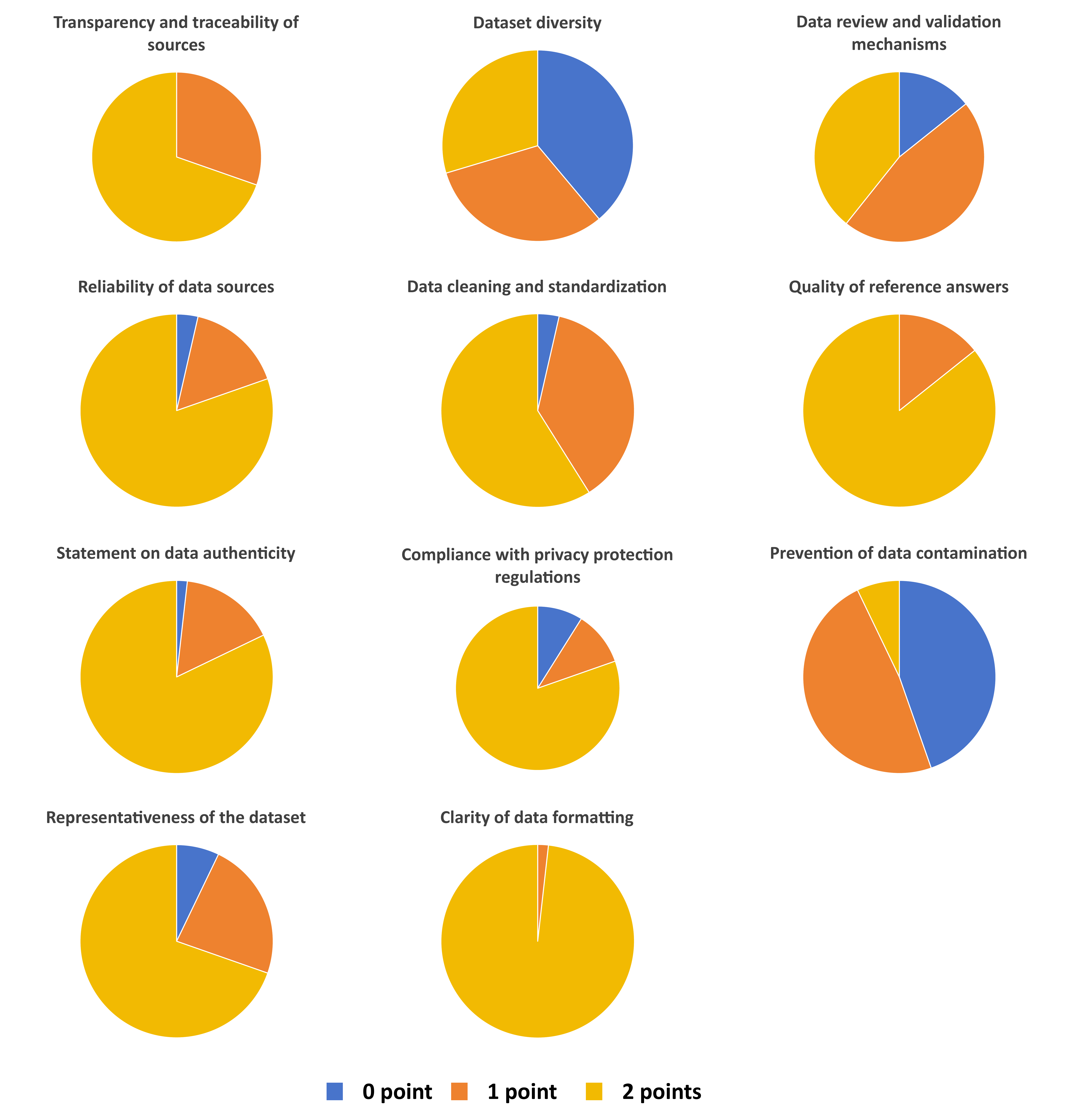}
\caption{Scoring Performance of the 56 Benchmarks in the Phase II: Dataset Construction and Management}
\label{fig2}
\end{figure}

This phase constitutes the empirical foundation of the benchmark, with an emphasis on constructing datasets that are realistic, diverse, representative, and ethically sourced. Key elements include stringent procedures for data quality assurance, privacy protection, and contamination prevention. The dataset sources and quality were assessed across five dimensions: transparency and traceability of data provenance, reliability of data sources, and data authenticity. Data processing and privacy protection were evaluated in six aspects, including data cleaning and standardization, privacy-preserving mechanisms, and data formatting protocols. As shown in Figure \ref{fig2}, the 56 benchmarks achieved an average score rate of 74.7 \% (16.4 out of 22) across 11 evaluation criteria, with a mean of 1.4 points per item. Overall, the results indicate general compliance with the proposed standards, although two specific criteria showed suboptimal performance.

\textbf{Dataset Diversity.} This criterion evaluates whether the benchmark explicitly defines diversity objectives—such as coverage of disease types and clinical departments—and provides quantitative evidence (e.g., disease distribution histograms, task type ratios) to demonstrate dataset coverage. Ensuring sufficient diversity is critical to evaluate model generalization across a broad range of cases and specialties, thereby mitigating evaluation bias.

The specific calculation is shown in Equation ~\ref{eq1}.
In this Equation, \( N_{\text{disease}} \) represents the number of diseases in the ICD-11 standard, specifically the first 23 diseases, serving as the benchmark for the types of diseases considered. \( N_{\text{department}} \) denotes the number of medical departments, typically referring to the medical specialties included in the model's evaluation. The variable \( N_{\text{disease}}^{\text{benchmark}} \) indicates the number of diseases covered by the benchmark, representing the actual disease categories included in the model’s evaluation. Similarly, \( N_{\text{department}}^{\text{benchmark}} \) refers to the number of medical departments involved in the benchmark. Finally, \( R_{\text{coverage}} \) represents the coverage ratio, which is calculated by comparing the sum of the diseases and medical departments covered by the benchmark \( N_{\text{disease}}^{\text{benchmark}} + N_{\text{department}}^{\text{benchmark}} \) to the total number of diseases and medical departments in the reference standard \( N_{\text{disease}} + N_{\text{department}} \). This formula provides a quantitative measure of the benchmark’s coverage across various diseases and medical departments, reflecting the model's diversity and comprehensiveness in its evaluation.

\begin{figure}[htb]
\centering
\includegraphics[width=0.5\textwidth]{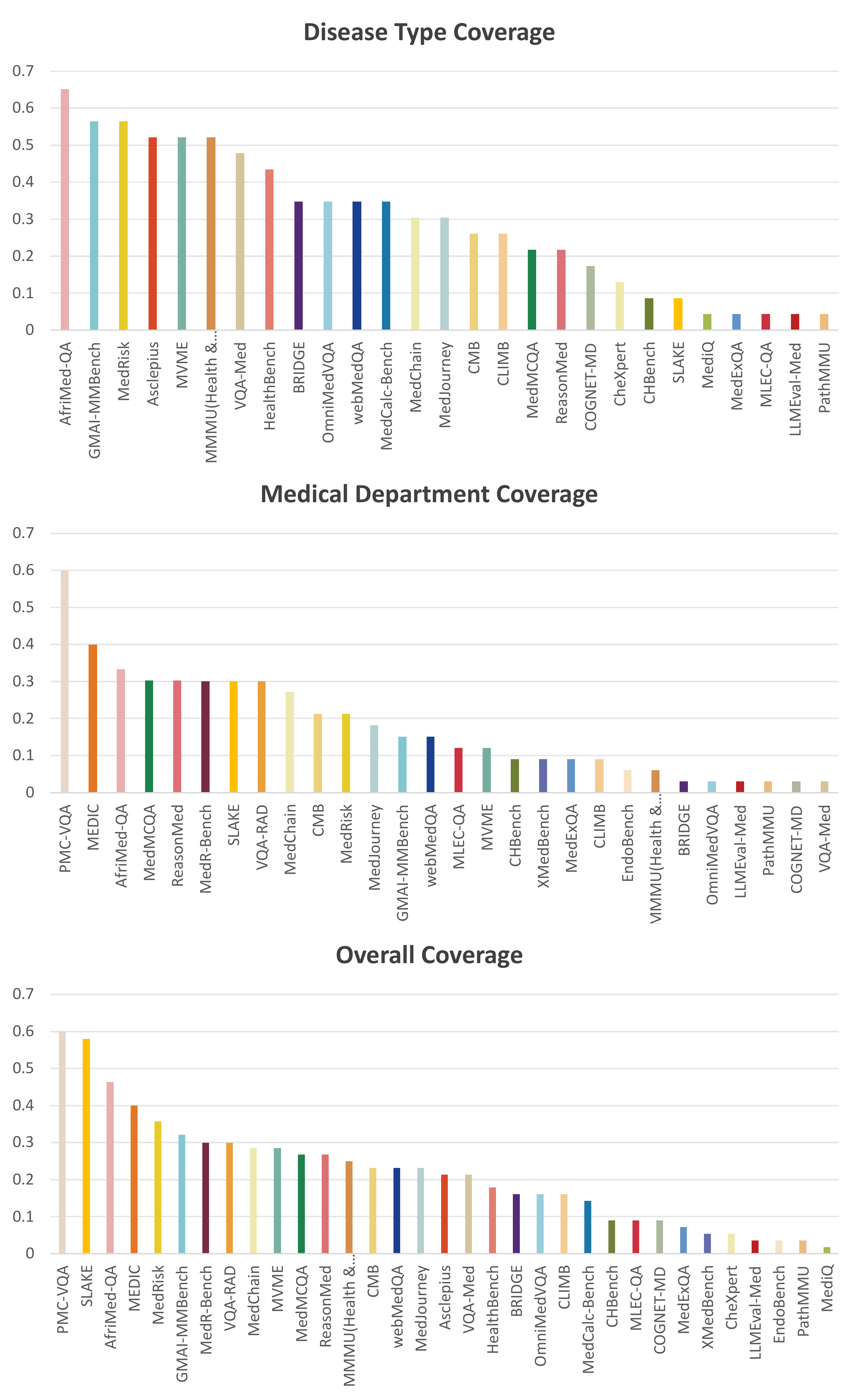}
\caption{Coverage of disease types, medical departments, and overall performance across 56 medical benchmarks.}
\label{fig30}
\end{figure}

\begin{figure}[htb]
\centering
\includegraphics[width=0.5\textwidth]{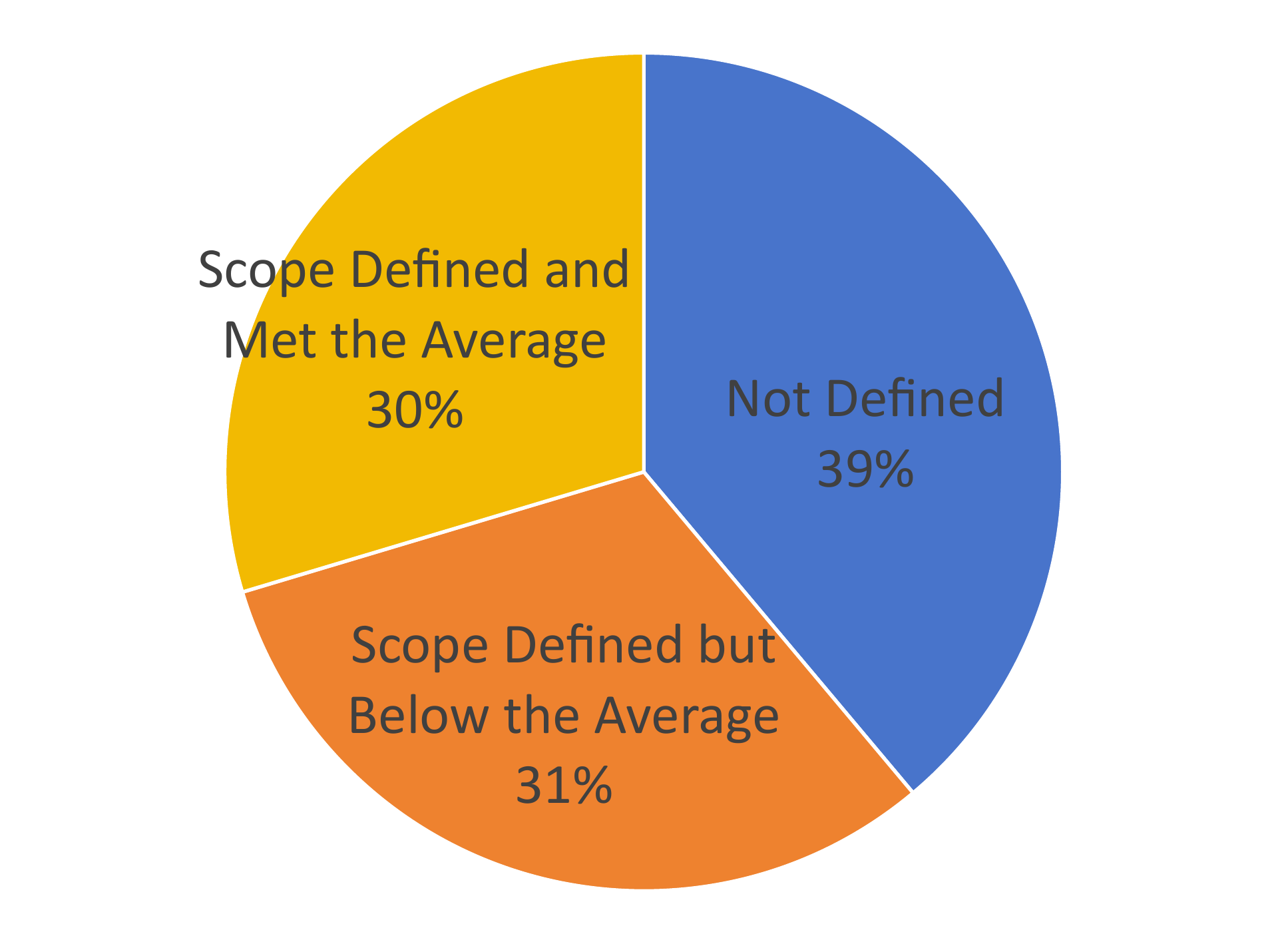}
\caption{Performance on Dataset diversity}
\label{fig18}
\end{figure}

The coverage of each benchmark after calculation is shown in Fig. \ref{fig30}. We scored each benchmark based on the average total coverage rate (21.8\%), and the results are presented in Fig \ref{fig18}, 39\% (21 out of 56) of the benchmarks lack a clear definition or discussion of dataset diversity. The absence of well-defined diversity goals and corresponding quantitative analyses hinders the ability to assess model robustness across different disease categories or clinical settings, potentially obscuring vulnerabilities in specific subgroups or task scenarios.

\begin{equation}
  \label{eq1}
    R_{\text{coverage}} = \frac{N_{\text{disease}}^{\text{benchmark}} + N_{\text{department}}^{\text{benchmark}}}{N_{\text{disease}} + N_{\text{department}}}
\end{equation}

\begin{figure}[htb]
\centering
\includegraphics[width=0.5\textwidth]{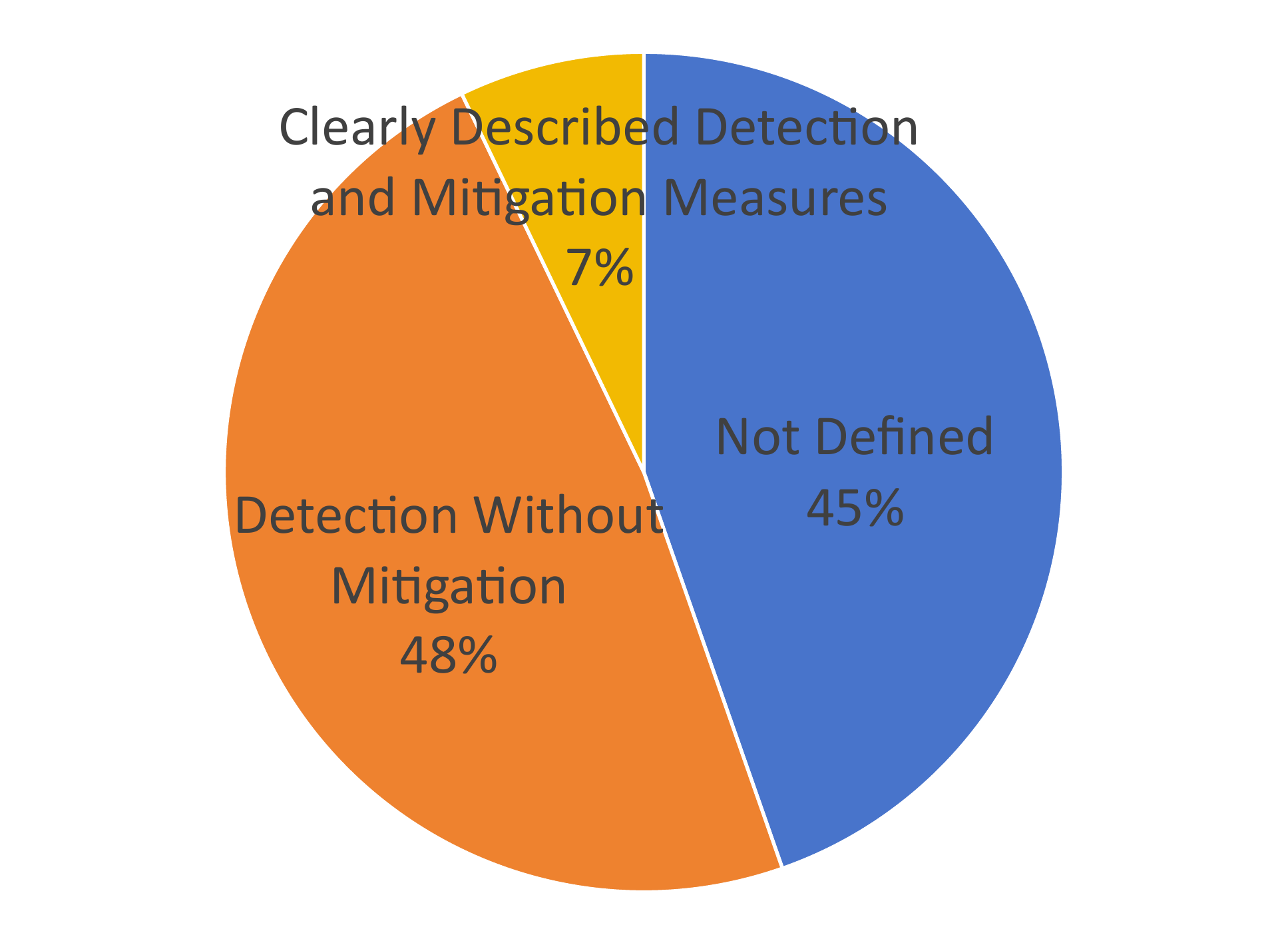}
\caption{Performance on Data Contamination Prevention}
\label{fig19}
\end{figure}

\textbf{Prevention of Data Contamination.} This dimension assesses whether the benchmark includes mechanisms to identify and mitigate potential data contamination—i.e., the presence of evaluation samples within the training corpus of LLMs. Such contamination can artificially inflate model performance through memorization, thereby compromising the fairness and validity of the evaluation. If present in high-risk tasks (e.g., diagnostic decisions or medication recommendations), contamination may mislead researchers and regulators in their assessment of model safety. Therefore, contamination detection is a prerequisite for ensuring the credibility and scientific rigor of benchmark evaluations. However, as depicted in Figure \ref{fig19}, only 7\% of the benchmarks implemented both detection and mitigation strategies. Another 48\% conducted detection without addressing identified contamination, while 45\% did not mention the issue at all.

\subsection{Statistics About Phase III: Technical Implementation and Evaluation Methodology}
\begin{figure}[htb]
\centering
\includegraphics[width=0.5\textwidth]{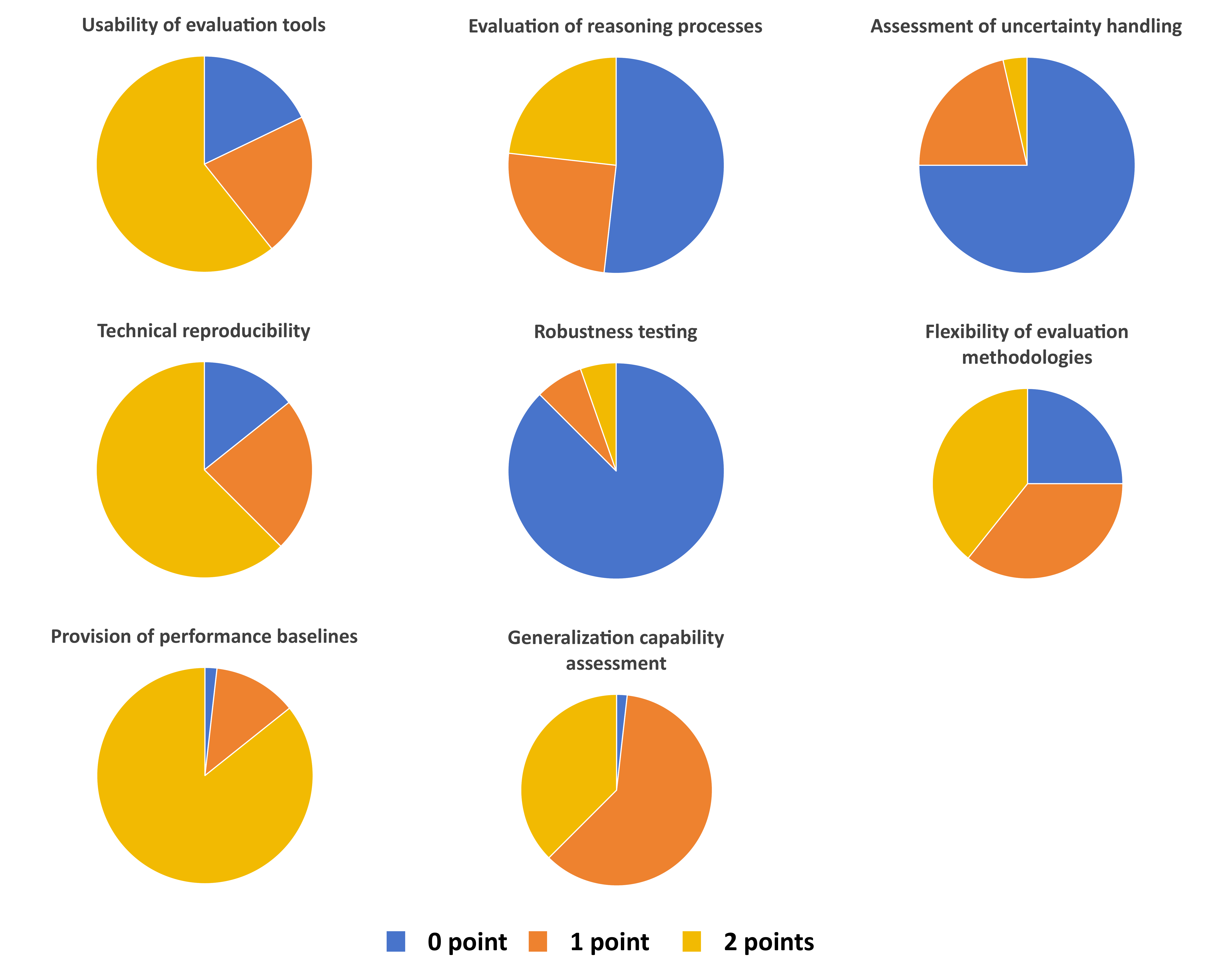}
\caption{Scoring Performance of the 56 Benchmarks in the Phase III: Technical Implementation and Evaluation Methodology}
\label{fig3}
\end{figure}

This is the operational backbone. It consists of 8 items, encompassing the development of accessible, reproducible evaluation scripts and the selection of metrics that go beyond simple accuracy to assess deeper capabilities such as reasoning, robustness, and uncertainty awareness. For the 56 evaluated benchmarks, the average score is 52.1\% (8.33 out of 16), with a mean score of 1.04 for each item. Results indicate that this is the second worst performing stage among the 5 stages. As shown in Figure \ref{fig3}, there are 3 items exhibiting critical underperformance, where majority of the benchmarks receive 0 mark.

\textbf{Reasoning Process Evaluation.} This criterion assesses whether the benchmarks include evaluation for the reasoning process of the models. Apart from the final answer, understanding the decision-making process of models equally important. 

\begin{figure}[htb]
\centering
\includegraphics[width=0.5\textwidth]{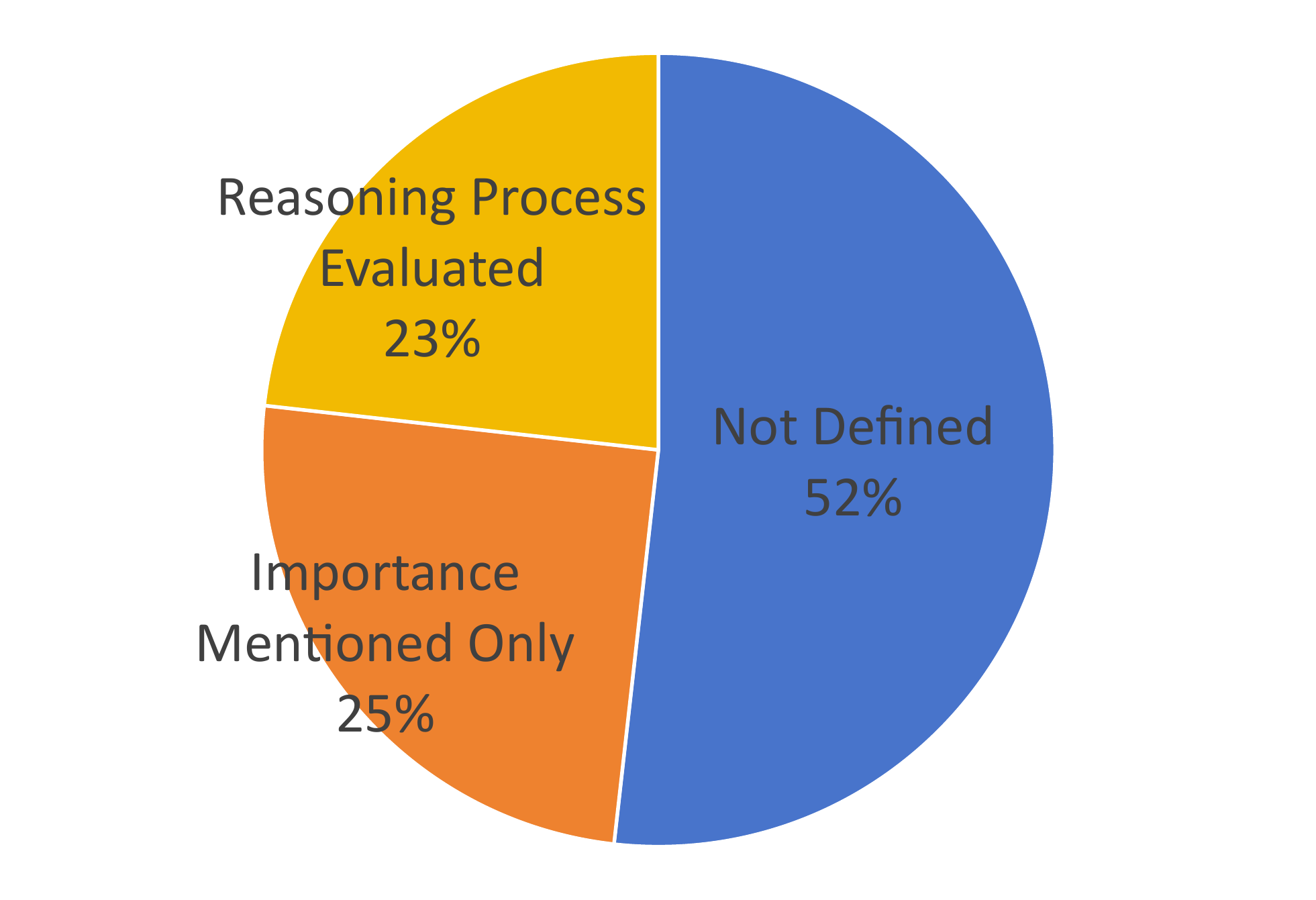}
\caption{Performance on Reasoning Process Evaluation}
\label{fig20}
\end{figure}

As shown in Figure \ref{fig20}, a concerning 55\% of benchmarks do not have any consideration for the models' reasoning process. While 25\% of benchmarks mention the importance of evaluating the reasoning process, only 23\% of benchmarks design concrete  assessment for reasoning process with clear methods and metrics.

Without evaluating the reasoning process, the flawed logic or hidden biases of models may left unchecked, potentially compromising patient safety. By assessing the reasoning path, how and why a model arrives at its conclusions can be revealed, fostering transparency and trustworthiness.

\textbf{Robustness Evaluation.} This item assesses whether there is evaluation for robustness in the benchmark. In practical application, a robust model should be able to reliably interpret different variations of input. Robustness evaluation help assess whether models are resilient to imperfections and perturbations, ensuring the reliability when adopted in high-stakes medical scenarios.

\begin{figure}[htb]
\centering
\includegraphics[width=0.5\textwidth]{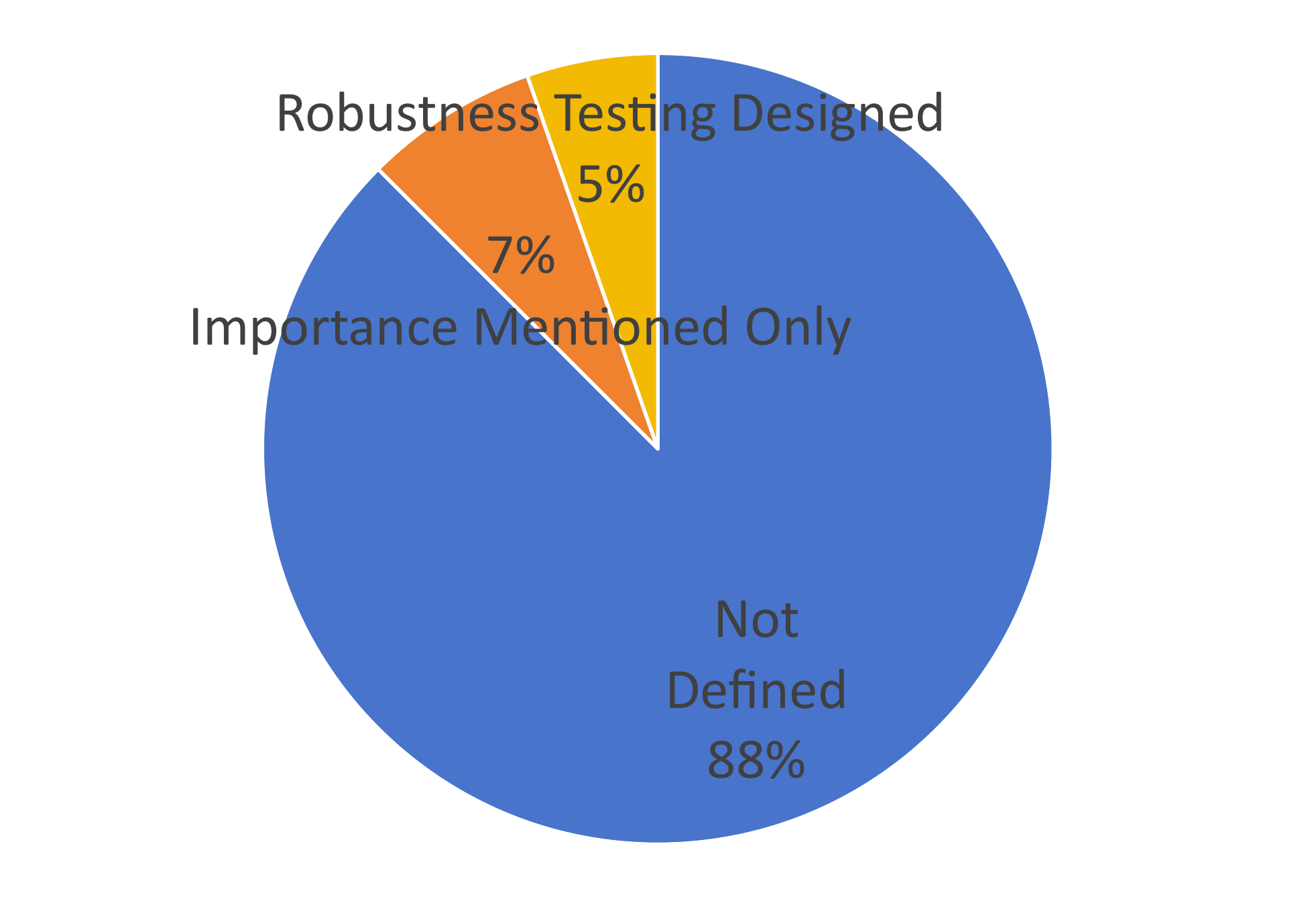}
\caption{Performance on Robustness evaluation}
\label{fig21}
\end{figure}

As depicted in Figure \ref{fig21}, an overwhelming 88\% of the covered benchmarks do not consider robustness in their evaluation. While 4 benchmarks recognize the importance of robustness evaluation, only 3 benchmarks design clear assessment for robustness.

Real-world environments are inherently variable and noisy. While models may perform well under idealized inputs in the benchmark, they may fail when facing subtle changes. Without robustness evaluation, benchmark results could be overly optimistic. This false confidence may lead to misleading decisions, causing risks when deployed in real-world environment. By evaluating robustness, users can gain a better understanding of models' stability and consistency, fostering safer and more dependable deployment in complex medical environments.

\textbf{Uncertainty Evaluation.} This criterion is designed for evaluating the ability of models to recognize and express its uncertainty. In high-stakes medical contexts, it is vital for models to recognize and communicate the limits of their knowledge. Assessing how safely and responsibly a model handles uncertainty promotes caution, reducing the risk of overconfident errors and misleading decisions.

\begin{figure}[htb]
\centering
\includegraphics[width=0.5\textwidth]{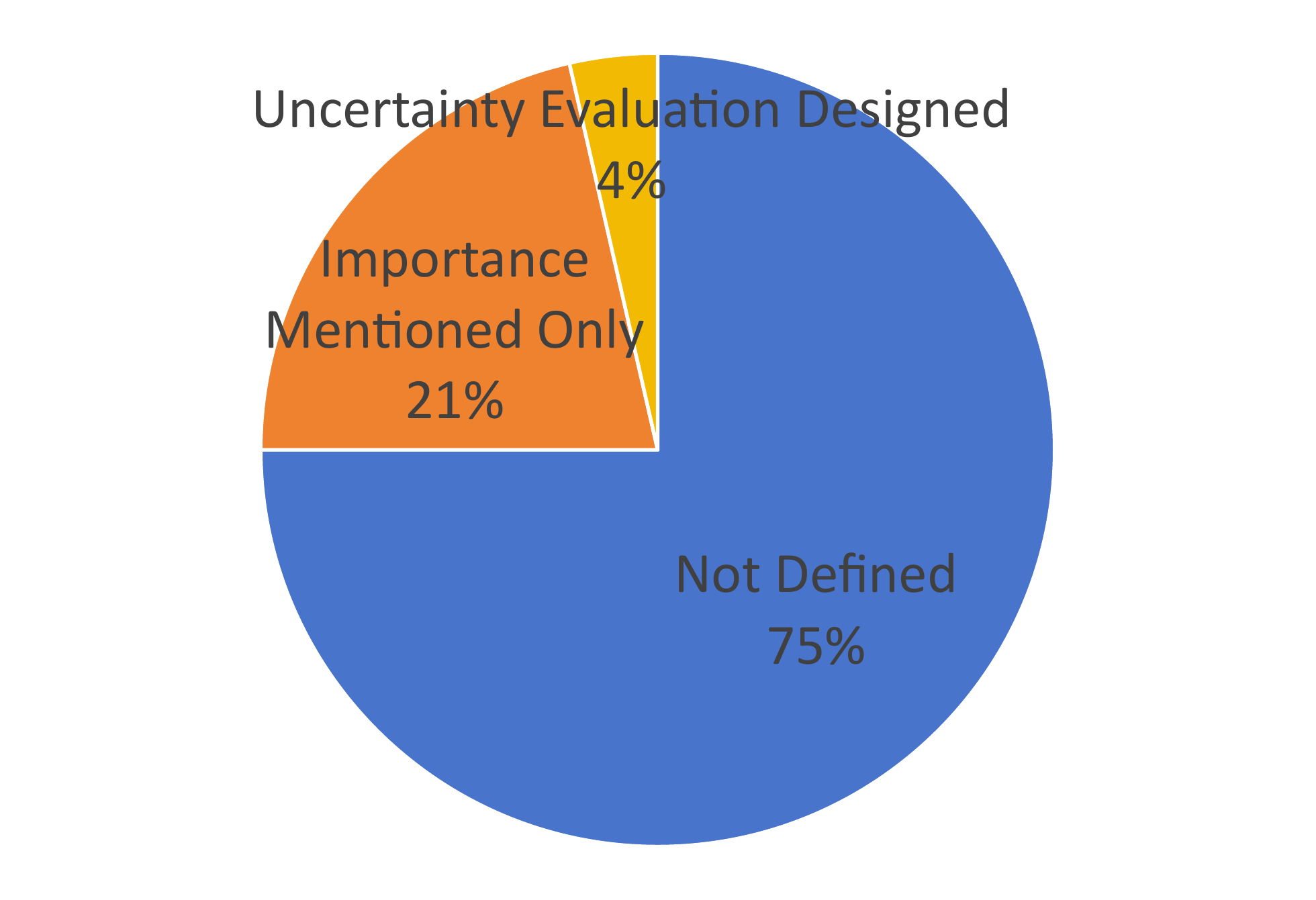}
\caption{Performance on Uncertainty Evaluation}
\label{fig22}
\end{figure}

As illustrated in Figure \ref{fig22}, 75\% benchmarks have no considerations for uncertainty in their evaluation. Even though 21\% acknowledge the importance of uncertainty evaluation, only 4\% incorporate evaluation related to uncertainty in the benchmark, highlighting a significant gap in current benchmarking practices.

If uncertainty evaluation is omitted, it can lead to misleading performance results that overstate a model’s safety and reliability. When facing uncertainty, if models provide overconfident answers, it could lead to harmful outcomes. Benchmarks without uncertainty evaluation fail to reward models that behave ethically and responsibly under uncertainty, undermining the development of safer AI tools in medicine. By evaluating uncertainty, benchmarks can better reflect clinical uncertainty, fostering more trustworthy and safer decision-making behavior

\subsection{Statistics About Phase IV: Benchmark Validity and Performance Verification}
\begin{figure}[htb]
\centering
\includegraphics[width=0.5\textwidth]{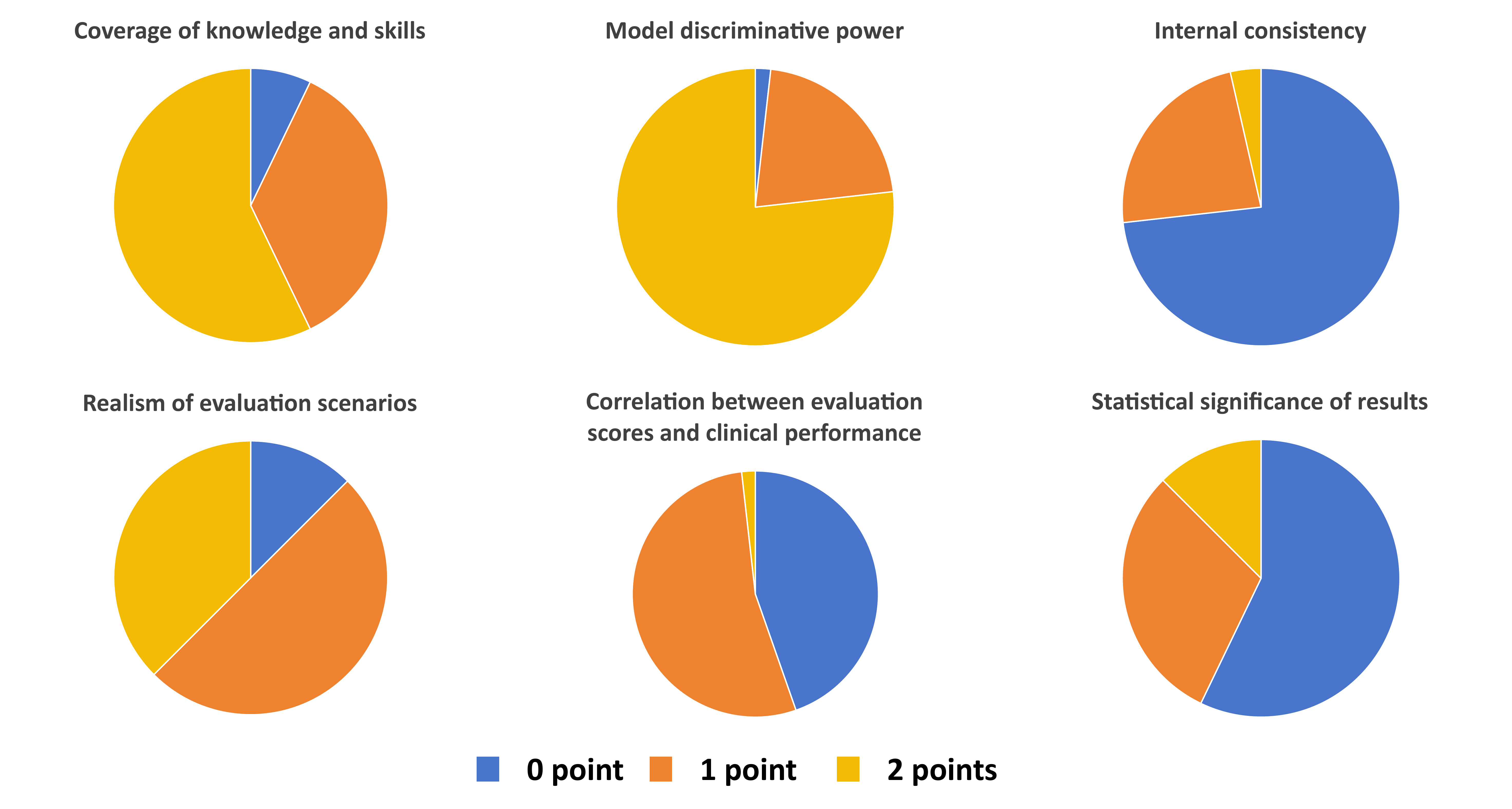}
\caption{Scoring Performance of the 56 Benchmarks in the Phase IV: Benchmark Validity and Performance Verification}
\label{fig4}
\end{figure}

This phase provides scientific validation for the benchmark. It encompasses evidence of content validity (i.e., whether the benchmark comprehensively covers the target domain) and construct validity (i.e., whether it measures the claimed latent capabilities), as well as its ability to distinguish performance differences among models of varying proficiency levels. Among the 56 benchmarks evaluated, the average score rate for this phase was 49.1\% (5.90 out of 12), with a mean of 0.98 points per item—representing the lowest performance among the five phases. As demonstrated in Figure \ref{fig4}, of the six evaluation criteria in this phase, three demonstrated notably poor results.

\textbf{Correlation with Clinical Performance.} This component assesses whether there is empirical evidence exploring the relationship between benchmark scores and model performance in real-world clinical applications. Specifically, it examines whether the benchmark scores can reliably reflect clinical effectiveness—an essential indicator of external validity and practical utility. In medical contexts, the ultimate goal of model performance is to enhance the accuracy, safety, and efficiency of clinical decision-making. If evaluation scores fail to map onto actual clinical outcomes, the benchmark may mislead researchers regarding model applicability. Moreover, neglecting this correlation may incentivize developers to over-optimize non-essential capabilities for better scores, thus deviating from real-world deployment needs.

\begin{figure}[htb]
\centering
\includegraphics[width=0.5\textwidth]{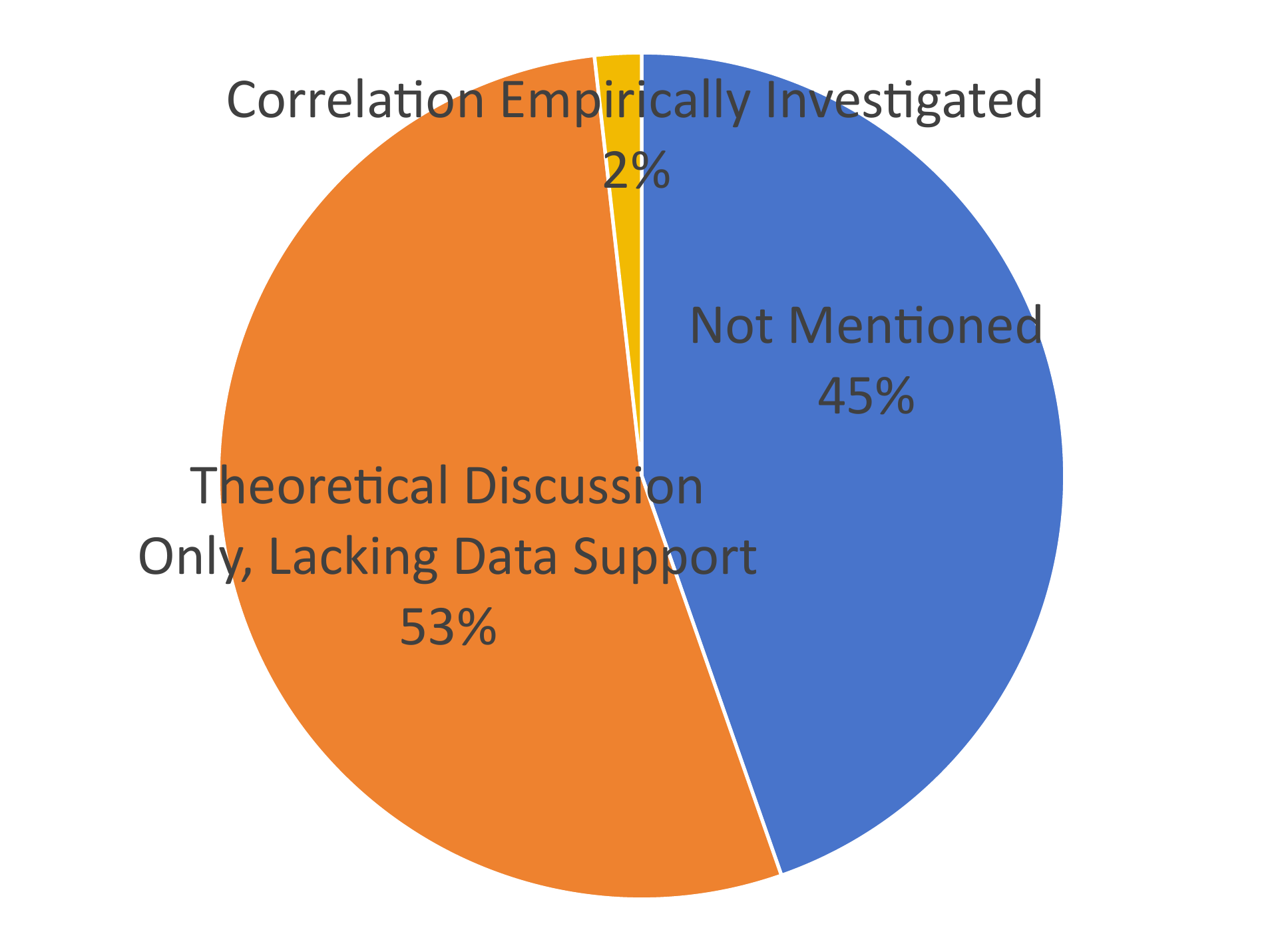}
\caption{Performance on Correlation with Clinical Performance}
\label{fig23}
\end{figure}

As presented in Figure \ref{fig23}, among the 56 benchmarks reviewed, 45\% did not address the correlation between evaluation scores and clinical performance. A further 53\% (28 benchmarks) discussed expected correlations at a theoretical level but lacked supporting experimental evidence. Only 2\% (1 benchmark) provided preliminary empirical analysis exploring the correlation, along with a discussion of the results. These findings indicate a significant gap in validating the practical relevance of benchmark scores. Future medical benchmarks should prioritize establishing and verifying correlations between evaluation outcomes and clinical workflows or expert assessments. This is essential to ensure that benchmark scores serve as credible and informative indicators for real-world model deployment.

\begin{figure}[htb]
\centering
\includegraphics[width=0.5\textwidth]{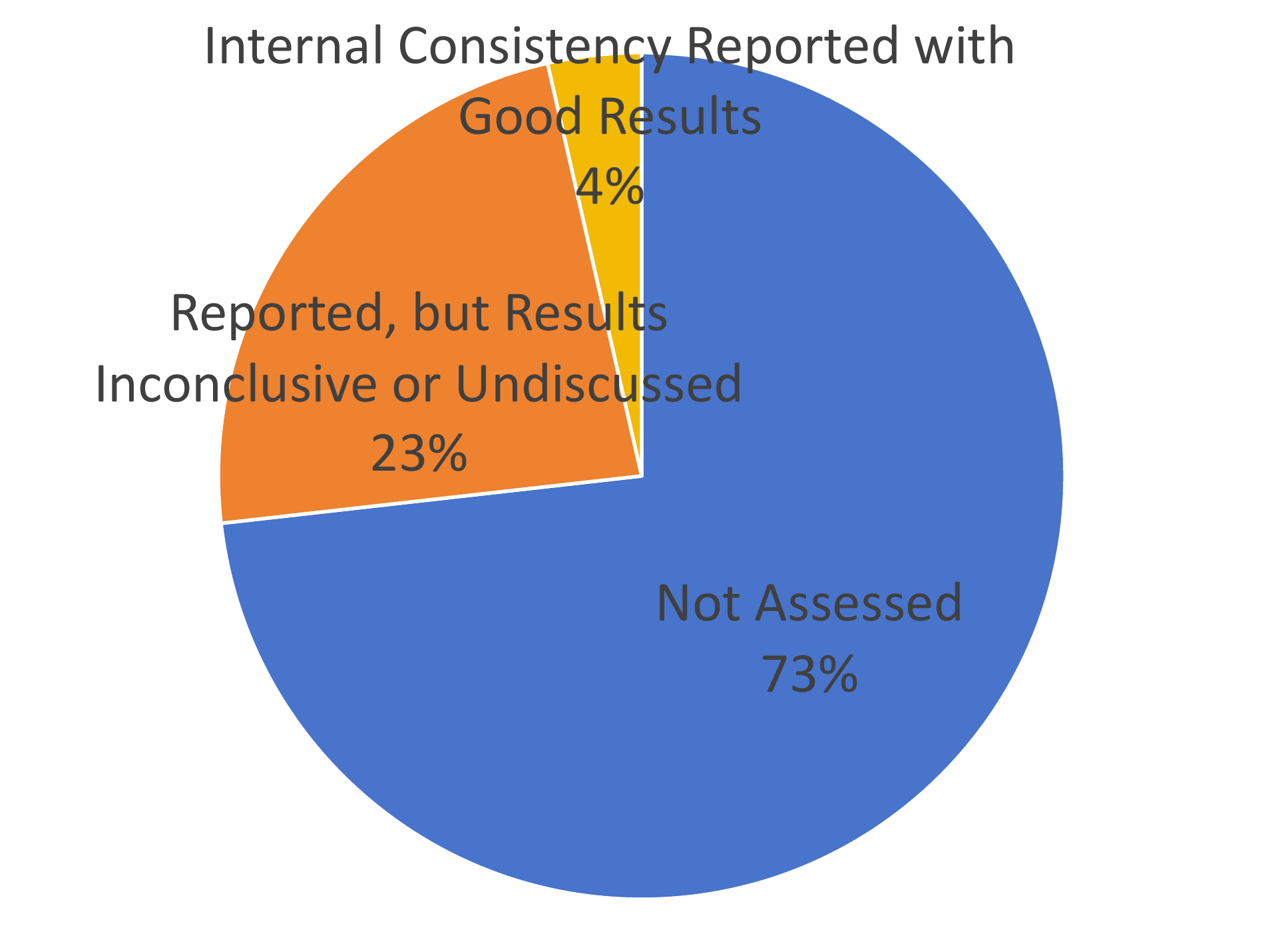}
\caption{Performance on Internal consistency}
\label{fig24}
\end{figure}

\textbf{Internal Consistency.} This metric examines whether different components or items within a benchmark consistently measure the same target capability, thereby ensuring internal reliability. High internal consistency reduces the risk that variability in item quality, phrasing, or task context will distort evaluation results. Figure \ref{fig24} shows that in this study, only 4\% of the benchmarks reported statistical indicators of internal consistency, with positive results. An additional 23\% also reported such metrics, but without clear discussion or yielded inconclusive findings. The remaining 73\% did not conduct internal consistency assessments. The absence of such analysis raises concerns regarding evaluation accuracy and fairness. Inconsistent benchmarks may obscure true model strengths or weaknesses across specific skill dimensions, impeding targeted model refinement. In clinical applications, this may further compromise risk assessment and regulatory oversight. Therefore, future benchmarks should prioritize the evaluation and reporting of internal consistency to enhance reliability.

\textbf{Statistical Significance Reporting.} This criterion evaluates whether statistical significance or uncertainty—such as confidence intervals or p-values—was reported when comparing model performances. In the absence of such analyses, observed performance differences may arise from sample randomness or dataset variability rather than genuine model capability disparities. This undermines the scientific rigor of the benchmark and may result in erroneous conclusions during model selection or optimization. In high-stakes medical scenarios, such inaccuracies pose safety risks if incorporated into decision-support systems or diagnostic tools. Consequently, the systematic reporting of statistical measures such as confidence intervals, p-values, or variance analyses is essential for constructing trustworthy and interpretable evaluation frameworks.

\begin{figure}[htb]
\centering
\includegraphics[width=0.5\textwidth]{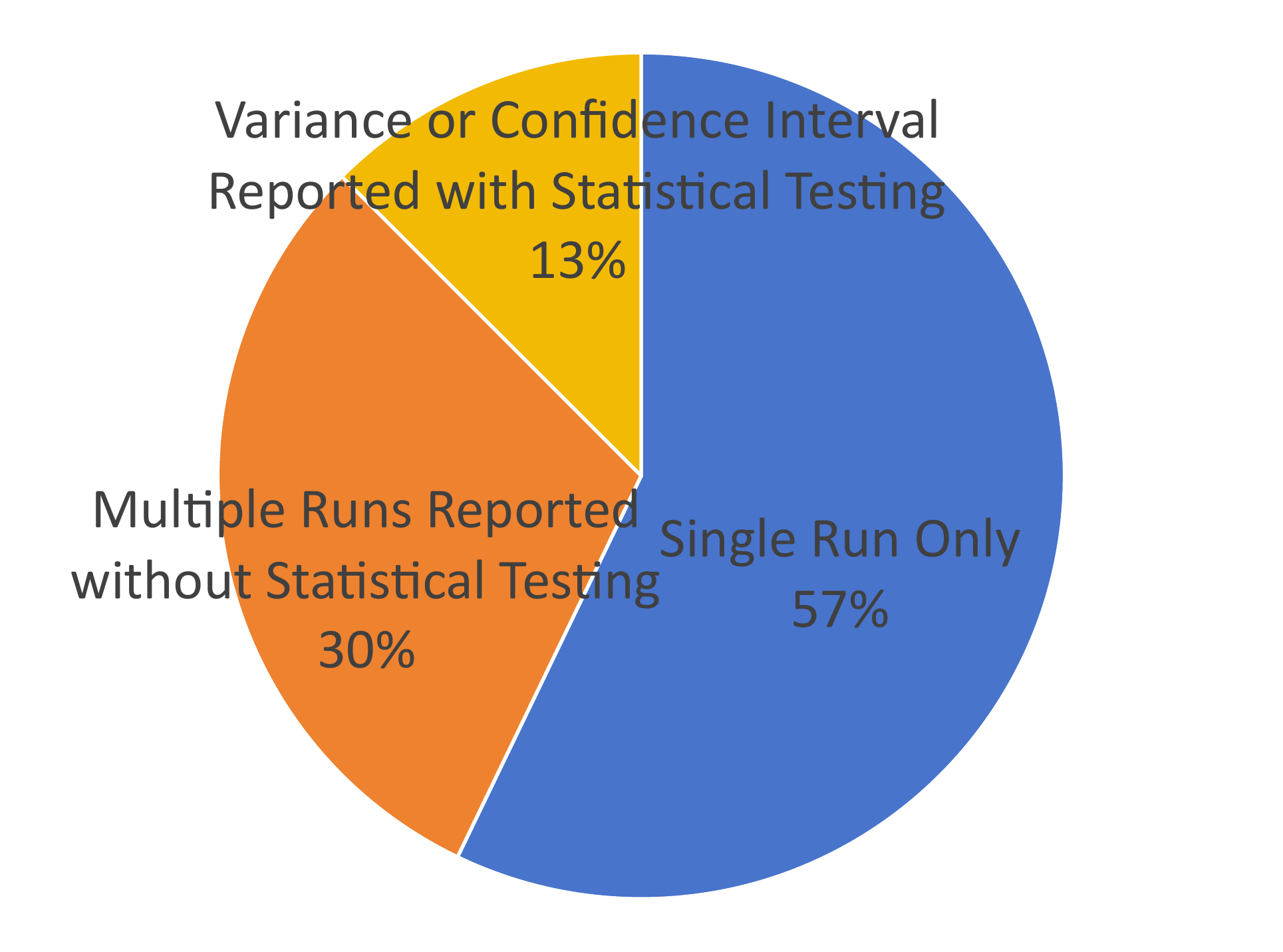}
\caption{Performance on Statistical significance Reporting}
\label{fig25}
\end{figure}

It can be seen from Figure \ref{fig25} that, in practice, 57\% of the benchmarks reported only single-run results. Although 30\% provided mean and standard deviation across multiple runs, they did not employ formal statistical tests to compare model performance. These findings highlight a lack of rigorous statistical standards in current benchmark design. Incorporating structured multi-run strategies and statistical significance testing is critical to enhance the reliability and validity of performance evaluations in medical AI.

\subsection{Statistics About Phase V: Documentation, Openness, and Governance}
\begin{figure}[htb]
\centering
\includegraphics[width=0.5\textwidth]{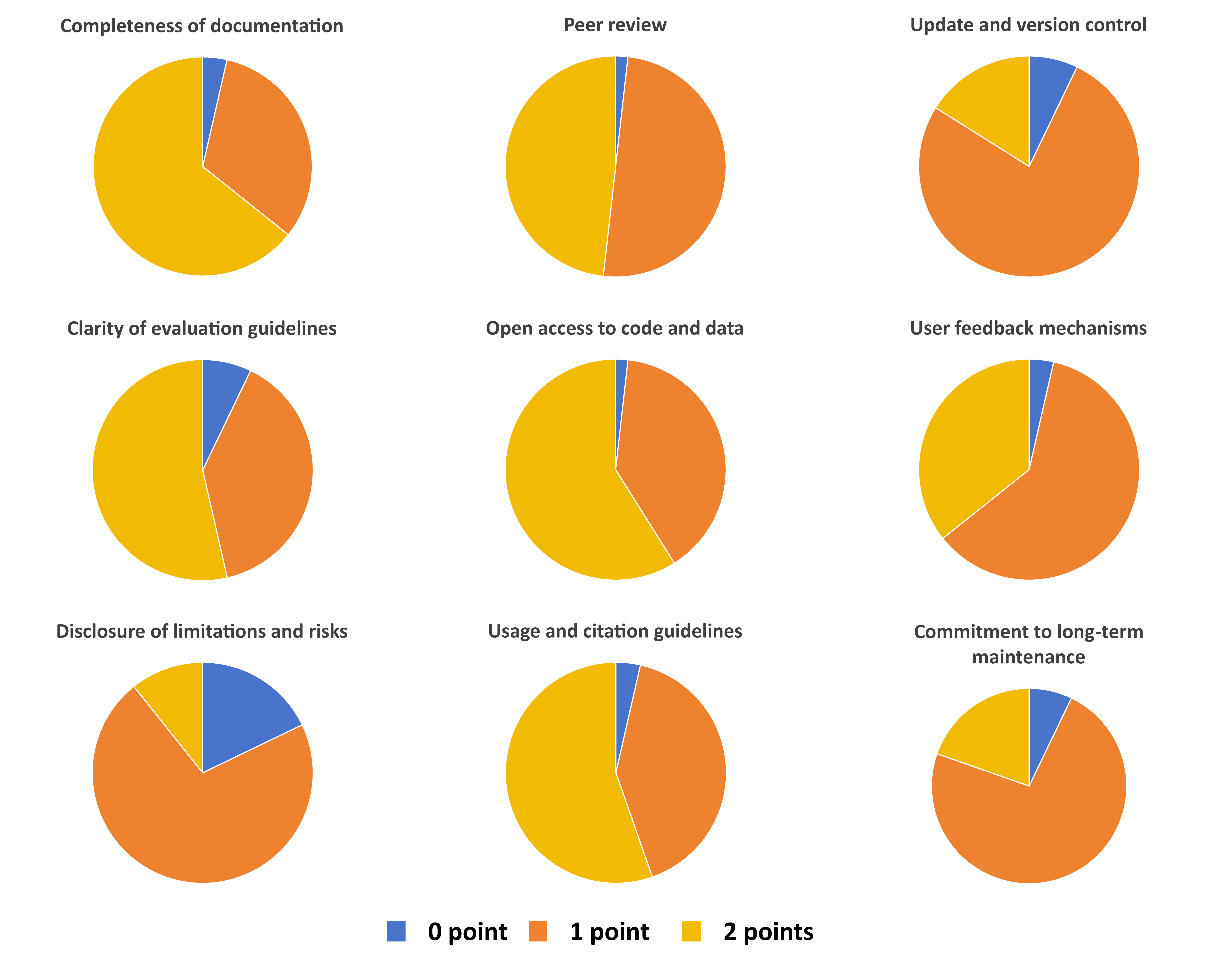}
\caption{Scoring Performance of the 56 Benchmarks in the Phase V: Documentation, Openness, and Governance}
\label{fig5}
\end{figure}

Documentation, Openness, and Governance represent community-facing components that are essential for ensuring the long-term value and impact of a benchmark. This phase includes the provision of clear documentation, adherence to open-source principles, and the establishment of governance mechanisms for maintenance, version control, and community feedback. Among the 56 benchmarks evaluated, as depicted in Figure \ref{fig5}, the average score rate for this phase was 66.2\% (11.9 out of 18), with a mean score of 1.3 per item. The scoring distribution in this phase was relatively balanced, with only one criterion showing slightly weaker performance.

\begin{figure}[htb]
\centering
\includegraphics[width=0.5\textwidth]{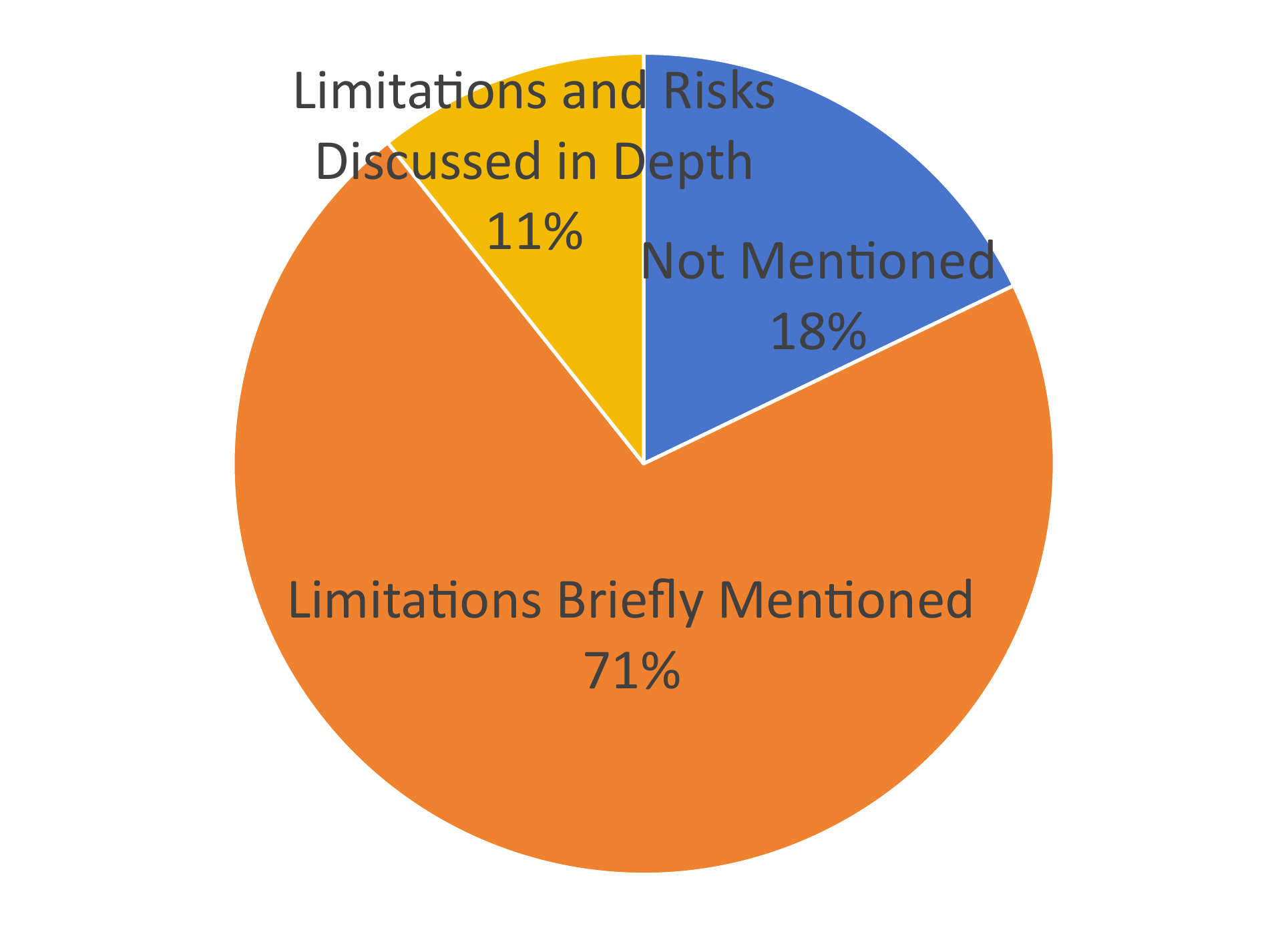}
\caption{Performance on Discussion of Limitations and Risks}
\label{fig26}
\end{figure}

\textbf{Discussion of Limitations and Risks.} This criterion evaluates whether the benchmark documentation openly and explicitly discusses its inherent limitations, potential societal risks, or misuse scenarios. No benchmark is flawless. Proactively disclosing such limitations and risks reflects scientific rigor and a strong sense of responsibility on the part of the developers. It also helps prevent overreliance on benchmark results or inappropriate applications. Figure \ref{fig26} shows that Only 11\% of the benchmarks provided thorough and candid discussions of limitations and associated risks. A majority—71\%—briefly mentioned limitations but lacked in-depth analysis. The remaining 18\% (10 out of 53) did not discuss any form of limitation or risk.

In the medical domain, the responsibilities of a benchmark extend beyond merely providing performance scores. Benchmarks should also serve as tools for risk communication and usage guidance. Failure to include such information undermines both scientific robustness and standardization, and diminishes the benchmark’s value for clinical practice and policy-making. Therefore, systematic disclosure of limitations and potential risks should not be considered optional, but rather a fundamental requirement in the design and dissemination of medical benchmarks.

\subsection{Scores per lifecycle Stage}

\begin{figure}[h]
\centering
\includegraphics[width=0.5\textwidth]{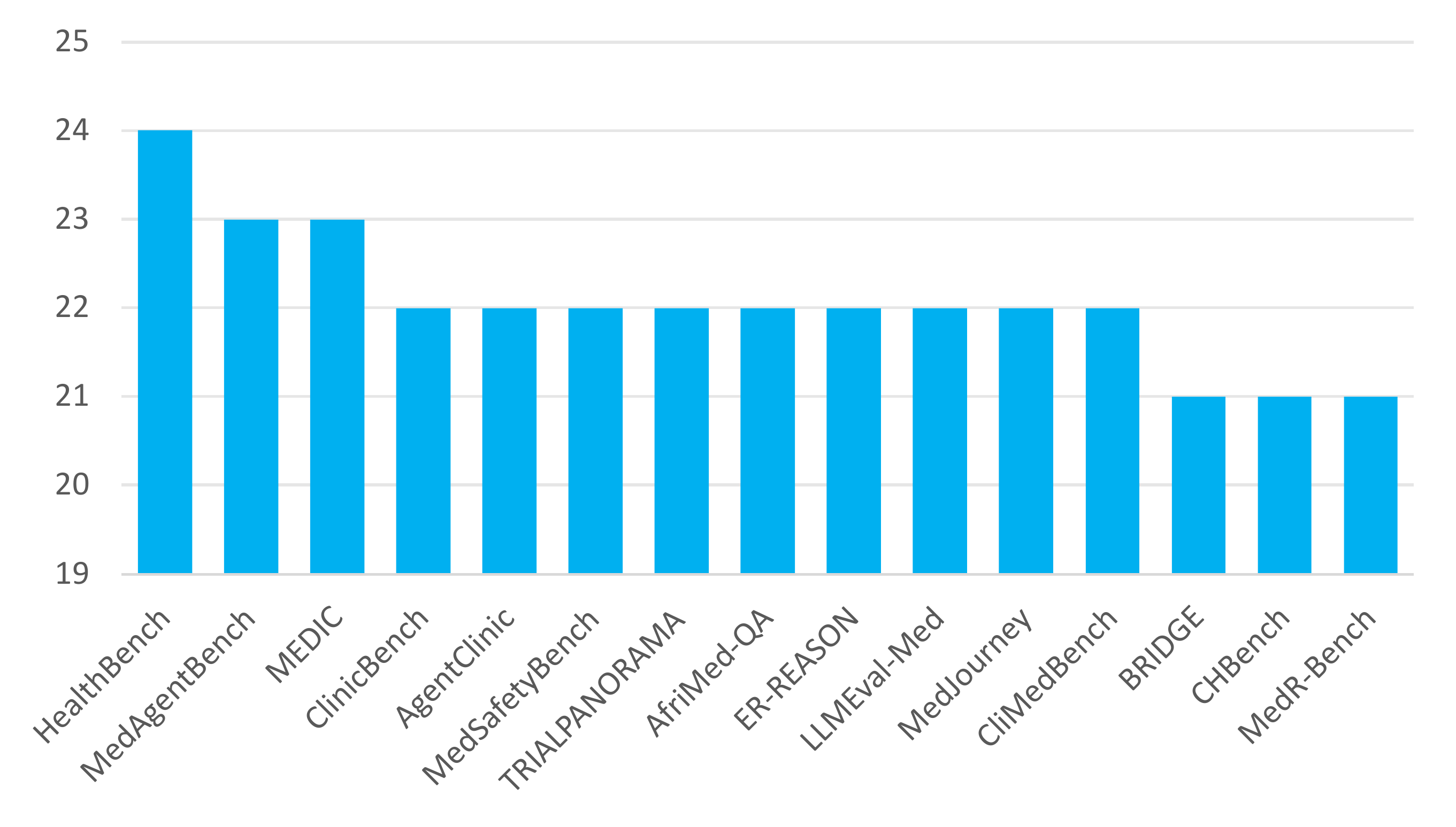}
\caption{Top 15 Benchmarks by Score in Phase I}
\label{score1}
\end{figure}

\begin{figure}[h]
\centering
\includegraphics[width=0.5\textwidth]{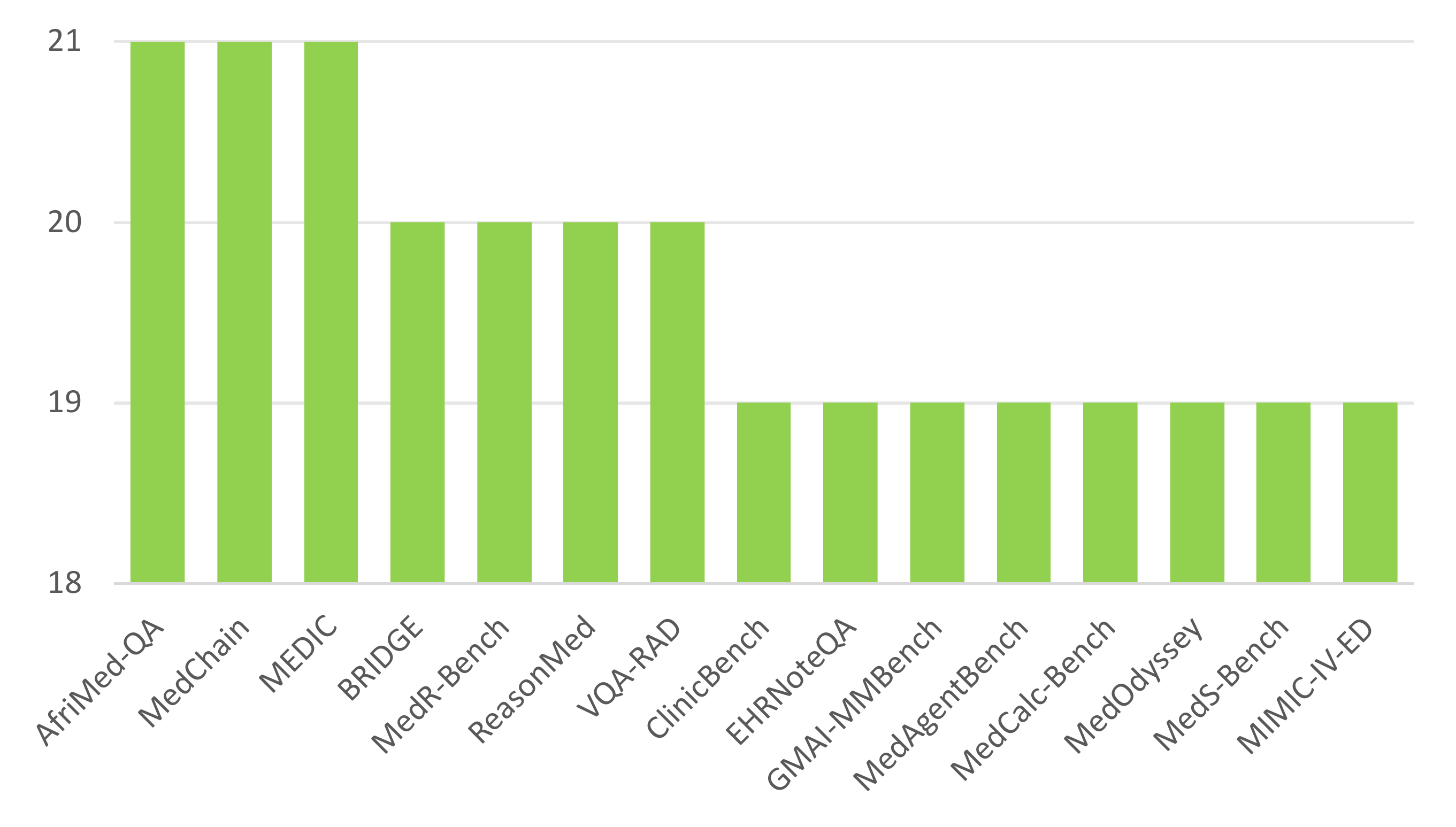}
\caption{Top 15 Benchmarks by Score in Phase II}
\label{score2}
\end{figure}

\begin{figure}[h]
\centering
\includegraphics[width=0.5\textwidth]{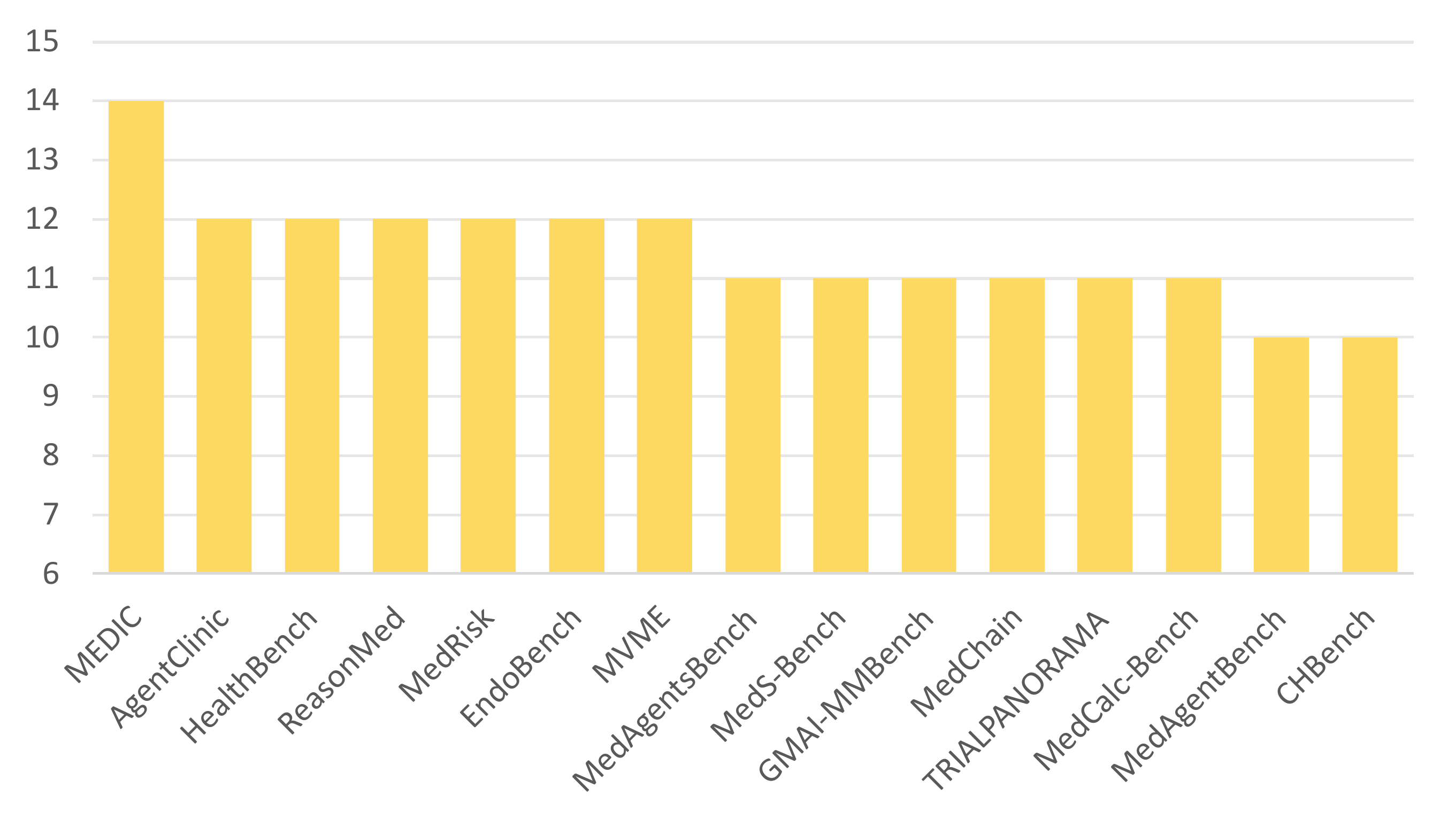}
\caption{Top 15 Benchmarks by Score in Phase III}
\label{score3}
\end{figure}

\begin{figure}[h]
\centering
\includegraphics[width=0.5\textwidth]{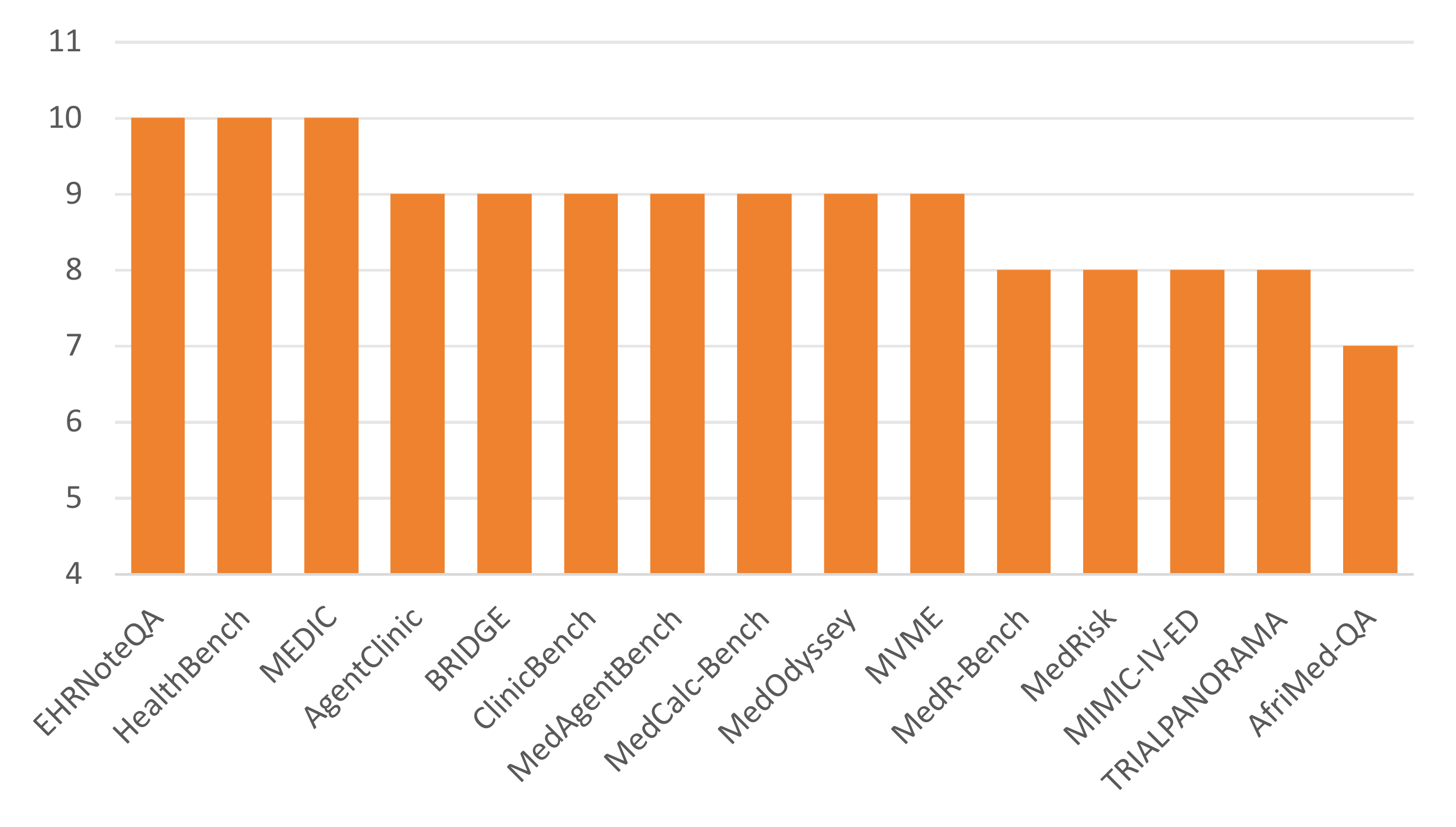}
\caption{Top 15 Benchmarks by Score in Phase IV}
\label{score4}
\end{figure}

\begin{figure}[h]
\centering
\includegraphics[width=0.5\textwidth]{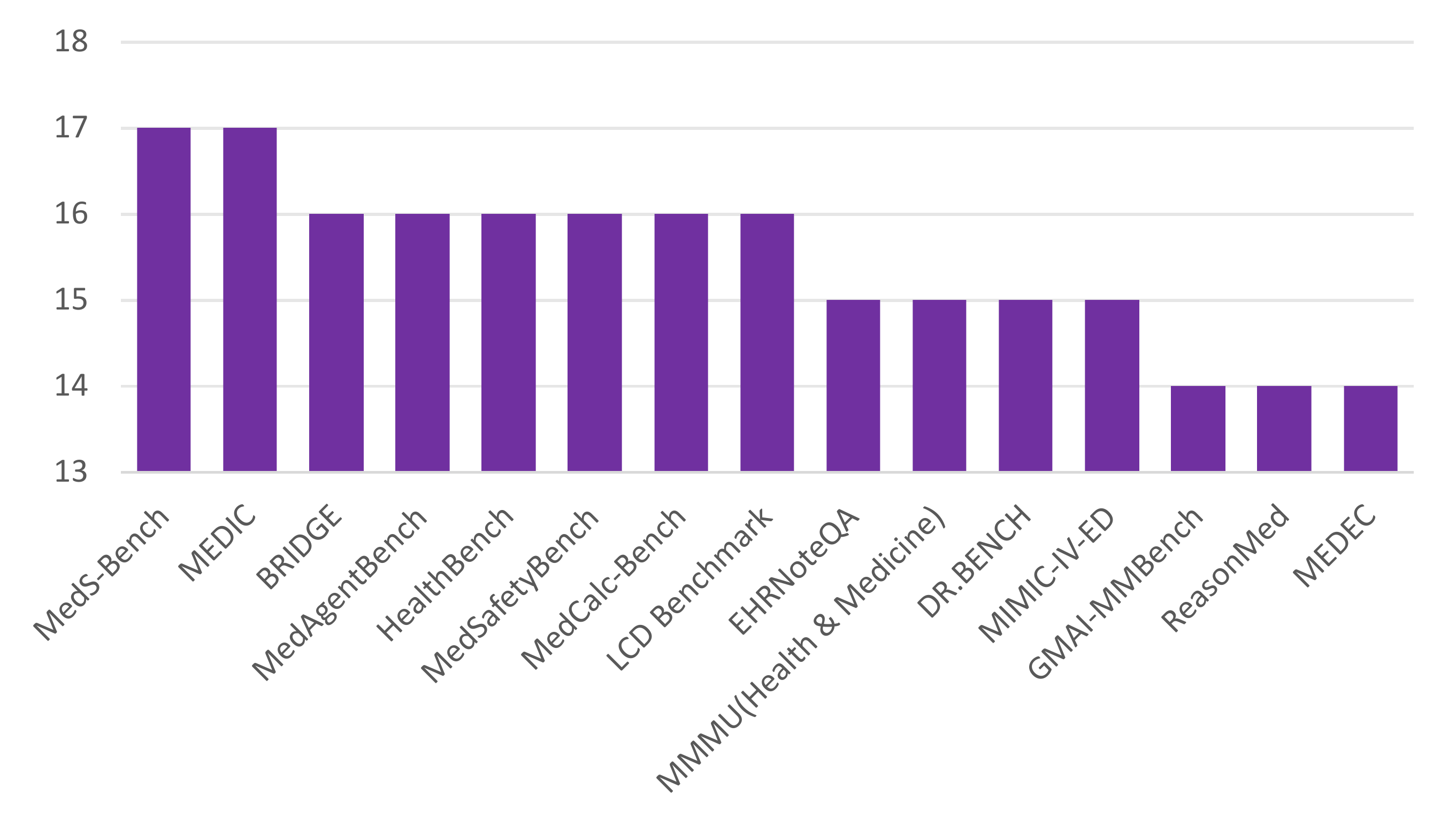}
\caption{Top 15 Benchmarks by Score in Phase V}
\label{score5}
\end{figure}

We present the top 15 benchmarks for each of the five lifecycle phases using bar charts(Figure\ref{score1} to Figure\ref{score5}). The scores for each phase reflect performance in areas such as design, dataset construction, technical implementation, validation, and documentation and governance. This visualization allows for a clear comparison of benchmark performance across phases, highlighting strengths in certain areas and identifying potential weaknesses in others.

\subsection{Validation of MedCheck’s Correlation with Clinical Safety}

\begin{table*}[t!]
\centering
\small 
\setlength{\tabcolsep}{6pt} 
\renewcommand{\arraystretch}{1.1} 
\caption{Correlation Analysis between MedCheck Scores and Safety Sensitivity. Higher MedCheck scores demonstrate a positive correlation with safety outcomes, whereas lower-scoring benchmarks (e.g., PubMedQA) show a negative correlation.}
\label{tab:medcheck_safety_corr}
\begin{tabular}{l c c c c c}
\toprule
\textbf{Benchmark} & \textbf{MedCheck Score} & \textbf{GPT-3.5} & \textbf{GPT-4} & \textbf{GPT-o3} & \textbf{\shortstack{Pearson Correlation ($r$) \\ w/ Safety Sensitivity}} \\
\midrule
HealthBench & 0.75 & 0.155 & 0.150 & 0.599 & \textbf{+0.701} \\
CMExam      & 0.62 & 0.232 & 0.257 & 0.579 & +0.647 \\
PubMedQA    & 0.31 & 0.796 & 0.752 & 0.160 & -0.647 \\
\bottomrule
\end{tabular}
\vspace{-10pt} 
\end{table*}

To explore whether MedCheck scores can serve as potential indicators of clinical utility and safety, we provide theoretical justification alongside an exploratory empirical analysis analyzing the relationship between our scoring framework and clinical safety outcomes.

MedCheck functions as a strict meta-evaluation of the evidence provided by benchmarks, assessing whether a given benchmark possesses the necessary construct validity to serve as a reliable clinical tool. Rather than generating new clinical data, the framework evaluates the extent to which a benchmark has documented its own relevance.or instance, Criterion 35 (Correlation with Clinical Performance) penalizes benchmarks that fail to provide empirical data linking their metrics to real-world outcomes. Similarly, criteria such as Criterion 12 (Safety and Fairness), Criterion 28 (Robustness Evaluation), and Criterion 30 (Uncertainty Evaluation) strictly assess the presence of necessary safeguards. Consequently, a high MedCheck score is not merely a theoretical accolade but a verification that a benchmark has demonstrably validated its own clinical utility and safety mechanisms.

We further conducted a correlation analysis to quantify the relationship between MedCheck scores and benchmark sensitivity to clinical safety issues. We averaged MedCheck scores across Phases I, III, and IV—which focus on safety-critical design, evaluation capabilities, and validation—and compared them against model performance on MedSafetyBench across GPT-3.5, GPT-4, and o3.

We selected three representative benchmarks with varying MedCheck scores: HealthBench (MedCheck score: 0.75), CMExam (0.62), and PubMedQA (0.31). The analysis reveals a strong positive correlation (Pearson r = 0.970, p = 0.156; Spearman $\rho$ = 1.000, p = 0.000) between MedCheck scores and safety sensitivity. Notably, the high-scoring HealthBench demonstrates the strongest positive correlation (r = +0.701) with safety outcomes. Conversely, PubMedQA exhibits a negative correlation (r = -0.647), indicating models performing well on lower-scoring benchmarks may actually perform worse on safety-critical tasks. This empirical evidence suggests that MedCheck effectively identifies benchmarks meaningfully associated with clinical safety and utility. However, it is important to clarify the methodological role of MedSafetyBench in this analysis. We utilize MedSafetyBench strictly as an established "ground truth" anchor for clinical safety to calibrate our framework. The goal is to demonstrate that \textit{MedCheck}'s safety-specific criteria possess the discriminative power to correctly identify high-relevance safety benchmarks, avoiding circular reasoning. 

Furthermore, while this empirical evidence suggests strong alignment, we explicitly acknowledge the limitation of the small sample size ($N=3$) in this specific correlation analysis. This analysis is intended as an exploratory proof-of-concept rather than a definitive statistical inference. Independent validation across a broader set of benchmarks and real-world clinical outcomes is required before \textit{MedCheck} scores can be definitively treated as predictive indicators.

\subsection{Validation of Construct Validity via Clinical Transferability}
\label{sec:jama_validation}

\begin{table*}[t!]
\centering
\small 
\setlength{\tabcolsep}{4pt} 
\renewcommand{\arraystretch}{1.1} 
\caption{Correlation Analysis between Benchmark Performance and Clinical Transferability (JAMA Clinical Challenge). Benchmarks with higher MedCheck scores demonstrate stronger alignment with model performance on real-world clinical cases.}
\label{tab:jama_correlation}
\begin{tabular}{l c c c c c}
\toprule
\textbf{Benchmark} & \textbf{MedCheck Score} & \textbf{GPT-4o} & \textbf{Gemini 2.0 Flash} & \textbf{Qwen2.5} & \textbf{\shortstack{Pearson Correlation ($r$) \\ w/ JAMA Ranking}} \\
\midrule
BRIDGE      & 0.83 & 0.530 & 0.530 & 0.510 & \textbf{+0.995} \\
MedQA       & 0.55 & 0.850 & 0.770 & 0.660 & +0.946 \\
COGNET-MD   & 0.35 & 0.910 & 0.810 & 0.690 & +0.932 \\
\midrule
\rowcolor{gray!10} \textit{JAMA (Holdout)} & ---  & \textbf{0.750} & \textbf{0.730} & \textbf{0.570} & --- \\
\bottomrule
\end{tabular}
\vspace{-10pt}
\end{table*}

To further validate MedCheck's construct validity, we conducted a correlation study using the \textbf{JAMA Clinical Challenge dataset} as a clinically-proximal holdout task ($N=100$ cases across 10 specialties). We selected three representative benchmarks with varying MedCheck scores: BRIDGE (0.83), MedQA (0.55), and COGNET-MD (0.35), and evaluated three leading models (GPT-4o, Gemini 2.0 Flash, and Qwen2.5) on these benchmarks alongside the JAMA holdout task.

Our analysis, summarized in \textbf{Table~\ref{tab:jama_correlation}}, reveals two key findings supporting MedCheck's validity:
\begin{itemize}
    \item \textbf{Ordered Correlation with Clinical Reality:} Benchmarks with higher MedCheck scores demonstrate stronger alignment with the clinical holdout task. As shown in the Pearson correlation ($r$) between benchmark rankings and JAMA rankings, BRIDGE exhibits the strongest correlation ($r = +0.995$), followed by MedQA ($r = +0.946$) and COGNET-MD ($r = +0.932$).
    \item \textbf{Predictive Power:} We observe a strong positive correlation between MedCheck scores and the JAMA alignment strength (Pearson $r = 0.975$, $p = 0.144$). This indicates that benchmarks scoring higher on MedCheck are significantly more effective at capturing capabilities that transfer to real-world clinical scenarios.
\end{itemize}

\section{Domain-Specific Analysis (Medical vs. Clinical)}

\begin{figure}[h]
\centering
\includegraphics[width=0.5\textwidth]{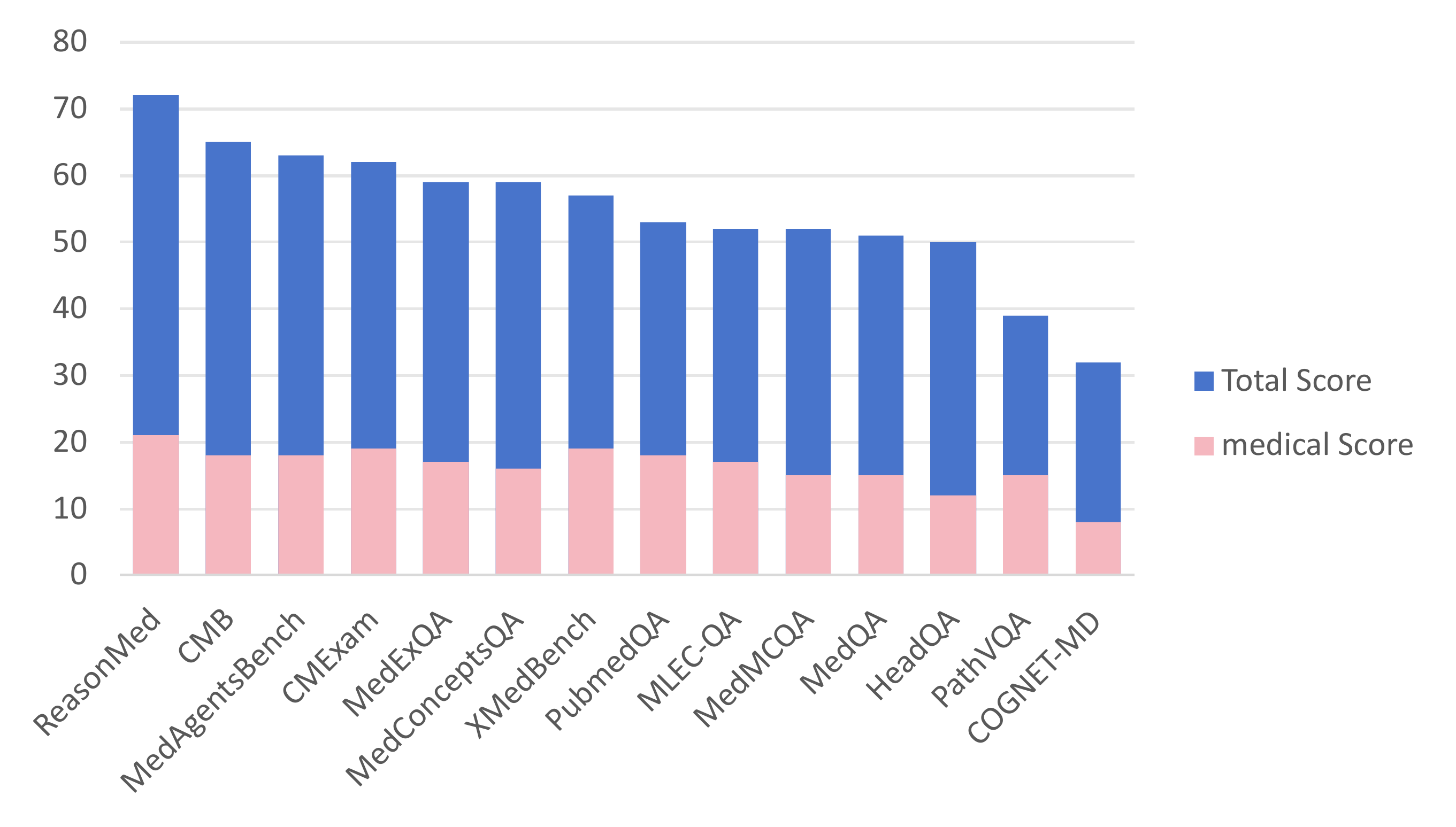}
\caption{Top 15 Medical Benchmarks by Total Score}
\label{medical}
\end{figure}

\begin{figure}[h]
\centering
\includegraphics[width=0.5\textwidth]{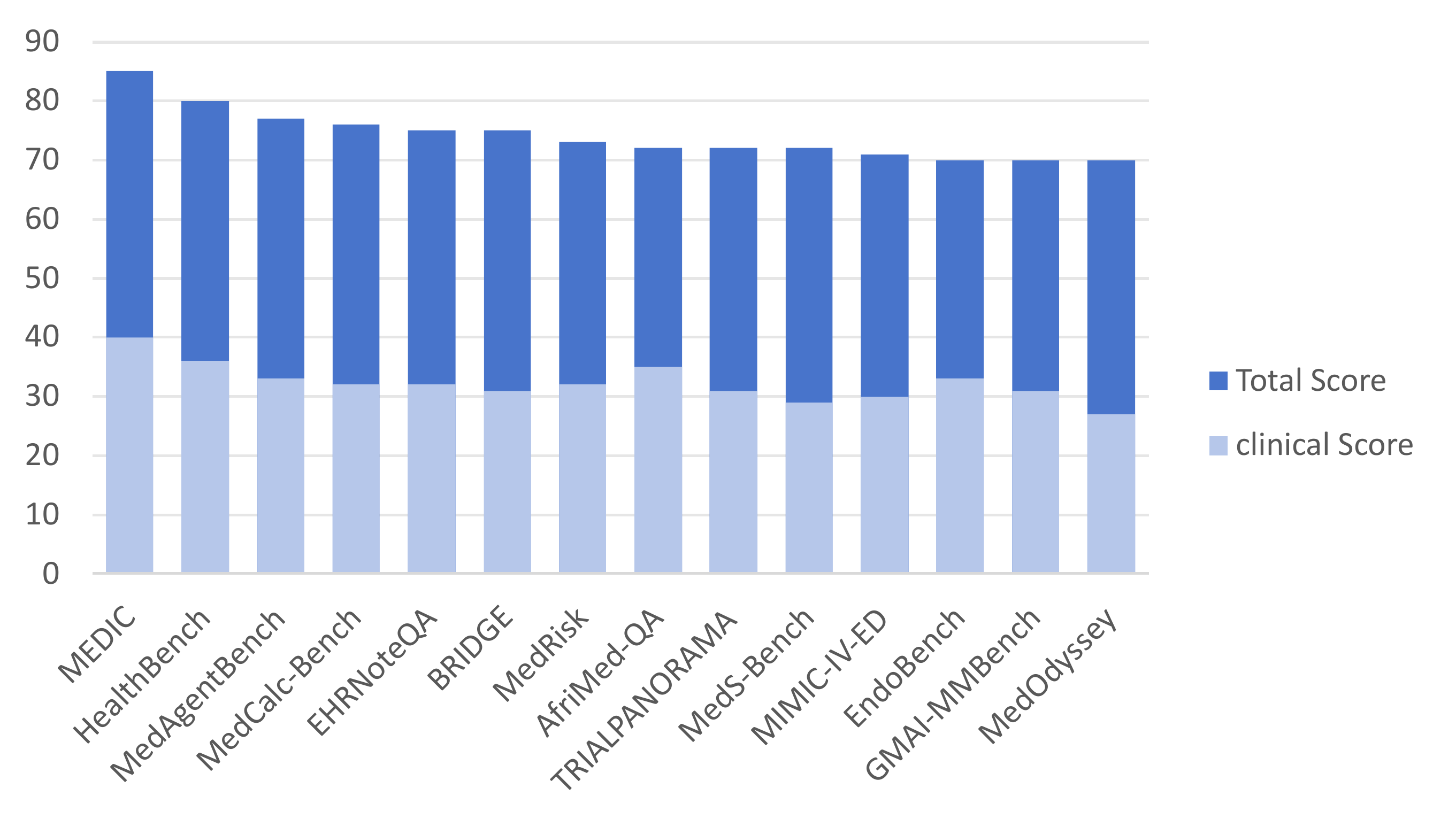}
\caption{Top 15 Clinical Benchmarks by Total Score}
\label{clinical}
\end{figure}

To address the important distinction between benchmarks for foundational medical knowledge and those for clinical practice, we tagged each of our 46 criteria as "medical," "clinical," or "general." We then re-scored all 56 benchmarks using only the criteria relevant to their primary domain.

The results confirm the utility of our framework across domains. For example, ReasonMed, a benchmark focused on medical reasoning from texts, achieves the highest score (21) among benchmarks classified as "medical" when evaluated on medical criteria alone. Conversely, MEDIC, a benchmark designed for real-world clinical applications, leads the "clinical" benchmarks with a 40 score on clinical-specific criteria. This demonstrates \textit{MedCheck}'s ability to appropriately evaluate benchmarks according to their intended purpose. The domain-specific scoring breakdown for the top 15 medical and clinical benchmarks is provided in Figure~\ref{medical} and Figure \ref{clinical}.

\section{Full Evaluation Criteria}
\label{sec:appendix_full_checklist}
We implemented a comprehensive tagging system categorizing our 46 criteria as Medical (M), Clinical (C), or General (G), and provided domain-specific scoring analysis for all 56 benchmarks, demonstrating that our framework appropriately evaluates benchmarks according to their intended purpose.
\subsection{Design and Conceptualization}

\subsubsection{Purpose and Intent}
\begin{enumerate}
    \item \textbf{Clarity of Evaluation Objectives (G)}
    \hypertarget{item:1}{}
        \begin{itemize}
            \item \textbf{Explanation}: Benchmark developers should clearly specifies the targeted capabilities of LLMs it aims to evaluate in the medical field (e.g., medical knowledge question answering, diagnosis, report generation).
            \item \textbf{Justification}: Clearly defined evaluation objectives can avoid ambiguity, facilitating subsequent data collection, task design, and metric selection. Also, it helps users determine whether the benchmark aligns with their specific evaluation needs.
            \item \textbf{Scoring}:
                \begin{itemize}
                    \item 0: Does not mention or define the evaluation objective
                    \item 1: Mentions the evaluation objectives but the definition is broad or insufficiently detailed.
                    \item 2: Explicitly defines the medical capabilities being evaluated and provides examples.
                \end{itemize}
        \end{itemize}
    \item \textbf{Clarity of Application Scenario (C\&M)}
    \hypertarget{item:2}{}
        \begin{itemize}
            \item \textbf{Explanation}: Benchmark developers should clearly describes the specific medical application scenarios it corresponds to and explains the potential value.
            \item \textbf{Justification}: Linking the benchmark to real-world application scenarios ensures that the results are meaningful. It facilitates the validation of model’s effectiveness in specific settings, ultimately better serving users such as doctors and researchers.
            \item \textbf{Scoring}:
                \begin{itemize}
                    \item 0: Does not describe any application scenarios.
                    \item 1: Mentions application scenarios but provides vague descriptions or fails to clarify the potential value.
                    \item 2: Clearly describes the application scenarios and elaborates on the value in detail.
                \end{itemize}
        \end{itemize}
    \item \textbf{Uniqueness and Novelty (G)}
    \hypertarget{item:3}{}
        \begin{itemize}
            \item \textbf{Explanation}: By comparing with relevant benchmarks, benchmark developers should explain their contributions and the uniqueness of the benchmark, such as filling a gap and proposing new evaluation methodology.
            \item \textbf{Justification}: Demonstrating the uniqueness demonstrates the necessity and justification of the new benchmark. It also helps the community to better understand its unique value, promoting continual innovation in evaluation and benchmarking.
            \item \textbf{Scoring}:
                \begin{itemize}
                    \item 0: Does not compare with relevant benchmarks or does not mention its uniqueness.
                    \item 1: Briefly mentions other benchmark, but the comparison is insufficient to explain its unique value.
                    \item 2: Provides detailed comparison with relevant benchmarks and clearly argues its uniqueness.
                \end{itemize}
        \end{itemize}
\end{enumerate}

\subsubsection{Scope and Applicability}
\begin{enumerate}
    \item \textbf{Target Capability of Evaluation (G)}
    \hypertarget{item:4}{}
        \begin{itemize}
            \item \textbf{Explanation}: Benchmark developers should clearly define the capability of LLMs intended to evaluate (e.g., text generation, multimodal understanding)
            \item \textbf{Justification}: Clearly defining the targeted capability helps clarify the scope of the benchmark, which can prevent misusing the benchmark.
            \item \textbf{Scoring}:
                \begin{itemize}
                    \item 0: Does not define the target LLM capability.
                    \item 1: Briefly mentions the target capability without clear definition.
                    \item 2: Clearly defines and explains the target LLM capability.
                \end{itemize}
        \end{itemize}
    \item \textbf{Medical Domain Coverage (M)}
    \hypertarget{item:5}{}
        \begin{itemize}
            \item \textbf{Explanation}: Benchmark developers should clearly define the scope of the medical specialties of the benchmark, such as clinical departments, disease types, or task types.
            \item \textbf{Justification}: By clearly defining the medical scope, it helps users better understand the breath and depth of the coverage of the benchmark. It also helps users determine whether the benchmark aligns with their needs.
            \item \textbf{Scoring}:
                \begin{itemize}
                    \item 0: The medical scope and coverage are not defined.
                    \item 1: The medical scope and coverage are only briefly mentioned.
                    \item 2: The medical scope and coverage are clearly defined and explained.
                \end{itemize}
        \end{itemize}
    \item \textbf{Demonstration of User Needs (C)}
    \hypertarget{item:6}{}
        \begin{itemize}
            \item \textbf{Explanation}: The benchmark should reflect the core concerns and assessment needs of its target users, such as addressing specific clinical challenges or overcoming technical obstacles.
            \item \textbf{Justification}: An effective benchmark should serve the needs of users. Evidence of user needs justifies the development of the benchmark. This boosts credibility, promotes adoption, and helps guide model improvements.
            \item \textbf{Scoring}:
                \begin{itemize}
                    \item 0: Does not reflect any consideration of user needs.
                    \item 1: Briefly mentions user needs, but lacks concrete external evidence.
                    \item 2: Clearly explains user needs to justify design choices and connects user needs to benchmark tasks (e.g., referencing relevant literature, user surveys, or expert interviews).
                \end{itemize}
        \end{itemize}
\end{enumerate}

\subsubsection{Medical Expertise and Professionalism}
\begin{enumerate}
    \item \textbf{Domain Experts Involvement (C\&M)}
    \hypertarget{item:7}{}
        \begin{itemize}
            \item \textbf{Explanation}: Domain expert (e.g., physicians or clinical researchers) should be involved in the development of the benchmark.
            \item \textbf{Justification}: Due to the professionalism and rigor required in the medical field, the development of a benchmark must involve deep engagement from domain experts. Their expertise are fundamental to ensuring data quality, task validity, authenticity and relevance.
            \item \textbf{Scoring}:
                \begin{itemize}
                    \item 0: No mention of medical expert participation.
                    \item 1: Briefly mentions expert involvement, but the information is unclear and lacks details on the roles or involved tasks.
                    \item 2: Clearly describes the experts’ qualifications and their involvement, including specific roles and tasks throughout the development process.
                \end{itemize}
        \end{itemize}
    \item \textbf{Authoritative Knowledge Sources (C\&M))}
    \hypertarget{item:8}{}
        \begin{itemize}
            \item \textbf{Explanation}: Benchmark developers must clearly specify which authoritative medical knowledge sources (e.g., clinical guidelines, textbooks, medical databases) the benchmark content is based on.
            \item \textbf{Justification}: Benchmark content should adhere to recognized, evidence-based medical knowledge sources. By using authoritative sources, it ensures that the benchmark is scientifically reliable, enhancing credibility and transparency.
            \item \textbf{Scoring}:
                \begin{itemize}
                    \item 0: No knowledge sources listed.
                    \item 1: Mentions knowledge sources, but the description is vague or the sources are not sufficiently authoritative.
                    \item 2: Clearly lists and explains the use of authoritative medical knowledge sources.
                \end{itemize}
        \end{itemize}
    \item \textbf{Medical Standards Alignment (C\&M)}
    \hypertarget{item:9}{}
        \begin{itemize}
            \item \textbf{Explanation}: Benchmarks should follow internationally or industry-recognized medical standards (e.g., ICD, SNOMED CT, LOINC) when medical terminology is involved.
            \item \textbf{Justification}: Adherence to medical standards ensures the clinical relevance and consistency, facilitating integration and comparison in reflect real-world medical practice.
            \item \textbf{Scoring}:
                \begin{itemize}
                    \item 0: Does not mention any medical standards.
                    \item 1: Mentions the importance of standardization, but does not specify which standards are followed, or the description is unclear.
                    \item 2: Clearly states the adherence to recognized medical standards.
                \end{itemize}
        \end{itemize}
\end{enumerate}

\subsubsection{Evaluation Metrics and Dimensions}
\begin{enumerate}
    \item \textbf{Validity of Core Metric (C\&M)}
    \hypertarget{item:10}{}
        \begin{itemize}
            \item \textbf{Explanation}: The core performance metric should be clearly defined and closely related to the clinical task or medical capability being assessed.
            \item \textbf{Justification}: Evaluation metrics directly shape the interpretation of results. Choosing suitable metric and explaining its relevance ensures a shared understanding and comprehensive interpretation.
            \item \textbf{Scoring}:
                \begin{itemize}
                    \item 0: Core metrics are not clearly defined or have weak relevance to the task.
                    \item 1: Core metrics are clearly defined, but the choice and relevance of the metrics are not explained and justified.
                    \item 2: Core metrics are clearly defined, with justification for the relevance to the evaluation task.
                \end{itemize}
        \end{itemize}
    \item \textbf{Multi-dimensional Evaluation (C)}
    \hypertarget{item:11}{}
        \begin{itemize}
            \item \textbf{Explanation}: Apart from the correctness, benchmark developers should include evaluation of other important dimensions (e.g., safety, fluency).
            \item \textbf{Justification}: In high-stakes medical domain, going beyond correctness is vital. Multi-dimensional evaluation offers a more comprehensive assessment of whether a model is truly reliable and trustworthy.
            \item \textbf{Scoring}:
                \begin{itemize}
                    \item 0: Only consider the correctness of answers.
                    \item 1: Other evaluation dimensions are briefly mentioned, without specific evaluation methods.
                    \item 2: Multiple evaluation dimensions are clearly designed, with well defined evaluation methods.
                \end{itemize}
        \end{itemize}
    \item \textbf{Safety and Fairness Considerations (C)}
    \hypertarget{item:12}{}
        \begin{itemize}
            \item \textbf{Explanation}: Benchmark should include evaluation for potential risks (e.g., unsafe recommendation, toxicity) and bias (e.g., gender, ethnicity) in model outputs.
            \item \textbf{Justification}: Evaluating risks and fairness facilitates bias-free, safe and equitable applications of LLMs in the medical domain, promoting responsible AI development and application.
            \item \textbf{Scoring}:
                \begin{itemize}
                    \item 0: Does not consider safety and fairness evaluation.
                    \item 1: Mentions the importance of safety and fairness, but lacks specific evaluation methods.
                    \item 2: Designs dedicated assessment and test cases for safety and fairness.
                \end{itemize}
        \end{itemize}
\end{enumerate}

\subsection{Dataset Construction and Management}

\subsubsection{Data Source and Quality}
\begin{enumerate}
    \item \textbf{Data source transparency and traceability (G)}
    \hypertarget{item:13}{}
        \begin{itemize}
            \item \textbf{Explanation}: Benchmark developers should clearly state the data source of the benchmark, along with relevant traceability information (e.g., data collection time frame, platforms).
            \item \textbf{Justification}: Clear and traceable data sources are critical to ensure transparency and ethical data usage, which is especially important when sensitive data is involved. Also, it ensures reproducibility, enhancing the credibility of the benchmark.
            \item \textbf{Scoring}:
                \begin{itemize}
                    \item 0: Does not state the original data source of the benchmark.
                    \item 1: States the data source of the benchmark, with incomplete traceability information.
                    \item 2: Clearly states the data source of the benchmark and provides detailed traceability information.
                \end{itemize}
        \end{itemize}
    \item \textbf{Data Source Reliability (C\&M)}
    \hypertarget{item:14}{}
        \begin{itemize}
            \item \textbf{Explanation}: Benchmark developers should clearly describe the selection criteria and explain the reliability of the data source.
            \item \textbf{Justification}: Collecting data from unreliable sources may lead to invalid results for medical applications. By explaining the reliability of the data source, credibility of the benchmark can be established.
            \item \textbf{Scoring}:
                \begin{itemize}
                    \item 0: Does not describe the quality or selection criteria of the data source.
                    \item 1: Mentions the reliability of the data source, but does not explain the selection criteria.
                    \item 2: Clearly explains the selection criteria and describes the reliability of the data source.
                \end{itemize}
        \end{itemize}
    \item \textbf{Data Authenticity (C)}
    \hypertarget{item:15}{}
        \begin{itemize}
            \item \textbf{Explanation}: Benchmark developers should clearly specify whether the data comes from the real world scenarios, synthetically generated, or a mixture of both. For synthetically generated data, the construction process and verification for its authenticity (e.g., expert review) should also be described.
            \item \textbf{Justification}: Real-world data reflects authentic scenarios but is harder to obtain, whereas synthetic data can be generated but requires careful verification. Clarifying data authenticity enhance transparency and ensures clinical relevance of the data used.
            \item \textbf{Scoring}:
                \begin{itemize}
                    \item 0: Does not state the data source and explain its authenticity.
                    \item 1: States the data source, but lacks sufficient details on synthetic data generation or verification.
                    \item 2: Clearly states the data source, and provides a detailed description of the generation and verification process for synthetic data.
                \end{itemize}
        \end{itemize}
    \item \textbf{Dataset Representativeness (C\&M)}
    \hypertarget{item:16}{}
        \begin{itemize}
            \item \textbf{Explanation}: The representativeness of key features (e.g., patient age, disease) of the dataset should be explained and statistically analyzed.
            \item \textbf{Justification}: A benchmark that lacks representativeness may lead to bias in evaluation, reducing the clinical relevance, generalizability and fairness of the results.
            \item \textbf{Scoring}:
                \begin{itemize}
                    \item 0: Does not mention the representativeness of the dataset.
                    \item 1: Provides qualitative descriptions of representativeness or partial and incomplete statistical analysis for some features.
                    \item 2: Provides quantitative statistical analysis of key features and compares them against the target population’s distribution.
                \end{itemize}
        \end{itemize}
    \item \textbf{Dataset Diversity (C\&M)}
    \hypertarget{item:17}{}
        \begin{itemize}
            \item \textbf{Explanation}: The benchmark should have a diverse coverage, and provide quantitative evidence of its diversity of disease or medical departments covered.
            \item \textbf{Justification}: Ensuring the dataset covers a variety helps comprehensively evaluate the model’s generalization ability, reducing bias in the evaluation results.
            \item \textbf{Scoring}:
                \begin{itemize}
                    \item 0: Does not clearly define or explain the disease and medical department coverage.
                    \item 1: Clearly states the disease or medical department coverage, but coverage (defined in Equation \ref{eq1}) is less than average (21.8\%).
                    \item 2: Clearly states the disease or medical department coverage with a coverage (defined in Equation \ref{eq1}) above average (21.8\%).
                \end{itemize}
        \end{itemize}
\end{enumerate}

\subsubsection{Data Processing and Privacy Protection}
\begin{enumerate}
    \item \textbf{Data Cleaning and Standardization (C)}
    \hypertarget{item:18}{}
        \begin{itemize}
            \item \textbf{Explanation}: The processes and steps of data preprocessing, including data cleaning and standardization, should be clearly described.
            \item \textbf{Justification}: It ensures that the final dataset is well-structured, enhancing reliability. Also, it allows a better understanding of the construction process of the benchmark, ensuring transparency.
            \item \textbf{Scoring}:
                \begin{itemize}
                    \item 0: Does not describe any data cleaning or standardization steps.
                    \item 1: The data preprocessing procedure is only briefly or partially mentioned.
                    \item 2: The data preprocessing procedure is clearly described, with details that cover all steps and aspects of the process.
                \end{itemize}
        \end{itemize}
    \item \textbf{Privacy Protection (C\&M)}
    \hypertarget{item:19}{}
        \begin{itemize}
            \item \textbf{Explanation}: If sensitive data is used, it should be de-identified. Methods of de-identification should be described and compliance with relevant regulations (e.g., HIPAA) should be clearly stated.
            \item \textbf{Justification}: Real-world clinical data may contain patient information. It is essential to ensure that the data and privacy protection aligns with ethical and legal standards, reducing the risk of privacy breaches and ensuring compliant data usage.
            \item \textbf{Scoring}:
                \begin{itemize}
                    \item 0: No privacy protection measures are mentioned.
                    \item 1: Privacy protection measures are mentioned, but the description of methods or regulations is unclear.
                    \item 2: Sensitive data is not used. Otherwise, privacy protection measures are clearly described and compliance with relevant regulations is stated.
                \end{itemize}
        \end{itemize}
    \item \textbf{Data Format Clarity and Consistency (G)}
    \hypertarget{item:20}{}
        \begin{itemize}
            \item \textbf{Explanation}: The data in the dataset, including questions, cases, or task descriptions, should be written clearly and presented in a consistent format.
            \item \textbf{Justification}: A clear and consistent format is essential for standardized evaluation. Inconsistent format may affect models' interpretation, compromising the accuracy of the evaluation.
            \item \textbf{Scoring}:
                \begin{itemize}
                    \item 0: Data format is disorganized and contains ambiguities.
                    \item 1: Minor inconsistencies or ambiguities exist in data format.
                    \item 2: Data format is clear and consistent.
                \end{itemize}
        \end{itemize}
    \item \textbf{Data Review and Audit (C\&M)}
    \hypertarget{item:21}{}
        \begin{itemize}
            \item \textbf{Explanation}: The dataset construction process should include a review procedure with involvement of medical experts.
            \item \textbf{Justification}: A dataset construction process without review mechanism is prone to errors. Careful review that involves medical experts can ensures clinical relevance, professionalism and reliable of the data.
            \item \textbf{Scoring}:
                \begin{itemize}
                    \item 0: No review procedure is  described.
                    \item 1: Briefly mentions the review procedure, or the review procedure is simple and without experts involvement.
                    \item 2: Provides a clear and detailed description of the review process, including the involvement of medical experts.
                \end{itemize}
        \end{itemize}
    \item \textbf{Quality of Reference Answer (G)}
    \hypertarget{item:22}{}
        \begin{itemize}
            \item \textbf{Explanation}: The benchmark should provide clear and accurate reference answers or scoring guidelines, and explain the construction and verification process (e.g., expert consensus).
            \item \textbf{Justification}: Clear reference answers or scoring guidelines ensures transparent and accurate evaluation. By explaining how are reference answers or scoring guidelines formulated and verified, it also enhances the credibility of evaluation results.
            \item \textbf{Scoring}:
                \begin{itemize}
                    \item 0: No reference answers or scoring guidelines are provided.
                    \item 1: Reference answers  or scoring guidelines are provided, but the formulation or verification process is not explained.
                    \item 2: Provides clear reference answers or scoring guidelines, and explains its formulation and verification process in details.
                \end{itemize}
        \end{itemize}
    \item \textbf{Data Contamination Prevention (G)}
    \hypertarget{item:23}{}
        \begin{itemize}
            \item \textbf{Explanation}: Benchmark developers should take actions to identify, address and prevent potential data contamination issues of the data.
            \item \textbf{Justification}: Data contamination may lead to inflated performance, which only reflect memorization instead of medical capability from the models. Preventing and controlling potential contamination ensures that the benchmark is effective, enhancing credibility and validity.
            \item \textbf{Scoring}:
                \begin{itemize}
                    \item 0: Does not mention or consider potential data contamination.
                    \item 1: Mentions the risks of data contamination but lacks concrete, actionable steps to handle it.
                    \item 2: Clearly describes procedures taken to prevent or detect contamination using feasible, developer-controlled practices (e.g., encrypting test sets, distributing data via a gated API, using Canary GUIDs, or conducting n-gram overlap checks against open pre-training corpora). We evaluate based on actionable preventive measures rather than an impossible guarantee of zero contamination, ensuring fair assessment even when closed-source models are involved.
                \end{itemize}
        \end{itemize}
\end{enumerate}

\subsection{Technical Implementation and Evaluation Methodology}

\begin{enumerate}
    \item \textbf{User-friendliness of evaluation tools (G)}
    \hypertarget{item:24}{}
        \begin{itemize}
            \item \textbf{Explanation}: Evaluation scripts or tools that is easy to obtain and use should be provided.
            \item \textbf{Justification}: It ensures that users can use the benchmark conveniently, promoting benchmark adoption and ensuring fair, transparent, and consistent evaluation.
            \item \textbf{Scoring}:
                \begin{itemize}
                    \item 0: No evaluation scripts or tools is provided.
                    \item 1: Provides evaluation scripts, but requires users to perform complex environment configurations or manual settings.
                    \item 2: Provides evaluation scripts or tools with detailed instructions for streamlined execution, that are user-friendly and require minimal user effort.
                \end{itemize}
        \end{itemize}
    \item \textbf{Technical Reproducibility (G)}
    \hypertarget{item:25}{}
        \begin{itemize}
            \item \textbf{Explanation}: Tools and technical documentation (e.g., environment configuration, dependency versions) should be provided.
            \item \textbf{Justification}: The availability of evaluation tools and clear technical documentation ensures reproducibility of reported results, thereby enhancing the credibility of the benchmark.
            \item \textbf{Scoring}:
                \begin{itemize}
                    \item 0: No information for technical reproducibility has been provided.
                    \item 1: Information for technical reproducibility is partially provided, but insufficient to guarantee reproducibility.
                    \item 2: Complete information for reproducibility is provided with detailed steps to replicate the results.
                \end{itemize}
        \end{itemize}
    \item \textbf{Provision of performance baselines (G)}
    \hypertarget{item:26}{}
        \begin{itemize}
            \item \textbf{Explanation}: Benchmark developers should provide multiple meaningful performance baselines, such as random, baseline models and human performance.
            \item \textbf{Justification}: Providing different performance baselines allows comparison against the model's performance, enabling a deeper understanding and better interpretability.
            \item \textbf{Scoring}:
                \begin{itemize}
                    \item 0: No performance baselines are provided.
                    \item 1: Only one type of performance baseline is provided, or performance baselines are mentioned without specific data.
                    \item 2: At least two meaningful types of performance baselines are provided, with clear explanation of the measurement methods.
                \end{itemize}
        \end{itemize}
    \item \textbf{Reasoning Process Evaluation (C)}
    \hypertarget{item:27}{}
        \begin{itemize}
            \item \textbf{Explanation}: Apart from the final answer, the benchmark includes evaluations for the model's reasoning process or explanation abilities.
            \item \textbf{Justification}: In medical domain, understanding the model's decision-making process is just as important as the final answer. Evaluating the reasoning process helps ensure that the model’s decisions are trustworthy and logically sound.
            \item \textbf{Scoring}:
                \begin{itemize}
                    \item 0: Does not consider reasoning process evaluation.
                    \item 1: Mentions the importance of evaluating the reasoning process, but lacks specific evaluation methods.
                    \item 2: Designs assessment for reasoning process with clear evaluation methods and metrics.
                \end{itemize}
        \end{itemize}
    \item \textbf{Robustness Evaluation (C)}
    \hypertarget{item:28}{}
        \begin{itemize}
            \item \textbf{Explanation}: Benchmark should include testing for the model’s stability and robustness (e.g., input perturbations, adversarial samples)
            \item \textbf{Justification}: In practical applications, models may encounter different variations of inputs. Robustness testing ensures model's output is consistent and reliable under different conditions.
            \item \textbf{Scoring}:
                \begin{itemize}
                    \item 0: Does not consider robustness evaluation.
                    \item 1: Mentions the importance of evaluating the robustness of models, but lacks specific evaluation
                    \item 2: Designs assessment for robustness with clear evaluation methods and metrics.
                \end{itemize}
        \end{itemize}
    \item \textbf{Generalization Capability Evaluation (C)}
    \hypertarget{item:29}{}
        \begin{itemize}
            \item \textbf{Explanation}: The benchmark design should help evaluate the generalization capability of models to unseen data or scenarios (e.g., careful train/test split, out-of-distribution testing).
            \item \textbf{Justification}: Due to the high variability in medical scenarios, assessing a model’s generalization capability is essential for determining its reliability across different real-world scenarios, ensuring that the performance is not caused by overfitting.
            \item \textbf{Scoring}:
                \begin{itemize}
                    \item 0: Does not consider generalization capability evaluation.
                    \item 1: Mentions the importance of evaluating the generalization capability of models, but lacks specific evaluation
                    \item 2: Designs mechanism for assessing the generalization capability.
                \end{itemize}
        \end{itemize}
    \item \textbf{Uncertainty Evaluation (C)}
    \hypertarget{item:30}{}
        \begin{itemize}
            \item \textbf{Explanation}: Benchmark should include evaluation for the model’s ability to recognize and express its own uncertainty (e.g., responding “I don’t know”)
            \item \textbf{Justification}: Incorrect overconfident answers can be dangerous in high-stakes medical applications. A model’s ability to accurately identify and express uncertainty is critical for preventing incorrect and misleading decisions, enhancing system safety.
            \item \textbf{Scoring}:
                \begin{itemize}
                    \item 0: Does not consider uncertainty evaluation.
                    \item 1: Mentions the importance of evaluating the uncertainty of models, but lacks specific evaluation
                    \item 2: Designs assessment for uncertainty with clear evaluation methods and metrics.
                \end{itemize}
        \end{itemize}
    \item \textbf{Evaluation Flexibility (G)}
    \hypertarget{item:31}{}
        \begin{itemize}
            \item \textbf{Explanation}: Evaluation code should have a modular interface and support different models. Limited flexibility (e.g., hardcoded model paths, strict dependency on a single framework) hinders usability, whereas a flexible design (e.g., providing an extensible base class or API wrapper) supports both closed-source APIs and local open-source models.
            \item \textbf{Justification}: It ensures that different types of models can be tested under the same interface.
            \item \textbf{Scoring}:
                \begin{itemize}
                    \item 0: Supports only one specific model or framework, with hardcoded logic requiring significant code rewriting to adapt to new models.
                    \item 1: Supports only one type of model, but provides clear architectural guidelines or interfaces for users to implement their own extensions.
                    \item 2: Natively supports both closed-source APIs and local open-source models through a modular, unified, and user-friendly interface.
                \end{itemize}
        \end{itemize}
\end{enumerate}

\subsection{Benchmark Validity and Performance Verification}

\begin{enumerate}
    \item \textbf{Knowledge and Skill Coverage (M)}
    \hypertarget{item:32}{}
        \begin{itemize}
            \item \textbf{Explanation}: Evidence (e.g., expert evaluation) is provided to demonstrate that the benchmark content sufficiently covers the medical knowledge and skills it claims to assess.
            \item \textbf{Justification}: Sufficient coverage of the core medical competencies the benchmark aims to measure is the prerequisite for establishing content validity. When the content validity is supported with evidence, it enhances the rigor and trustworthiness of the evaluation.
            \item \textbf{Scoring}:
                \begin{itemize}
                    \item 0: No evidence regarding coverage of knowledge and skills is provided.
                    \item 1: Provide brief content analysis or expert evaluation to demonstrate the coverage of knowledge and skills.
                    \item 2: Provides both comprehensive content analysis and expert evaluation to demonstrate the coverage of knowledge and skills.
                \end{itemize}
        \end{itemize}
    \item \textbf{Scenario Authenticity (C)}
    \hypertarget{item:33}{}
        \begin{itemize}
            \item \textbf{Explanation}: The evaluation tasks of the benchmark should effectively simulate and mirror real-world medical scenario.
            \item \textbf{Justification}: Ensuring that the evaluation task closely mirrors the targeted clinical practice in real-world scenarios enhance the relevance of the benchmark. It helps assess whether the model can be transferred to real-world applications.
            \item \textbf{Scoring}:
                \begin{itemize}
                    \item 0: The evaluation task have weak relevance to real clinical scenarios (e.g., abstract knowledge  question answering)
                    \item 1: The evaluation task reflects a single clinical decision point, but does not constitute a complete workflow.
                    \item 2: The evaluation consists of multiple tasks that closely simulate realistic clinical workflows, with validation by domain experts.
                \end{itemize}
        \end{itemize}
    \item \textbf{Model Discrimination Ability (G)}
    \hypertarget{item:34}{}
        \begin{itemize}
            \item \textbf{Explanation}: Benchmark developers should provide experiment data and analysis, indicating that the benchmark can effectively distinguish the capability between different models.
            \item \textbf{Justification}: An effective benchmark should be capable of differentiating and distinguishing models of varying capabilities. It provides meaningful insights into model strengths and weaknesses, guiding future improvements and development.
            \item \textbf{Scoring}:
                \begin{itemize}
                    \item 0: No evidence is provided regarding the discrimination ability of the benchmark.
                    \item 1: The benchmark has been tested on some models and score differences are reported, but without statistical validation.
                    \item 2: The benchmark has been tested on several model, showing significant score differences with statistical validation.
                \end{itemize}
        \end{itemize}
    \item \textbf{Correlation with Clinical Performance (C)}
    \hypertarget{item:35}{}
        \begin{itemize}
            \item \textbf{Explanation}: Benchmark developers should provide evidence that preliminarily explores the correlation between benchmark performance and the model’s performance in actual clinical applications.
            \item \textbf{Justification}: Validating whether benchmark results have meaningful indication on the model’s performance in real-world clinical scenarios ensures the external validity and practical value of the benchmark and results.
            \item \textbf{Scoring}:
                \begin{itemize}
                    \item 0: Does not consider or mention the correlation between benchmark performance and clinical performance.
                    \item 1: Provides discussion about the correlation between benchmark performance and clinical performance theoretically without experiment data or concrete evidence.
                    \item 2: Preliminary research or evidence (e.g., small-scale clinical validation, consistency analysis between models and domain experts) is provided to explore and discuss the correlation between benchmark performance and clinical performance.
                \end{itemize}
        \end{itemize}
    \item \textbf{Internal Consistency (G)}
    \hypertarget{item:36}{}
        \begin{itemize}
            \item \textbf{Explanation}: For different sections or items of the benchmark evaluating the same capability, benchmark developers should demonstrate good internal consistency (e.g., Cronbach’s $\alpha$)
            \item \textbf{Justification}: It ensures that different sections or items reliably assess the same capability, enabling meaningful comparisons and interpretation of results.
            \item \textbf{Scoring}:
                \begin{itemize}
                    \item 0: Does not mention or conduct any internal consistency measurement.
                    \item 1: Mentions the importance of internal consistency, without concrete measurement.
                    \item 2: Conducts internal consistency measurement and demonstrates good consistency.
                \end{itemize}
        \end{itemize}
    \item \textbf{Statistical Significance Reporting (G)}
    \hypertarget{item:37}{}
        \begin{itemize}
            \item \textbf{Explanation}: When comparing the performance of different models, statistical significance (e.g., confidence intervals, and p-values) should be report.
            \item \textbf{Justification}: Statistical significance testing allows for a more informed interpretation of results by differentiating the performance differences between models that stem from actual capabilities, and those arises from randomness and noise.
            \item \textbf{Scoring}:
                \begin{itemize}
                    \item 0: No statistical significance is reported or considered.
                    \item 1: Reports simple statistics (e.g., mean and standard deviation of multiple executions) but does not conduct statistical testing.
                    \item 2: Reports and uses statistical significance tests (e.g., confidence interval).
                \end{itemize}
        \end{itemize}
\end{enumerate}

\subsection{Documentation, Openness and Governance}

\subsubsection{Documentation and Transparency}
\begin{enumerate}
    \item \textbf{Documentation Completeness (G)}
    \hypertarget{item:38}{}
        \begin{itemize}
            \item \textbf{Explanation}: Benchmark developers should provide a clear and comprehensive documentation of the benchmark, systematically describing the relevant details, including the design, objectives, scope, construction process, task definitions and evaluation procedures.
            \item \textbf{Justification}: A complete and clear documentation help users understand and use the benchmark properly, enhancing the usability, reproducibility and transparency.
            \item \textbf{Scoring}:
                \begin{itemize}
                    \item 0: The documentation is missing or overly simple.
                    \item 1: The documentation exists, but some important details are missing or unclearly described.
                    \item 2: The documentation is comprehensive and clear, covering all key components.
                \end{itemize}
        \end{itemize}
    \item \textbf{Clarity of Evaluation Guidelines (G)}
    \hypertarget{item:39}{}
        \begin{itemize}
            \item \textbf{Explanation}: Benchmark developers should provide evaluation guidelines, with definitions of evaluation metrics, detailed scoring criteria, and easy-to-follow usage instructions for users to replicate the evaluation process.
            \item \textbf{Justification}: Clear evaluation guidelines help users better understand how model performance is quantified, ensuring a shared interpretation and understanding. It also enhances the consistency by documenting how to replicate the evaluation process.
            \item \textbf{Scoring}:
                \begin{itemize}
                    \item 0: No evaluation guidelines are provided.
                    \item 1: Provides evaluation guidelines, but some areas are only briefly described.
                    \item 2: Provides a clear and comprehensive evaluation guidelines.
                \end{itemize}
        \end{itemize}
    \item \textbf{Discussion of Limitations and Risks (C)}
    \hypertarget{item:40}{}
        \begin{itemize}
            \item \textbf{Explanation}: The benchmark developers should openly discuss the limitations and potential social risks of the benchmark.
            \item \textbf{Justification}: Disclosing limitations and risks demonstrates scientific rigor and responsibility. This transparency helps prevent users from misunderstanding and misusing the benchmark, especially in high-stakes contexts.
            \item \textbf{Scoring}:
                \begin{itemize}
                    \item 0: Discussion of limitations and risks is not provided.
                    \item 1: Limitations and risks are briefly mentioned without in-depth discussion.
                    \item 2: In-depth and comprehensive discussion of limitations and risks is provided.
                \end{itemize}
        \end{itemize}
    \item \textbf{Peer Review (G)}
    \hypertarget{item:41}{}
        \begin{itemize}
            \item \textbf{Explanation}: The benchmark and its corresponding paper was accepted at peer-reviewed venue.
            \item \textbf{Justification}: Going through peer review process means that the design, validity and results of a benchmark has been rigorous evaluated. It enhances credibility and ensures quality.
            \item \textbf{Scoring}:
                \begin{itemize}
                    \item 0: The benchmark has not been accepted at a peer-reviewed venue.
                    \item 1: The benchmark is under review or published on platform without strict peer review (e.g., arXiv preprints).
                    \item 2: The benchmark has been published at a peer-reviewed venue.
                \end{itemize}
        \end{itemize}
\end{enumerate}

\subsubsection{Openness and Accessibility}
\begin{enumerate}
    \item \textbf{Accessibility of Evaluation Code and Data (G)}
    \hypertarget{item:42}{}
        \begin{itemize}
            \item \textbf{Explanation}: Access to the evaluation code and data, which can be shared within legal and ethical boundaries, is provided (e.g. on platforms like GitHub or Hugging Face) along with the applicable license.
            \item \textbf{Justification}: Accessible code and data are prerequisites for reproducibility. Moreover, it allows the community to review, improve, and expand the benchmarks.
            \item \textbf{Scoring}:
                \begin{itemize}
                    \item 0: Access to the evaluation code and data is not provided.
                    \item 1: The evaluation code and data are partially accessible, or accessible without clear licensing.
                    \item 2: The evaluation code and data are fully accessible and the applicable license is clearly specified.
                \end{itemize}
        \end{itemize}
    \item \textbf{Usage and Citation Guidelines (G)}
    \hypertarget{item:43}{}
        \begin{itemize}
            \item \textbf{Explanation}: Clear guidelines are provided to standardize benchmark usage, result reporting, and correct citation formats in academic papers or technical reports.
            \item \textbf{Justification}: Proper usage and citation guidelines help maintain academic integrity. It also promote standardized reporting of results, facilitating comparisons in further research.
            \item \textbf{Scoring}:
                \begin{itemize}
                    \item 0: No usage or citation guidelines are provided.
                    \item 1: Partial guidelines are provided but are incomplete.
                    \item 2: Clear and complete usage and citation guidelines are provided.
                \end{itemize}
        \end{itemize}
\end{enumerate}

\subsubsection{Continuous Maintenance and Governance}
\begin{enumerate}
    \item \textbf{Update and Version Management (G)}
    \hypertarget{item:44}{}
        \begin{itemize}
            \item \textbf{Explanation}: A clear plan or mechanism should be in place to regularly (or when necessary) update the contents of the benchmark (e.g., incorporating new data) and to effectively manage and archive different versions.
            \item \textbf{Justification}: Medical knowledge is constantly evolving, and a static Benchmark may become outdated. Establishing a mechanism for updates and version control is essential to ensure its long-term relevance and alignment with current developments.
            \item \textbf{Scoring}:
                \begin{itemize}
                    \item 0: No update plans or version management.
                    \item 1: Intention to update is mentioned, without any concrete actions.
                    \item 2: There is a clear plan for update and version management.
                \end{itemize}
        \end{itemize}
    \item \textbf{Feedback Channel for Users (G)}
    \hypertarget{item:45}{}
        \begin{itemize}
            \item \textbf{Explanation}: A feedback channel should be maintained for users to report problems, provide feedback and suggestions.
            \item \textbf{Justification}: Maintaining an effective feedback channel allows users to provide feedback when issues with the benchmark are discovered. This is crucial for continuously improving benchmark quality and fixing potential bugs and issues.
            \item \textbf{Scoring}:
                \begin{itemize}
                    \item 0: No public feedback channel is established.
                    \item 1: A public feedback channel is set up, but responses are either delayed or absent.
                    \item 2: A public feedback channel is set up with promptly response to user feedback.
                \end{itemize}
        \end{itemize}
    \item \textbf{Long-term Maintenance Responsibility (G)}
    \hypertarget{item:46}{}
        \begin{itemize}
            \item \textbf{Explanation}: The individuals, teams, or institutions responsible for its long-term maintenance and development should be clearly stated.
            \item \textbf{Justification}: Clarifying who holds long-term responsibility reassures the community that it will be actively supported and improved, ensuring usability and credibility.
            \item \textbf{Scoring}:
                \begin{itemize}
                    \item 0: No long-term maintenance responsibility is mentioned.
                    \item 1: The responsible party is implied, but not explicitly mentioned.
                    \item 2: Details of the person responsible  for long-term maintenance are clearly stated.
                \end{itemize}
        \end{itemize}
\end{enumerate}

\section{Case Study: Scoring and Explanations for a Representative Benchmark}
\label{sec:appendix_case_study}

To demonstrate the application of the \textit{MedCheck} framework, we provide the detailed scoring and explanations for a representative benchmark. This case study focuses on the objective verification of factual evidence within the benchmark's documentation across all 46 criteria.

\begin{enumerate}
    \item \textbf{Clarity of Evaluation Objectives}
    \begin{itemize}
        \item \textbf{Score}: 2
        \item \textbf{Explanation}: It clearly defines the objective of evaluating LLMs on medical knowledge question-answering and provides clear examples.
    \end{itemize}

    \item \textbf{Clarity of Application Scenario}
    \begin{itemize}
        \item \textbf{Score}: 0
        \item \textbf{Explanation}: It does not describe any specific application scenarios.
    \end{itemize}

    \item \textbf{Uniqueness and Novelty}
    \begin{itemize}
        \item \textbf{Score}: 2
        \item \textbf{Explanation}: It states that it was the first free-form multiple-choice OpenQA dataset for solving medical problems collected from the professional medical board exams, with comparisons against existing medical QA datasets.
    \end{itemize}

    \item \textbf{Target Capability of Evaluation}
    \begin{itemize}
        \item \textbf{Score}: 2
        \item \textbf{Explanation}: It clearly explains that the targeted LLM capability is knowledge understanding and require multi-hop logical reasoning.
    \end{itemize}

    \item \textbf{Medical Domain Coverage}
    \begin{itemize}
        \item \textbf{Score}: 0
        \item \textbf{Explanation}: The medical scope and coverage of specialties or disease types are not defined.
    \end{itemize}

    \item \textbf{Demonstration of User Needs}
    \begin{itemize}
        \item \textbf{Score}: 1
        \item \textbf{Explanation}: It briefly mentions the NLP community's need for OpenQA benchmarks.
    \end{itemize}

    \item \textbf{Domain Experts Involvement}
    \begin{itemize}
        \item \textbf{Score}: 2
        \item \textbf{Explanation}: It clearly describes that there are 2 medical experts with the MD degree involved to ensure the reference books in the data covered sufficient knowledge to answer the questions.
    \end{itemize}

    \item \textbf{Authoritative Knowledge Sources}
    \begin{itemize}
        \item \textbf{Score}: 2
        \item \textbf{Explanation}: The benchmark is based on authoritative medical licensing exams and relevant references books.
    \end{itemize}

    \item \textbf{Medical Standards Alignment}
    \begin{itemize}
        \item \textbf{Score}: 0
        \item \textbf{Explanation}: It does not mention any recognized medical standards.
    \end{itemize}

    \item \textbf{Validity of Core Metric}
    \begin{itemize}
        \item \textbf{Score}: 2
        \item \textbf{Explanation}: Accuracy is used as the core metric which is suitable for question-answering task.
    \end{itemize}

    \item \textbf{Multi-dimensional Evaluation}
    \begin{itemize}
        \item \textbf{Score}: 0
        \item \textbf{Explanation}: It only evaluates the correctness of answers.
    \end{itemize}

    \item \textbf{Safety and Fairness Considerations}
    \begin{itemize}
        \item \textbf{Score}: 0
        \item \textbf{Explanation}: It does not consider safety and fairness evaluation.
    \end{itemize}

    \item \textbf{Data source transparency and traceability}
    \begin{itemize}
        \item \textbf{Score}: 2
        \item \textbf{Explanation}: It clearly states the data sources are medical licensing exams and provides relevant URLs of the data sources.
    \end{itemize}

    \item \textbf{Data Source Reliability}
    \begin{itemize}
        \item \textbf{Score}: 2
        \item \textbf{Explanation}: Data is based on authoritative offical medical licensing exams.
    \end{itemize}

    \item \textbf{Data Authenticity}
    \begin{itemize}
        \item \textbf{Score}: 2
        \item \textbf{Explanation}: Data sources are offical medical licensing exams and reference books
    \end{itemize}

    \item \textbf{Dataset Representativeness}
    \begin{itemize}
        \item \textbf{Score}: 2
        \item \textbf{Explanation}: It provides quantitative statistical analysis of features including the distribution of question length and types.
    \end{itemize}

    \item \textbf{Dataset Diversity}
    \begin{itemize}
        \item \textbf{Score}: 0
        \item \textbf{Explanation}: The medical scope and coverage of specialties or disease types are not clearly described.
    \end{itemize}

    \item \textbf{Data Cleaning and Standardization}
    \begin{itemize}
        \item \textbf{Score}: 1
        \item \textbf{Explanation}: It breifly mentions the preprocessing and standardization procedures, but the description lacks detail.
    \end{itemize}

    \item \textbf{Privacy Protection}
    \begin{itemize}
        \item \textbf{Score}: 2
        \item \textbf{Explanation}: It uses publicly available data containing no sensitive patient information.
    \end{itemize}

    \item \textbf{Data Format Clarity and Consistency}
    \begin{itemize}
        \item \textbf{Score}: 2
        \item \textbf{Explanation}: The data format is clear and consistent.
    \end{itemize}

    \item \textbf{Data Review and Audit}
    \begin{itemize}
        \item \textbf{Score}: 2
        \item \textbf{Explanation}: There are 2 medical experts to ensure the reference books in the data covered sufficient knowledge to answer the questions.
    \end{itemize}

    \item \textbf{Quality of Reference Answer}
    \begin{itemize}
        \item \textbf{Score}: 2
        \item \textbf{Explanation}: The reference answers are based on authoritative medical licensing exams.
    \end{itemize}

    \item \textbf{Data Contamination Prevention}
    \begin{itemize}
        \item \textbf{Score}: 0
        \item \textbf{Explanation}: It does not mention or consider potential data contamination issues.
    \end{itemize}

    \item \textbf{User-friendliness of evaluation tools}
    \begin{itemize}
        \item \textbf{Score}: 2
        \item \textbf{Explanation}: It provides open-source code and evaluation tools in GitHub.
    \end{itemize}

    \item \textbf{Technical Reproducibility}
    \begin{itemize}
        \item \textbf{Score}: 2
        \item \textbf{Explanation}: It provides detailed technical documentation and reproduction guides.
    \end{itemize}

    \item \textbf{Provision of performance baselines}
    \begin{itemize}
        \item \textbf{Score}: 2
        \item \textbf{Explanation}: It provides at least two meaningful performance baselines (random and baseline models) with clear explanations.
    \end{itemize}

    \item \textbf{Reasoning Process Evaluation}
    \begin{itemize}
        \item \textbf{Score}: 0
        \item \textbf{Explanation}: It does not consider evaluating the reasoning process.
    \end{itemize}

    \item \textbf{Robustness Evaluation}
    \begin{itemize}
        \item \textbf{Score}: 0
        \item \textbf{Explanation}: It does not design specific assessments for robustness.
    \end{itemize}

    \item \textbf{Generalization Capability Evaluation}
    \begin{itemize}
        \item \textbf{Score}: 1
        \item \textbf{Explanation}: The importance of evaluating the generalization capability is mentioned and the cross-lingual evaluation demonstrates a certain level of generalization capability.
    \end{itemize}

    \item \textbf{Uncertainty Evaluation}
    \begin{itemize}
        \item \textbf{Score}: 0
        \item \textbf{Explanation}: It does not design specific assessments for the model's uncertainty handling.
    \end{itemize}

    \item \textbf{Evaluation Flexibility}
    \begin{itemize}
        \item \textbf{Score}: 1
        \item \textbf{Explanation}: It supports multiple evaluation methods but offers limited flexibility, requiring users to perform extensions themselves, though guidelines are provided.
    \end{itemize}

    \item \textbf{Knowledge and Skill Coverage}
    \begin{itemize}
        \item \textbf{Score}: 1
        \item \textbf{Explanation}: There are 61097 questions from medical licensing exams in different countries covering questions of different length and types.
    \end{itemize}

    \item \textbf{Scenario Authenticity}
    \begin{itemize}
        \item \textbf{Score}: 0
        \item \textbf{Explanation}: The evaluation task is abstract knowledge question-answering, which has weak relevance to realistic clinical workflows or decision-making scenarios.
    \end{itemize}

    \item \textbf{Model Discrimination Ability}
    \begin{itemize}
        \item \textbf{Score}: 1
        \item \textbf{Explanation}: The benchmark has been tested on several models and score differences are reported, but without statistical validation.
    \end{itemize}

    \item \textbf{Correlation with Clinical Performance}
    \begin{itemize}
        \item \textbf{Score}: 0
        \item \textbf{Explanation}: It does not mention or provide evidence exploring the correlation between benchmark performance and actual clinical application performance.
    \end{itemize}

    \item \textbf{Internal Consistency}
    \begin{itemize}
        \item \textbf{Score}: 0
        \item \textbf{Explanation}: It does not mention or conduct any internal consistency measurement.
    \end{itemize}

    \item \textbf{Statistical Significance Reporting}
    \begin{itemize}
        \item \textbf{Score}: 0
        \item \textbf{Explanation}: It only reports results from single runs.
    \end{itemize}

    \item \textbf{Documentation Completeness}
    \begin{itemize}
        \item \textbf{Score}: 1
        \item \textbf{Explanation}: Documentations are available but some details are missing including the data cleaning process. 
    \end{itemize}

    \item \textbf{Clarity of Evaluation Guidelines}
    \begin{itemize}
        \item \textbf{Score}: 1
        \item \textbf{Explanation}: Evaluation guidelines are provided, but there could be more detailed instructions.
    \end{itemize}

    \item \textbf{Discussion of Limitations and Risks}
    \begin{itemize}
        \item \textbf{Score}: 0
        \item \textbf{Explanation}: It does not discuss any limitations or potential social risks of the benchmark.
    \end{itemize}

    \item \textbf{Peer Review}
    \begin{itemize}
        \item \textbf{Score}: 2
        \item \textbf{Explanation}: The paper was published in Applied Sciences.
    \end{itemize}

    \item \textbf{Accessibility of Evaluation Code and Data}
    \begin{itemize}
        \item \textbf{Score}: 2
        \item \textbf{Explanation}: The evaluation code and data are accessible on GitHub, and licensing is specified.
    \end{itemize}

    \item \textbf{Usage and Citation Guidelines}
    \begin{itemize}
        \item \textbf{Score}: 1
        \item \textbf{Explanation}: Evaluation guidelines are provided, but there could be more detailed instructions.
    \end{itemize}

    \item \textbf{Update and Version Management}
    \begin{itemize}
        \item \textbf{Score}: 1
        \item \textbf{Explanation}: The benchmark is managed in GitHub but there are no clear update plans.
    \end{itemize}

    \item \textbf{Feedback Channel for Users}
    \begin{itemize}
        \item \textbf{Score}: 2
        \item \textbf{Explanation}: A public feedback channel is available via GitHub Issue with responses from the authors.
    \end{itemize}

    \item \textbf{Long-term Maintenance Responsibility}
    \begin{itemize}
        \item \textbf{Score}: 1
        \item \textbf{Explanation}: Team information is available, but the explicit, long-term maintenance responsibility is not clearly stated.
    \end{itemize}
\end{enumerate}

\section{Actionable Diagnostic Report Example}
\label{sec:appendix_clinical_actionable_report}

This section provides a complete actionable diagnostic report based on our lifecycle assessment of a high-impact clinical-oriented benchmark. By focusing on criteria where the benchmark scored 0 or 1, this report illustrates how \textit{MedCheck} surfaces actionable technical and procedural recommendations.

\subsection*{Phase I: Design and Conceptualization}
\begin{itemize}
    \item \textbf{Criterion 9 (Medical Standards Alignment) - Score: 1} \\
    \textit{Weakness:} Mentions standardization but lacks explicit mapping to standard ontologies. \\
    \textit{Actionable Recommendation:} Require model outputs to strictly map to standardized terminologies such as SNOMED CT or LOINC codes.
    \item \textbf{Criterion 12 (Safety and Fairness Considerations) - Score: 1} \\
    \textit{Weakness:} Conceptual discussion of safety without concrete empirical test cases. \\
    \textit{Actionable Recommendation:} Introduce a dedicated "clinical red-teaming" subset to test for harmful hallucinations or demographic biases.
\end{itemize}

\subsection*{Phase II: Dataset Construction and Management}
\begin{itemize}
    \item \textbf{Criterion 16 (Dataset Representativeness) - Score: 1} \\
    \textit{Weakness:} Qualitative demographic description lacks rigorous statistical analysis. \\
    \textit{Actionable Recommendation:} Publish detailed statistical tables comparing dataset demographics to real-world clinical population distributions.
    \item \textbf{Criterion 23 (Data Contamination Prevention) - Score: 0} \\
    \textit{Weakness:} High risk of data memorization due to the use of public clinical databases. \\
    \textit{Actionable Recommendation:} Conduct n-gram overlap analysis against common pre-training corpora and inject unique "canary strings" into the test set.
\end{itemize}

\subsection*{Phase III: Technical Implementation and Evaluation Methodology}
\begin{itemize}
    \item \textbf{Criterion 27 (Reasoning Process Evaluation) - Score: 0} \\
    \textit{Weakness:} "Black box" evaluation that cannot verify clinical logic. \\
    \textit{Actionable Recommendation:} Implement Chain-of-Thought (CoT) metrics or expert-defined reasoning path verification.
    \item \textbf{Criterion 28 (Robustness Evaluation) - Score: 0} \\
    \textit{Weakness:} Overestimates performance by assuming noise-free clinical inputs. \\
    \textit{Actionable Recommendation:} Introduce programmatic input perturbations (e.g., medical abbreviations, simulated typos) to test resilience.
    \item \textbf{Criterion 30 (Uncertainty Evaluation) - Score: 0} \\
    \textit{Weakness:} Rewards overconfident hallucinations over safe abstention. \\
    \textit{Actionable Recommendation:} Include "unanswerable" EHR cases and reward the model for correctly outputting "Insufficient data to decide."
\end{itemize}

\subsection*{Phase IV: Benchmark Validity and Performance Verification}
\begin{itemize}
    \item \textbf{Criterion 35 (Correlation with Clinical Performance) - Score: 1} \\
    \textit{Weakness:} Reliance on NLP metrics (e.g., ROUGE) that may misalign with clinical utility. \\
    \textit{Actionable Recommendation:} Conduct a clinician-in-the-loop study comparing automated scores with physician preference ratings.
    \item \textbf{Criterion 37 (Statistical Significance Reporting) - Score: 1} \\
    \textit{Weakness:} Reports point estimates without confidence intervals. \\
    \textit{Actionable Recommendation:} Use bootstrapping to report p-values and confidence intervals for all model comparisons.
\end{itemize}

\subsection*{Phase V: Documentation, Openness, and Governance}
\begin{itemize}
    \item \textbf{Criterion 40 (Discussion of Limitations and Risks) - Score: 1} \\
    \textit{Weakness:} Insufficient risk communication for actual clinical deployment. \\
    \textit{Actionable Recommendation:} Add a dedicated "Broader Impacts" section analyzing clinical deployment risks.
    \item \textbf{Criterion 46 (Long-term Maintenance Responsibility) - Score: 1} \\
    \textit{Weakness:} High risk of becoming "abandonware" due to lack of institutional commitment. \\
    \textit{Actionable Recommendation:} Explicitly state the responsible maintaining body and outline a 3-year sustainability plan.
\end{itemize}

\end{document}